\documentclass{article}

\PassOptionsToPackage{numbers, sort, compress}{natbib}

 \usepackage[main, final]{neurips_2026}

\usepackage[utf8]{inputenc} 
\usepackage[T1]{fontenc}    
\usepackage{hyperref}       
\usepackage{url}            
\usepackage{booktabs}       
\usepackage{amsfonts}       
\usepackage{nicefrac}       
\usepackage{microtype}      
\usepackage{xcolor}         
\usepackage{algorithm}
\usepackage{algorithmic}
\usepackage{tikz}
\usepackage{float}
\usepackage{xcolor}
\usepackage{amsmath, amssymb, amsthm}
\theoremstyle{plain}

\usepackage{graphicx}
\usepackage{subcaption}

\usepackage{array}
\usepackage{tabularx}
\usepackage{makecell}
\usepackage{caption}
\usepackage{placeins}
\usepackage{titletoc}

\newcommand{\grouprownine}[1]{\midrule\multicolumn{9}{@{}l@{}}{\textit{#1}}\\\midrule}
\newcommand{\grouprowsix}[1]{\midrule\multicolumn{6}{@{}l@{}}{\textit{#1}}\\\midrule}
\definecolor{OptiFlow}{rgb}{0.05,0.35,0.75}

\newcommand{\val}[2]{\ensuremath{#1{\scriptstyle\pm #2}}}

\newcommand{\scorecell}[4]{%
    \pgfmathsetmacro{\xmin}{#1-1}\pgfmathsetmacro{\xmax}{#1}\pgfmathsetmacro{\xmid}{#1-0.5}
    \pgfmathsetmacro{\ymin}{#2-1}\pgfmathsetmacro{\ymax}{#2}\pgfmathsetmacro{\ymid}{#2-0.5}
    \if -#3%
        \filldraw[fill=gray!10, draw=white, line width=1.2pt] (\xmin,\ymin) rectangle (\xmax,\ymax);
        \node[font=\small,gray] at (\xmid,\ymid) {--};
    \else
        \pgfmathtruncatemacro{\hmshade}{min(75,max(12,round(#3*0.75)))}
        \ifnum#3<3 \def\cellcolor{gray!20}\else \edef\cellcolor{OptiFlow!\hmshade!white}\fi
        \filldraw[fill=\cellcolor, draw=white, line width=1.2pt] (\xmin,\ymin) rectangle (\xmax,\ymax);
        \node[font=\large] at (\xmid,\ymid+0.10) {#3};
        \node[font=\tiny]  at (\xmid,\ymid-0.18) {$\pm #4$};
    \fi
}
\newcommand{\itres}[1]{\ensuremath{\mathit{#1}}}
\newcommand{\scorecellnoerr}[3]{%
    \pgfmathsetmacro{\xmin}{#1-1}
    \pgfmathsetmacro{\xmax}{#1}
    \pgfmathsetmacro{\xmid}{#1-0.5}
    \pgfmathsetmacro{\ymin}{#2-1}
    \pgfmathsetmacro{\ymax}{#2}
    \pgfmathsetmacro{\ymid}{#2-0.5}
    \pgfmathtruncatemacro{\hmshade}{min(75,max(12,round(#3*0.75)))}

    \ifnum#3<3
        \def\cellcolor{gray!20}
    \else
        \edef\cellcolor{green!\hmshade!white}
    \fi

    \filldraw[fill=\cellcolor, draw=white, line width=1.2pt]
        (\xmin,\ymin) rectangle (\xmax,\ymax);

    \node[font=\Large] at (\xmid,\ymid) {#3};
}

\newcommand{\OptiFlowaxes}{%
    \node[anchor=east, font=\small] at (-0.10,2.5) {$\eta^*/10$};
    \node[anchor=east, font=\small] at (-0.10,1.5) {$\eta^*$};
    \node[anchor=east, font=\small] at (-0.10,0.5) {$10{\times}\eta^*$};

    \node[font=\small] at (0.5,-0.25) {$\tau^*/2$};
    \node[font=\small] at (1.5,-0.25) {$\tau^*$};
    \node[font=\small] at (2.5,-0.25) {$2{\times}\tau^*$};
}
\newcounter{step}

\DeclareMathOperator*{\argmin}{arg\,min}
\newcommand{\cS}{\mathcal{S}}
\newcommand{\cA}{\mathcal{A}}
\newcommand{\cD}{\mathcal{D}}
\newcommand{\cN}{\mathcal{N}}
\newcommand{\cL}{\mathcal{L}}
\newcommand{\R}{\mathbb{R}}
\newcommand{\E}{\mathbb{E}}
\newcommand{\red}[1]{{\textcolor{red}{#1}}}

\title{Learning Multimodal One-step Flow Policy via Value-weighted Optimal Transport}

\author{%
  Jaehun Shon\textsuperscript{*} \qquad
  Jinha Choi\textsuperscript{*} \qquad
  Jongwook Jeon \qquad
  Jongmin Lee\textsuperscript{\dag} \\
  Department of Artificial Intelligence, Yonsei University \\
  \texttt{\{manfromearth, danielc174, jjwook2602, jongminlee\}@yonsei.ac.kr} \\
  \textsuperscript{*}Equal contribution \quad
  \textsuperscript{\dag}Corresponding author
}

\begin{document}

\maketitle

\begin{abstract}
Offline reinforcement learning aims to learn a policy solely from fixed datasets, which often contain multimodal action distributions. Flow policies can naturally represent such multimodal behaviors, but learning an efficient one-step flow policy remains challenging: standard value guidance often leads to mode collapse or exploits overestimation bias in out-of-distribution regions. To address this, we introduce \textbf{One-step Flow policy via Optimal Transport (OptiFlow)}, a framework for one-step flow policy learning as a structured sample-allocation problem. OptiFlow jointly trains a value-aware reference flow policy and an efficient one-step policy, coupling their action samples through state-wise entropic optimal transport. For each state, critic-estimated values define the priority of distillation target actions, while the action-distance cost ensures geometrically compatible pairings. By avoiding direct critic maximization, our transport-guided approach enables in-distribution exploitation by anchoring the one-step policy to high-value, dataset-supported modes without the risk of out-of-distribution divergence. Experimental results demonstrate that OptiFlow effectively captures optimal multimodal behaviors and achieves strong performance across diverse offline RL benchmarks. Our code is available at \url{https://github.com/Yonsei-DILLab/OptiFlow}.
\end{abstract}

\section{Introduction}

Offline reinforcement learning (RL) aims to learn decision-making policies from fixed datasets without further environment interaction~\citep{Lange2012, DBLP:journals/corr/abs-2005-01643}. This setting is crucial in domains such as robotics and control, where online exploration can be costly or unsafe. A central challenge is distribution shift: when a learned policy selects actions poorly supported by the offline data, value estimates can become unreliable and be amplified during policy optimization. Standard methods mitigate this through conservative value estimation~\citep{kumar2020conservative,kostrikov2022offline}, behavior-constrained policy learning~\citep{wu2019behavior,fujimoto2021minimalist}, \textcolor{black}{or distribution correction via dual optimization~\citep{lee2021optidiceofflinepolicyoptimization, nachum2019algaedicepolicygradientarbitrary}.}

Beyond distribution shift, modern offline datasets often exhibit multimodal behavior: for a given state, multiple distinct actions may be valid and lead to high return. Standard unimodal policies can struggle in such settings, often collapsing toward a mean action that is suboptimal or physically invalid~\citep{fujimoto2021minimalist,park2024value}. This issue is especially important in high-dimensional manipulation and control tasks, where multiple successful strategies may coexist. Expressive generative policies, including diffusion-based~\citep{ho2020denoising,esser2024scaling,liu2022flow} and flow-based~\citep{lipman2023flow} policies, provide a natural way to model such multimodal action distributions~\citep{hansenestruch2023idql,wang2022diffusion,park2025flow,tiofack2026guided,jang2025qguided}. Flow policies are particularly appealing because their deterministic probability paths make one-step policy learning a natural target for efficient deployment~\citep{esser2024scaling,liu2022flow}.

However, learning an effective one-step flow policy remains challenging. Multi-step sampling is often too expensive for real-time control, but compressing a reference policy into one step can lose multimodal structure~\citep{salimans2022progressive}. Pointwise distillation, which matches the same noise input to a target action, can force a one-step model to memorize complex paths and collapse distinct behaviors into compromised averages~\citep{frans2024one}. Direct critic maximization introduces a different failure mode: the policy may exploit local optima or overestimated out-of-distribution actions instead of capturing high-return in-distribution behavior.

We address these limitations by reframing one-step flow policy learning as a structured sample-allocation problem. We propose \textbf{Flow Policy via Optimal Transport (OptiFlow)}, which couples samples from a value-aware reference flow policy and an efficient one-step policy through entropic optimal transport~\citep{cuturi2013sinkhorn}. OptiFlow separates value from geometry: critic-estimated values determine the supervision mass by prioritizing high-return reference actions, while the transport cost encourages geometrically compatible pairings between one-step-policy samples and reference actions.

This transport-guided update avoids direct critic maximization of the deployed policy. Instead of following unreliable critic gradients into OOD regions or local optima, the one-step policy is anchored to high-value actions proposed by the behavior-regularized reference policy. At the same time, by relaxing strict pointwise matching OptiFlow lets the one-step policy flexibly map noise to actions and preserve multimodal reference-policy structure. We empirically show that OptiFlow captures multimodal behavior in controlled diagnostics and achieves strong performance across offline RL benchmarks.

\section{Related Work}

\paragraph{Expressive generative policies for offline RL.}
Recent work has introduced diffusion and flow-based policies as expressive alternatives to unimodal actors for modeling multimodal action distributions in offline RL~\citep{hansenestruch2023idql,wang2022diffusion,park2025flow,tiofack2026guided,jang2025qguided}.
\textcolor{black}{ The same expressiveness has also been leveraged in multi-agent imitation learning, where decentralized diffusion policies recover multiple coordination modes from multimodal demonstrations~\citep{dong2026mimicdmultimodalimitationmultiagent}.}
FQL~\citep{park2025flow} learns flow-based policies and uses critic-guided policy improvement to obtain strong offline RL performance. GFP~\citep{tiofack2026guided} introduces value-aware behavior cloning for flow policies together with one-step distillation. 
OptiFlow builds on this line of work, but addresses a different question: how should value-guided supervision be allocated across multiple action samples when the reference flow policy contains several high-value modes? Rather than directly optimizing the one-step policy against the critic or applying guidance only at inference time, OptiFlow uses critic-estimated values to define a reference-policy marginal in an optimal-transport coupling.

\paragraph{Value-guided policy learning and behavior constraints.}
A central theme in offline RL is to improve policy quality while avoiding unsupported actions. Prior methods address this through conservative value estimation~\citep{kumar2020conservative,kostrikov2022offline}, behavior regularization~\citep{wu2019behavior,fujimoto2021minimalist}, or value-weighted policy extraction~\citep{peng2019advantage,wang2020critic}. These approaches provide different mechanisms for balancing value improvement and support preservation. OptiFlow follows the same broad principle, but uses value information differently: critic-estimated values do not directly optimize the deployed one-step policy, but instead shape the reference-policy marginal used for transport-guided supervision.

\paragraph{Matching, retrieval, and optimal transport.}
Several offline RL and imitation learning methods use matching or retrieval mechanisms to keep learned policies close to data-supported actions. For example, retrieval-based policy constraints regularize the policy toward nearby dataset actions~\citep{ran2023policy}. Optimal transport has also been used to compare or align behavior distributions in imitation learning and offline decision making~\citep{luo2023optimal,chang2023imitation,asadulaev2024rethinking}. OptiFlow differs in both purpose and construction. We do not use optimal transport merely as a distributional distance or trajectory-level alignment objective. Instead, OptiFlow solves a state-wise entropic optimal transport problem between one-step-policy and reference-policy action samples. The resulting coupling combines a value-weighted reference-policy marginal with action-space geometry and produces transport-guided regression targets for efficient one-step flow-policy learning.


\section{Preliminaries}

\subsection{Offline Reinforcement Learning}

We consider a discounted Markov decision process 
$\mathcal M=\langle \cS,\cA,P,r,\rho_0,\gamma \rangle$~\citep{Sutton1998}, where 
$\mathcal S$ and $\mathcal A$ denote the state and action spaces, $P: \cS \times \cA \rightarrow \Delta(\cS)$ is the transition kernel,
$r: \mathcal{S} \times \mathcal{A} \rightarrow \R$ is the reward function, $\rho_0 \in \Delta(\cS)$ is the initial-state distribution, and $\gamma\in[0,1)$ is the discount factor. 
In offline RL, the learner is given a fixed dataset 
$\mathcal D=\{(s_i,a_i,r_i,s_i')\}_{i=1}^{|\mathcal D|}$ 
collected by potentially diverse and unknown behavior policies and cannot collect additional data from environment interactions.

For a policy $\pi$, we write 
$Q^\pi(s,a)$ and $V^\pi(s)$ for its action-value and value functions, 
$J(\pi)=\mathbb E_{\pi}\left[\sum_{t=0}^{\infty}\gamma^t r(s_t,a_t)\right]$ 
for its discounted return, and 
$d^\pi(s)=(1-\gamma)\sum_{t=0}^{\infty}\gamma^t 
\Pr_{\pi}(s_t=s)$ for its discounted state occupancy. 
OptiFlow learns a critic estimate $Q_\phi(s,a)$ from $\mathcal D$ and uses critic-estimated values to construct a value-weighted transport problem for updating a one-step flow policy.

\subsection{Flow Policies}

Flow matching~\citep{lipman2023flow, albergo2022building} learns a time-dependent vector field that transports samples from a simple base distribution to a target distribution. 
In offline RL, we apply this to conditional action generation, where the vector field $v_\theta(s, x_t, t)$ is conditioned on the state $s$. For behavioral cloning, we set the base sample as noise $z \sim \cN(0, I)$ and the target action as an action $a \sim \cD$ from the dataset. Defining the linear interpolation $x^t=(1-t)x^0+t x^1$ for $t \in [0,1]$, $z=x^0$, and $a=x^1$, the most basic flow-matching objective for BC is as follows:
\begin{align}
\min_{\theta} \cL_{\mathrm{BC}}(\theta) = \mathbb E_{\substack{(s, a=x^1) \sim \cD, \\ z=x^0 \sim \cN(0, I_d), \\ t \sim \mathrm{Unif}([0,1])}}
\left[
\left\|v_\theta(s,x^t,t)-(x^1-x^0)\right\|_2^2
\right].
\end{align}

At inference time, an action is generated by sampling $z \sim \cN(0, I_d)$ and integrating the ODE from $t=0$ to $t=1$:
$\frac{d x^t}{dt}=v_\theta(s,x^t,t)$, starting from $z = x^0$, to obtain $a = x^1$.
In practice, this integration is approximated by a finite number of Euler steps:
\begin{align}
x^{k+1}=x^k + \tfrac{1}{K} v_\theta\left(s,x^k,\tfrac{k}{K}\right) \text{ for } k=0,\dots,K-1.
\end{align}
Larger $K$ gives an expressive multi-step flow policy but requires iterative action generation. To enable efficient deployment, a one-step flow policy directly learns a mapping $a = \mu_\theta(s,z)$ from noise to action, which induces a stochastic policy $\pi_\theta(a|s)$ due to the stochasticity of the base noise $z$.
OptiFlow uses a value-aware \emph{multi-step} reference flow policy to model multimodal behavior and learns an efficient \emph{one-step} flow policy $\mu_\theta(s,z)$ through transport-guided updates.

\subsection{Entropic Optimal Transport}

Let $p\in\Delta^N$ and $q\in\Delta^M$ be discrete probability distributions over two finite sets, and let $C\in\mathbb R^{N\times M}$ be a cost matrix. The discrete optimal transport problem seeks a coupling 
$P\in\mathbb R_+^{N\times M}$ whose marginals match $p$ and $q$:
\begin{align}
\Pi(p,q) =
\left\{
P\ge 0 \mid P\mathbf 1_M=p,\; P^\top \mathbf 1_N=q
\right\}.
\end{align}
Exact optimal transport requires solving a linear program, which scales poorly and cannot efficiently leverage GPU parallelization. For efficient and scalable computation, we consider the entropy-regularized optimal transport problem~\citep{cuturi2013sinkhorn}:
\begin{align}
P^* =
\argmin_{P\in\Pi(p,q)} \Big[ \langle P,C \rangle
+
\varepsilon\sum_{i,j}P_{ij}(\log P_{ij}-1)
\Big],
\end{align}%
where $\varepsilon>0$ controls the smoothness of the transport plan.
This regularized problem admits a unique solution that can be computed efficiently using the Sinkhorn algorithm~\citep{cuturi2013sinkhorn}, an iterative matrix scaling procedure. The detailed formulation and the corresponding pseudocode are provided in Appendix~\ref{app:sinkhorn}.
In the following section, we detail how OptiFlow leverages this optimal transport framework state-wise to align an efficient one-step flow policy with a multimodal reference policy.


\begin{figure}[t]
    \centering
    \includegraphics[width=\linewidth]{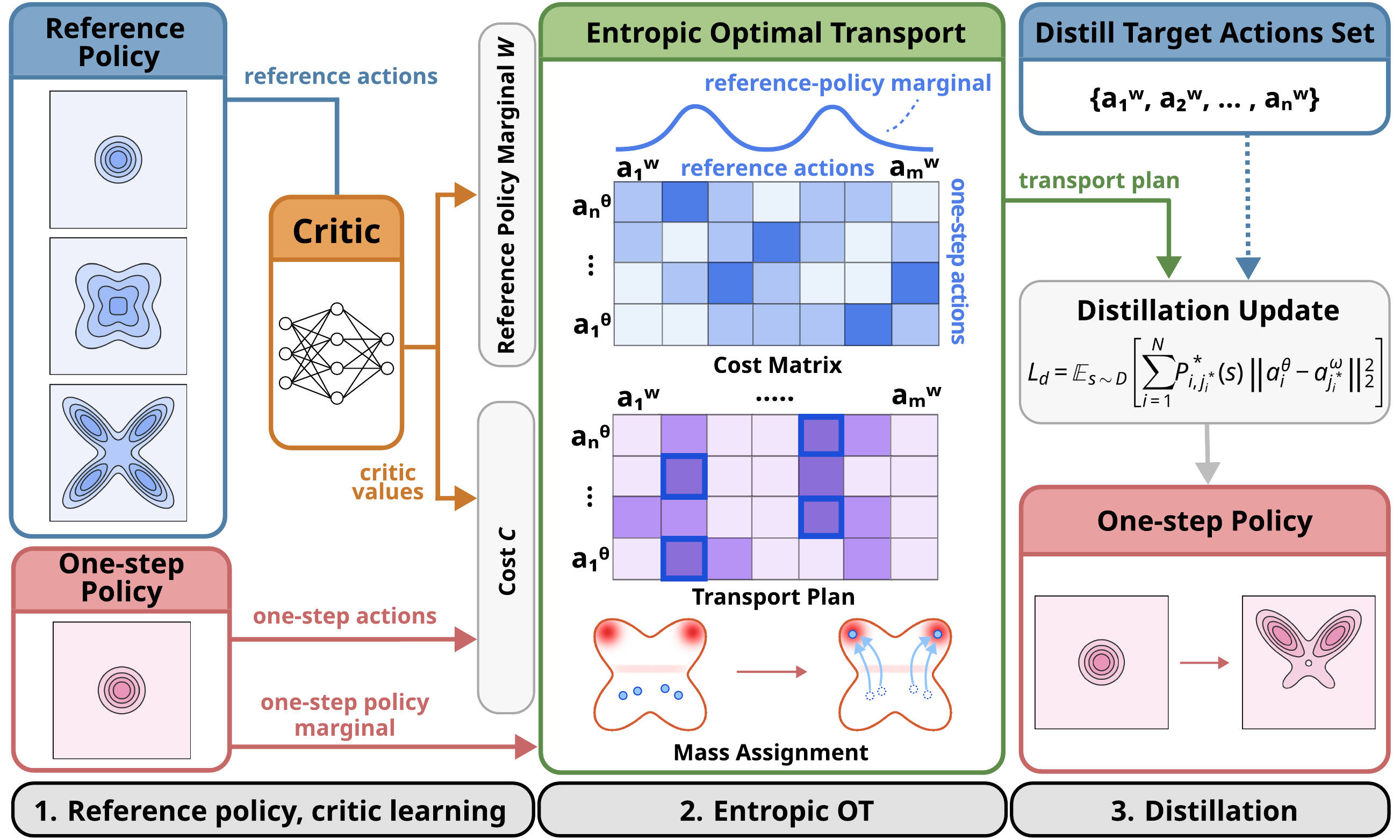}
    \caption{Overview of OptiFlow. OptiFlow learns a critic and a reference flow policy from offline data, then couples reference and one-step policy action samples with entropic optimal transport. Critic values define a reference marginal over reference actions, and the resulting transport plan provides structured distillation targets for the one-step policy.}
    \label{fig:OptiFlow}
\end{figure}

\section{Flow Policy via Optimal Transport}
\label{sec:method}

We introduce \textbf{Flow Policy via Optimal Transport (OptiFlow)}, a framework that learns efficient one-step flow policies for offline RL through a value-weighted optimal transport formulation.
While one-step flow policies are expressive enough to represent multimodal action distributions in principle, existing approaches suffer from two distinct failure modes: (i)~Mode collapse during mapping, where pointwise noise-to-action matching is prone to collapsing into compromised averages, (ii)~Prone to local optima or OOD actions, where direct critic maximization can pull the one-step policy toward local optima or out-of-distribution regions.
OptiFlow addresses these challenges by reframing one-step policy learning as a \textbf{structured sample-allocation problem}. We decouple the policy improvement process into three mechanisms: (i)~learning a value-aware multi-step reference flow policy anchored in the dataset, (ii)~constructing a value-aware geometric coupling via entropic OT, and (iii)~extracting the one-step policy through transport-guided target selection. An overview is illustrated in Figure~\ref{fig:OptiFlow}.



\paragraph{Step 1: Learning the critic.}
The first step is to train an action-value function $Q_\phi(s,a)$ from the offline dataset $\cD=\{(s_i,a_i,r_i,s_i')\}_{i=1}^{|\cD|}$ using a standard temporal-difference objective for the one-step policy $\mu_\theta$:
\begin{align}
\label{eq:q_loss}
\min_{\phi} \mathcal{L}_Q(\phi)
=
\E_{(s,a,r,s') \sim \cD,\, z' \sim \mathcal{N}(0,I_d)}
\left[
\Big( Q_\phi(s,a)
- \big( r + \gamma Q_{\bar{\phi}}(s', \mu_\theta(s',z')) \big)
\Big)^2
\right],
\end{align}%
where $Q_{\bar{\phi}}$ denotes a target critic network~\citep{mnih2013playing}. In OptiFlow, the critic is used in two ways: it defines the priority weights for training the reference flow policy, and it constructs the reference-side marginal for the optimal transport problem. Additional implementation variants of the TD target are described in Appendix~\ref{app:impl}.

\paragraph{Step 2: Learning the multi-step reference flow policy.}
The next step of OptiFlow is to establish a data-supported, high-value reference policy $\mu_\omega$. To this end, we adopt the Value-aware behavior cloning (VaBC)~\citep{tiofack2026guided} objective to train a multi-step reference flow policy. Given a dataset action $a$ and noise $\epsilon \sim \cN(0, I_d)$, the velocity field $v_\omega$ is trained via:
\begin{align}
\label{eq:ref_loss}
\min_{\omega} \mathcal{L}_{\mathrm{ref}}(\omega)
&=
\mathbb{E}_{(s,a)\sim\cD,\epsilon \sim \cN(0, I_d),t\sim\textrm{Unif}([0,1])}
\left[
g_\eta(s,a)
\left\| v_\omega(s,a_t,t) - (a-\epsilon)\right\|_2^2
\right] \\
\text{where }
g_\eta(s,a)
&=
\frac{
\exp\Big( \tfrac{\lambda}{\eta} Q_{\phi}(s,a)\Big)
}{
\exp\Big(\tfrac{\lambda}{\eta} Q_{\phi}(s,a)\Big)
+
\exp\Big(\tfrac{\lambda}{\eta} Q_{\phi}(s,\mu_\theta(s,z)) \Big)
}, \quad z \sim \cN(0, I_d).
\end{align}
where 
$\lambda = \left(
\frac{1}{B}\sum_{i=1}^{B}\left|Q_{{\phi}}(s_i,a_i)\right|
+ \epsilon_q \right)^{-1}
$ normalizes the critic scale within a minibatch of size $B$, with a small constant $\epsilon_q > 0$.
Here, $g_\eta(s,a)$ assigns higher weights to dataset actions that outperform the current one-step policy $\mu_\theta(s,z)$ as estimated by the critic $Q_\phi$. This objective clones dataset actions with high values. Consequently, $\mu_\omega$ successfully captures the multi-modal structure of the data while concentrating probability density on high-return regions Importantly, the reference multi-step flow policy $\mu_\omega$ serves as a \emph{support-preserving proposal distribution} that provides a high-quality candidate action set. This ensures that the subsequent OptiFlow's sample allocation is restricted to regions that are both in-distribution and high-performing, effectively filtering out suboptimal or unsupported actions before the final one-step policy extraction.

\setcounter{step}{3}\refstepcounter{step}\label{step:ot}
\paragraph{Step 3: Value-weighted transport coupling.}
With the \emph{multi-step} reference policy $\mu_\omega$ providing a set of high-quality, in-distribution candidate actions, the next challenge is to allocate these targets to the \emph{one-step} policy $\mu_\theta$ without inducing mode collapse or geometric misalignment.
For each state $s$, we sample $N$ actions from the one-step policy and $M$ actions from the reference flow policy:
\begin{align}
a_{\theta,i}^s &=\mu_\theta(s,z_i),
\hspace{23pt}
z_i \sim\cN(0,I_d),
\qquad
i=1,\dots,N,
\\
\tilde a_{\omega,j}^s &=\mu_\omega(s,z_j),
\hspace{20pt}
z_j \sim\cN(0,I_d),
\qquad
j=1,\dots,M.
\end{align}
To decouple the \emph{value guidance} from \emph{geometric alignment}, OptiFlow formulates the sample allocation as an optimal transport problem with \textbf{asymmetric marginal constraints}. Specifically, we impose a uniform marginal constraint $p \in \Delta^N$, with $p_i = \frac{1}{N}$, on the one-step policy samples, and a value-weighted marginal constraint $q^s \in \Delta^M$ on the reference policy actions:
\begin{align}
q_j^s
=
\frac{\exp( Q_\phi(s,\tilde a_{\omega,j}^s)/\tau)}
{\sum_{k=1}^M \exp(Q_\phi(s,\tilde a_{\omega,k}^s)/\tau)}.
\label{eq:qjs}
\end{align}
where $\tau > 0 $ is a temperature parameter. This asymmetric marginal design is essential for learning a valid one-step flow policy. The uniform constraint $p$ ensures that all base noise samples $z_i \sim \cN(0, I_d)$ are used equally, preserving the base distribution. Conversely, the value-weighted constraint $q$ injects the critic's guidance directly into the allocation process. By assigning larger probability mass to high-return reference actions, the transport plan maps the majority of one-step samples to high-value modes.

Then, to complete this decoupling, we define the optimal transport cost solely based on geometric proximity in the action space, without any value-based terms.
\begin{align}
C_{ij}^s = \frac{1}{\bar c^s} \|a_{\theta,i}^s-\tilde a_{\omega,j}^s\|_2^2,
\label{eq:cost}
\end{align}
where $\bar c^s$ denotes the per-state mean squared distance over all action pairs. With the asymmetric marginals $(p, q^s)$ that determines the target weights and the cost matrix $C$ that encodes the geometry, we compute the entropic optimal transport plan:
\begin{align}
P^{s,*}
=
\argmin_{P\in\Pi(p,\textcolor{red}{q^s})}
\left[
\langle P,C^s\rangle
+
\varepsilon \sum_{i,j}P_{ij}(\log P_{ij}-1)
\right],
\end{align}
where $\Pi(p, \red{q^s}) = \{P\ge 0\mid P\mathbf 1_M = \tfrac{1}{N} \mathbf 1_N,~ \red{P^\top\mathbf 1_N=q^s} \}$ denotes the set of valid couplings, and $\varepsilon > 0$ is the entropic regularization hyperparameter. The optimal transport plan $P^{s,*}$ is efficiently computed by Sinkhorn algorithm (see Appendix~\ref{app:sinkhorn} for more details).
By keeping the critic's influence isolated within the $q^s$-marginal constraints and the spatial structure isolated within the cost matrix, OptiFlow ensures that probability mass is allocated proportionally to high-value action modes without compromising the geometric shape of the action distribution.
This is in contrast to existing behavior-regularized offline RL methods that rely on an explicit behavioral penalty term. While strong regularization prevents out-of-distribution actions, it forces the policy to cover suboptimal dataset actions. OptiFlow circumvents this trade-off by leveraging the optimal transport coupling for one-step policy extraction. Rather than relying on a global behavioral penalty, the transport plan directly maps the one-step policy only to high-value modes, naturally discarding suboptimal candidates.


\paragraph{Step 4: Transport-guided distillation.}
Given the optimal coupling $P^{s,*}$, the final step is to learn the efficient one-step policy through a structured, transport-guided distillation. While the optimal coupling provides a soft assignment, practical one-step policy learning requires explicit, unambiguous regression targets. Thus, we extract a per-row reference action index via maximum assignment:
\begin{align}
j_i^* \in \arg\max_j P^{s,*}_{ij}, \qquad i=1,\dots,N.
\end{align}
The one-step policy is then updated by minimizing the mass-weighted distillation loss:
\begin{align}
\label{eq:distill_loss}
\min_{\theta} \mathcal L_{\mathrm{distill}}(\theta)
=
\mathbb E_{s\sim\mathcal D} \left[ \sum_{i=1}^N P^{s,*}_{i,j_i^*} \left\| \mu_\theta(s,z_i) -
\tilde a_{\omega, j_i^*}^s \right\|_2^2
\right].
\end{align}
During this update, the reference actions $\tilde a_{\omega, j_i^*}^s$ serve as fixed distillation targets, while the transport weights $P_{i,j_i^*}^{s,*}$ and hard assignments $j_i^*$ are treated as constants; gradients flow exclusively through the one-step-policy actions $\mu_\theta(s,z_i)$. Unlike pointwise distillation that simply matches shared noise inputs and action outputs and often collapses multimodal behaviors into a compromised mean, each distillation target in OptiFlow is explicitly routed to a geometrically aligned, distinct high-value mode via optimal transport, preserving the multimodal structure.

\paragraph{Algorithm Summary.}
To sum up, the full OptiFlow (Algorithm~\ref{alg:OptiFlow}) extracts an effective one-step policy that preserves the multimodal structure of high-value behaviors. It leverages a learned critic (Step~1) and a reference multi-step flow policy (Step~2) to solve per-state optimal transport problems (Step~3). By decoupling value guidance (target marginals) from behavioral geometry (cost matrix), the final transport-guided distillation (Step 4) explicitly maps noise to distinct, in-distribution high-value action modes.
Notably, OptiFlow avoids direct critic maximization, which often exploits unreliable out-of-distribution regions. Instead, it anchors the policy to high-return in-distribution targets via transport plan. This structural separation mitigates the need for explicit behavior regularization while ensuring stable and mode-preserving one-step policy extraction.



\begin{algorithm}[t]
\caption{OptiFlow training}
\label{alg:OptiFlow}
\small
\begin{algorithmic}[1]
\FOR{each training iteration}
    \STATE Sample a mini-batch $\mathcal B=\{(s,a,r,s')\}$ from $\mathcal D$
    \STATE Update critic $Q_\phi$ using $\nabla_\phi \mathcal L_Q(\phi)$ \quad({\footnotesize Eq.~\eqref{eq:q_loss}})
    \STATE Update reference flow policy $\mu_\omega$ using $\nabla_\omega \mathcal L_{\mathrm{ref}}(\omega)$\quad({\footnotesize Eq.~\eqref{eq:ref_loss}})
    \FOR{each state $s \in \mathcal B$}
        \STATE Sample one-step actions $\{a_{\theta,i}^s\}_{i=1}^N$ from $\mu_\theta(s,\cdot)$
        \STATE Sample reference actions $\{\tilde a_{\omega,j}^s\}_{j=1}^M$ from $\mu_\omega(s,\cdot)$
        \STATE Set $p_i=\tfrac{1}{N}$ and compute the value-weighted reference-policy marginal $q_j^s$ by Eq.~\eqref{eq:qjs}
        \STATE Construct $C_{ij}^s= \tfrac{1}{\bar c^s} \|a_{\theta,i}^s-\tilde a_{\omega,j}^s\|_2^2$
        \STATE Compute $P^{s,*}$ by Sinkhorn on $(C^s,p,q,\varepsilon,T)$\quad({\footnotesize Alg.~\ref{alg:sinkhorn}})
        \STATE Set $j_i^*=\arg\max_j P^{s,*}_{ij}$
    \ENDFOR
    \STATE Update one-step policy $\mu_\theta$ using $\nabla_\theta \mathcal L_{\mathrm{distill}}(\theta)$\quad({\footnotesize Eq.~\eqref{eq:distill_loss}})
    \STATE Update target critic $Q_{\bar\phi}$
\ENDFOR
\end{algorithmic}
\end{algorithm}

\section{Experiments}

Our experiments are designed to answer three questions:
(i) whether OptiFlow preserves multimodal action structure in controlled settings,
(ii) which design choices are responsible for the gains, including the use of a value-weighted reference-policy marginal and an OT coupling, and
(iii) whether value-weighted, transport-guided distillation improves offline RL performance over conventional and flow-based baselines.

\textcolor{black}{\subsection{Controlled Multimodal Experiments}
}
\paragraph{\textcolor{black}{Bandit diagnostic.}}

\begin{figure}[H]
\centering
\captionsetup[subfigure]{font=scriptsize}
\newcommand{\panelw}{0.14\linewidth}

\begin{subfigure}[t]{\panelw}
    \centering
    \includegraphics[width=\linewidth]{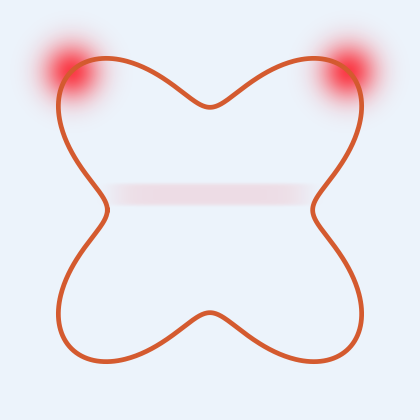}
    \caption{True Reward}
\end{subfigure}\hspace{0.02\linewidth}
\begin{subfigure}[t]{\panelw}
    \centering
    \includegraphics[width=\linewidth]{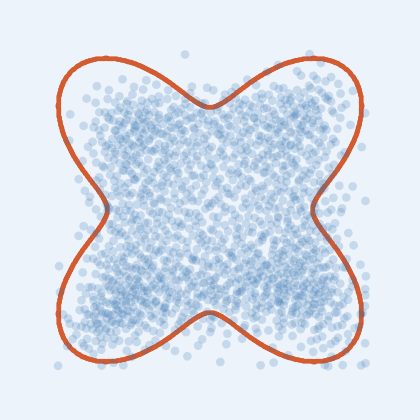}
    \caption{BC}
\end{subfigure}\hspace{0.02\linewidth}
\begin{subfigure}[t]{\panelw}
    \centering
    \includegraphics[width=\linewidth]{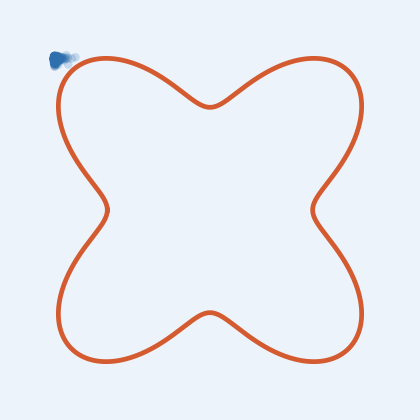}
    \caption{ReBRAC\\ w/ Gaussian Policy}
\end{subfigure}\hspace{0.02\linewidth}
\begin{subfigure}[t]{\panelw}
    \centering
    \includegraphics[width=\linewidth]{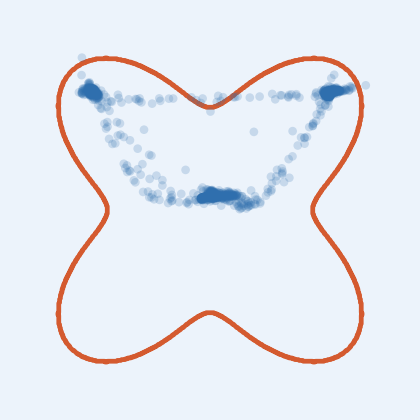}
    \caption{FQL-D}
\end{subfigure}\hspace{0.02\linewidth}
\begin{subfigure}[t]{\panelw}
    \centering
    \includegraphics[width=\linewidth]{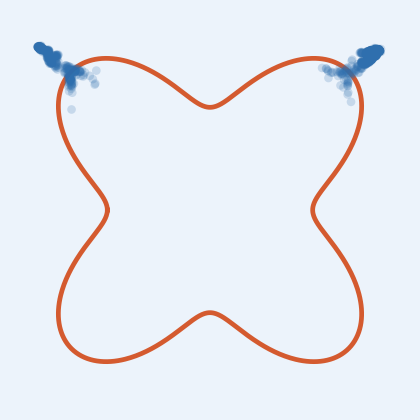}
    \caption{FQL-Q}
\end{subfigure}
\begin{subfigure}[t]{\panelw}
    \centering
    \includegraphics[width=\linewidth]{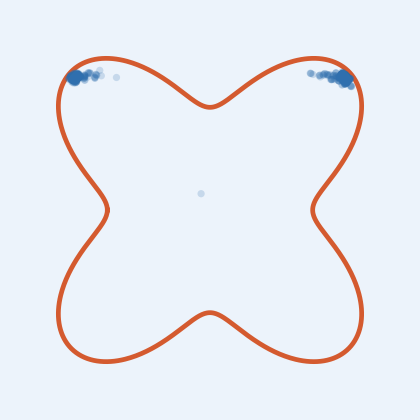}
    \caption{\textbf{OptiFlow (ours)}}
\end{subfigure}

\caption{\textbf{Bandit multimodal diagnostic.} A bimodal Reward landscape illustrating two high-value modes and a suboptimal region. The figure evaluates whether different methods preserve multimodality, avoid suboptimal assignments, and remain within behavior support. The experimental setup is detailed in Appendix~\ref{app:bandit_diagnostic_setup}}
\label{fig:bandit_toy}
\end{figure}

Figure~\ref{fig:bandit_toy} illustrates three requirements for offline policy distillation: preserving multimodality, avoiding suboptimal target assignments, and remaining within behavior support. Experimental results of FQL~\citep{park2025flow} highlight the trade-off between behavior matching and value maximization: emphasizing the dataset regularization term~(FQL-D in Figure~\ref{fig:bandit_toy}d) biases the policy toward the data distribution but under-prioritizes high-value modes, whereas emphasizing the critic maximization term~(FQL-Q in Figure~\ref{fig:bandit_toy}e) can deviate from behavior support due to out-of-distribution value overestimation. Gaussian-based policies~(Figure~\ref{fig:bandit_toy}c) with direct critic maximization can similarly drift outside behavior support under value overestimation and fail to capture multimodality.

In contrast, OptiFlow uses a mechanism in which the critic shapes the reference-policy marginal, while the OT coupling determines pairings according to the action-distance cost. This combination enables structured supervision that preserves multimodality~(Figure~\ref{fig:bandit_toy}f). Even when suboptimal actions appear among the reference samples, they are rarely selected after the row-wise maximum-mass assignment. Finally, because the targets are drawn from the behavior-regularized flow reference policy, OptiFlow remains grounded in dataset support while preserving multiple high-value modes.

\paragraph{Why OT coupling and value-weighted marginals?}

\begin{figure}[H]
\centering
\captionsetup[subfigure]{font=scriptsize}
\newcommand{\panelw}{0.14\linewidth}

\begin{subfigure}[t]{\panelw}
    \centering
    \includegraphics[width=\linewidth]{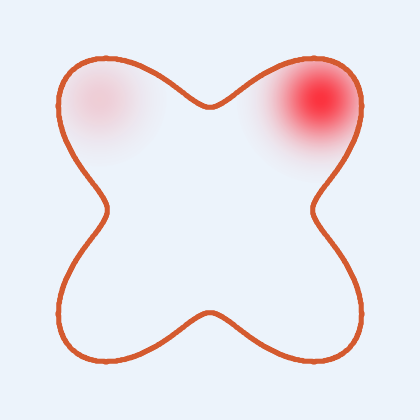}
    \caption{Asym. Q}
\end{subfigure}\hfill
\begin{subfigure}[t]{\panelw}
    \centering
    \includegraphics[width=\linewidth]{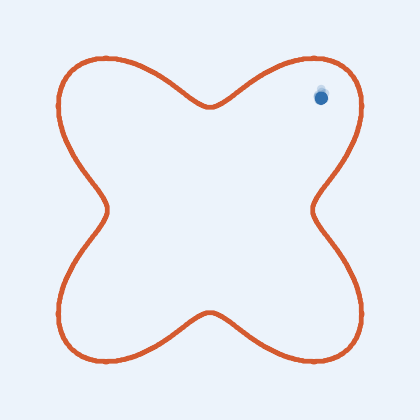}
    \caption{VWD}
\end{subfigure}\hfill
\begin{subfigure}[t]{\panelw}
    \centering
    \includegraphics[width=\linewidth]{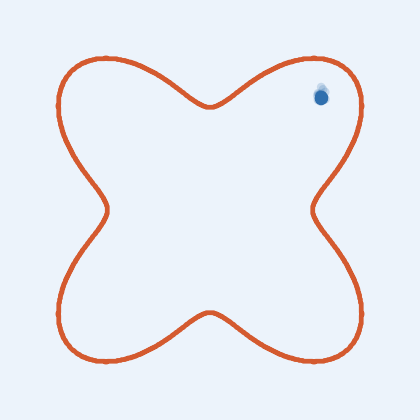}
    \caption{Top1-NN}
\end{subfigure}\hfill
\begin{subfigure}[t]{\panelw}
    \centering
    \includegraphics[width=\linewidth]{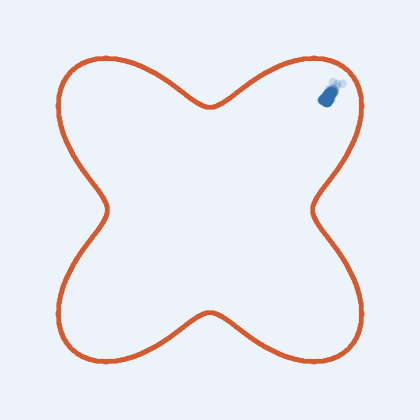}
    \caption{Top3-NN}
\end{subfigure}\hfill
\begin{subfigure}[t]{\panelw}
    \centering
    \includegraphics[width=\linewidth]{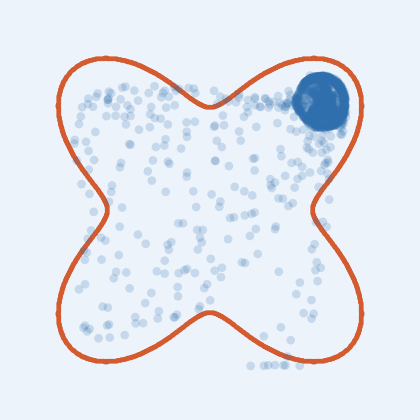}
    \caption{Q-cost OT}
\end{subfigure}\hfill
\begin{subfigure}[t]{\panelw}
    \centering
    \includegraphics[width=\linewidth]{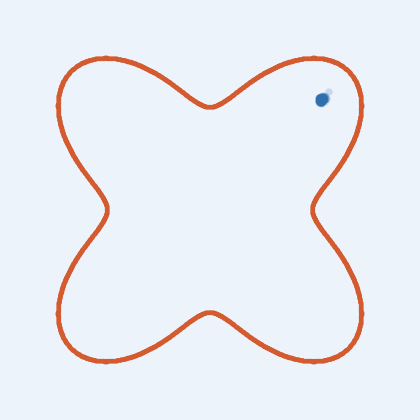}
    \caption{\textbf{OptiFlow (ours)}}
\end{subfigure}\hfill

\vspace{1mm}

{\color{gray}\rule{1\linewidth}{0.3pt}}

\vspace{1mm}

\begin{subfigure}[t]{\panelw}
    \centering
    \includegraphics[width=\linewidth]{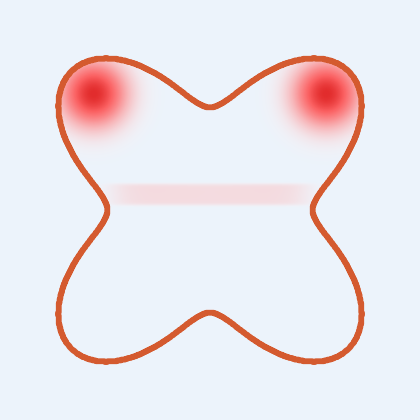}
    \caption{Bimodal Q}
\end{subfigure}\hfill
\begin{subfigure}[t]{\panelw}
    \centering
    \includegraphics[width=\linewidth]{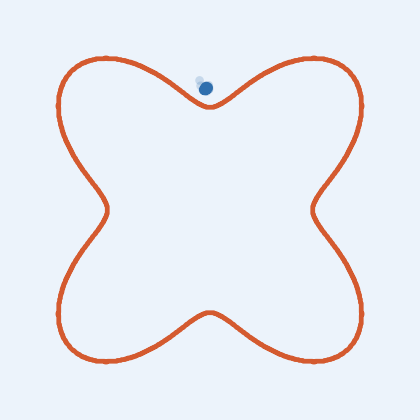}
    \caption{VWD}
\end{subfigure}\hfill
\begin{subfigure}[t]{\panelw}
    \centering
    \includegraphics[width=\linewidth]{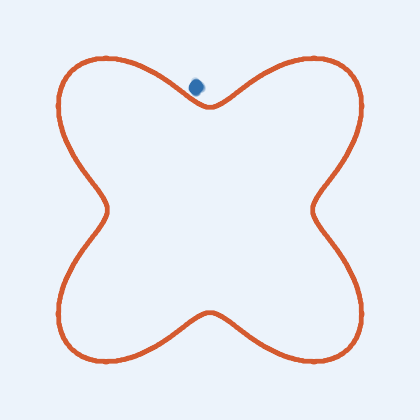}
    \caption{Top1-NN}
\end{subfigure}\hfill
\begin{subfigure}[t]{\panelw}
    \centering
    \includegraphics[width=\linewidth]{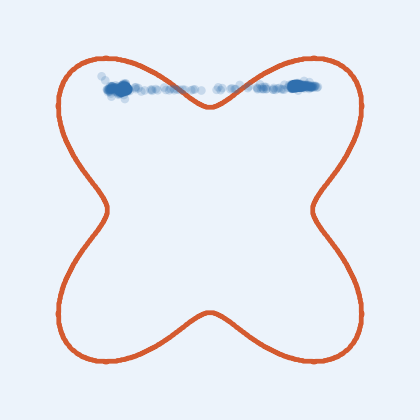}
    \caption{Top3-NN}
\end{subfigure}\hfill
\begin{subfigure}[t]{\panelw}
    \centering
    \includegraphics[width=\linewidth]{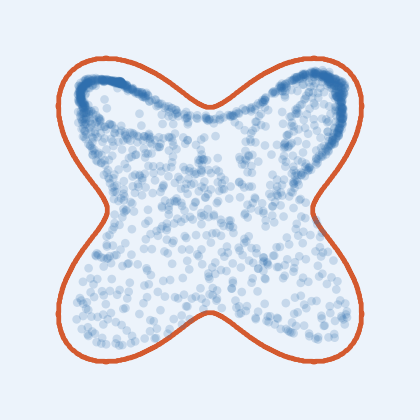}
    \caption{Q-cost OT}
\end{subfigure}\hfill
\begin{subfigure}[t]{\panelw}
    \centering
    \includegraphics[width=\linewidth]{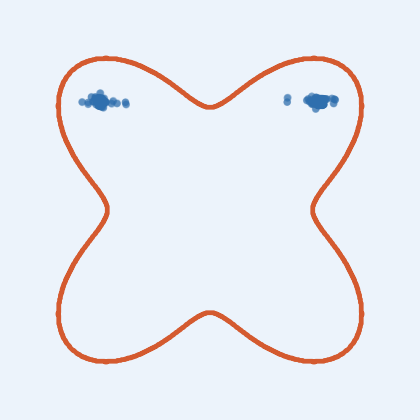}
    \caption{\textbf{OptiFlow (ours)}}
\end{subfigure}

\caption{
\textbf{Bandit diagnostic for value-aware transport.}
(a,g) show asymmetric and bimodal value landscapes. The figure evaluates whether different distillation strategies can capture high-value regions while preserving the distribution over optimal modes.
}
\label{fig:bandit_toy2}
\end{figure}

Figure~\ref{fig:bandit_toy2} compares several alternatives for offline policy distillation under asymmetric and bimodal value landscapes.
Value-weighted distillation (VWD) guides the one-step policy toward high-value targets, but collapses toward a single dominant mode~(Figure~\ref{fig:bandit_toy2}b,h). Top-\(K\) nearest-neighbor distillation partially preserves multimodality by selecting multiple reference actions, but still struggles to fully capture the multimodal structure~(Figure~\ref{fig:bandit_toy2}c,d,i,j). These results suggest that local or row-wise matching alone is insufficient for preserving the distribution over optimal modes.

A more structured alternative is to introduce OT coupling between one-step-policy and reference-policy samples. However, not all OT formulations behave similarly. One natural alternative approach of our method is to keep the reference-policy marginal uniform while injecting value directly into the transport cost, $C_{ij}=f(d_{ij}, Q(s,a_{\theta,i}^s), Q(s,\tilde a_{\omega,j}^s)).$ Although this formulation trades off value and distance, Figure~\ref{fig:bandit_toy2}e,k shows that it still misallocates samples because the uniform reference marginal forces every reference action, including low-value ones, to receive equal total mass.

OptiFlow instead places value into the \emph{reference-policy marginal} itself: high-value reference actions receive more supervision mass, while OT coupling determines geometrically compatible assignments between one-step and reference-policy samples. As shown in Figure~\ref{fig:bandit_toy2}f,l, this asymmetric marginal design preserves multimodal structure while concentrating supervision on high-value, dataset-supported regions.

\textcolor{black}{We additionally provide a counterexample illustrating a structural limitation of uniform-marginal OT, generalize this limitation beyond the counterexample, and empirically demonstrate the importance of the reference marginal on real-world benchmarks in Appendix~\ref{app:uniform_marginal}.}
We describe the value-weighted, Top-\(K\) nearest-neighbor distillation formulations  and more general value-, distance-aware distillation formulation in Appendix~\ref{app:swd-tkd}, and discuss their failure modes on mode collapse to motivate the need for OptiFlow formulation.

\begin{figure}[H]
\centering
\captionsetup[subfigure]{font=scriptsize}
\newcommand{\panelw}{0.2\linewidth}

\begin{subfigure}[t]{\panelw}
    \centering
    \includegraphics[width=\linewidth]{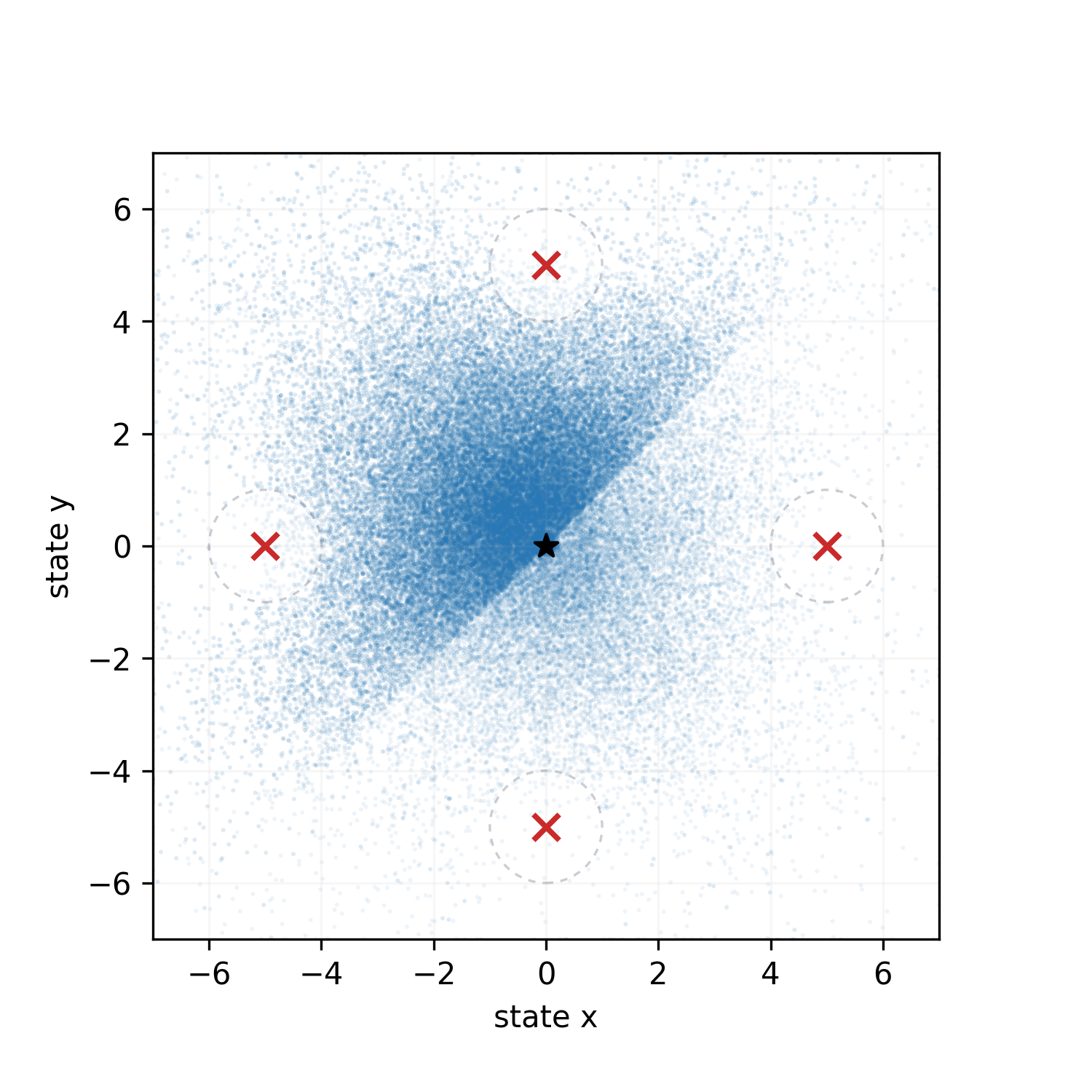}
    \caption{Dataset Distribution}
\end{subfigure}\hfill
\begin{subfigure}[t]{\panelw}
    \centering
    \includegraphics[width=\linewidth]{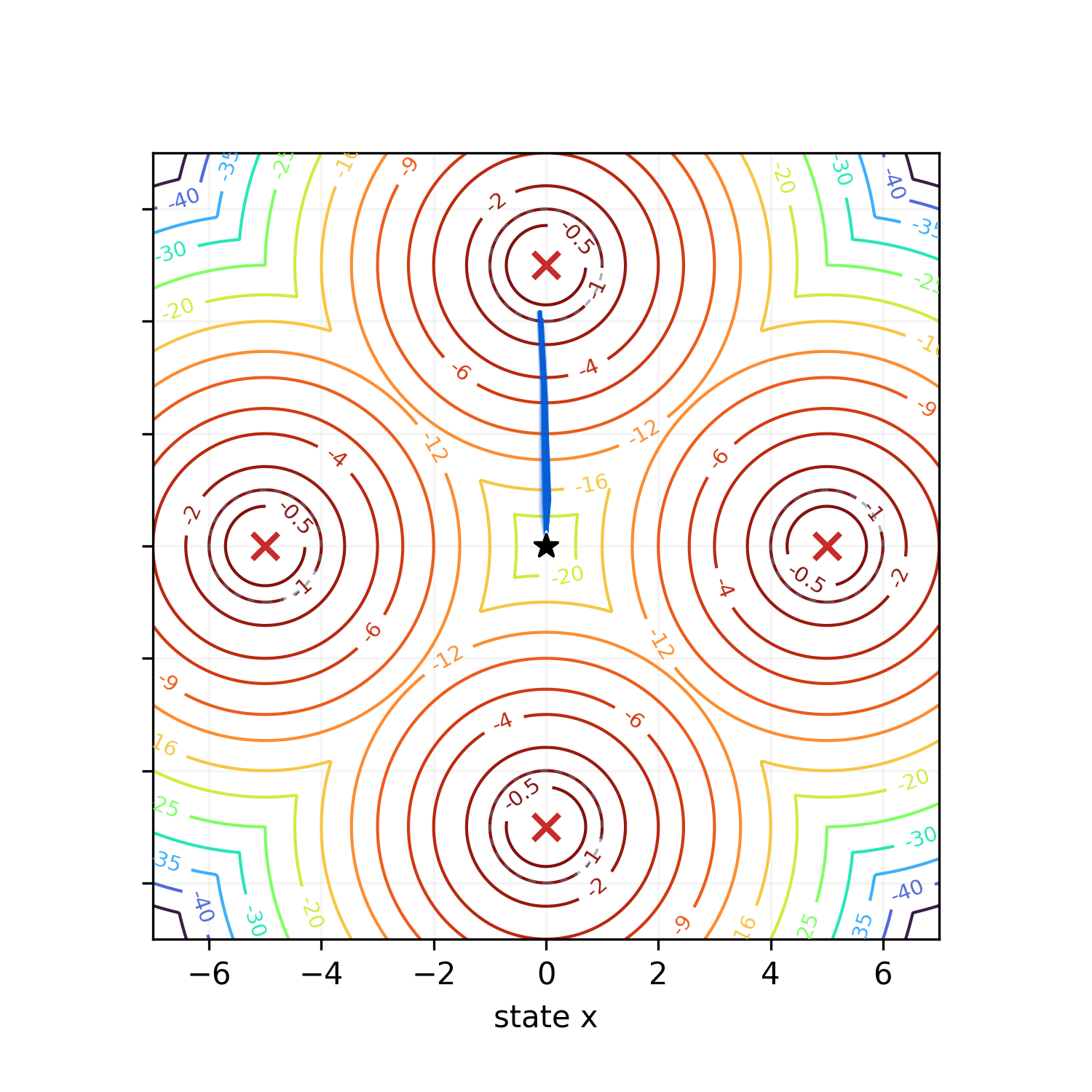}
    \caption{FQL ($\alpha=0.1$)}
\end{subfigure}\hfill
\begin{subfigure}[t]{\panelw}
    \centering
    \includegraphics[width=\linewidth]{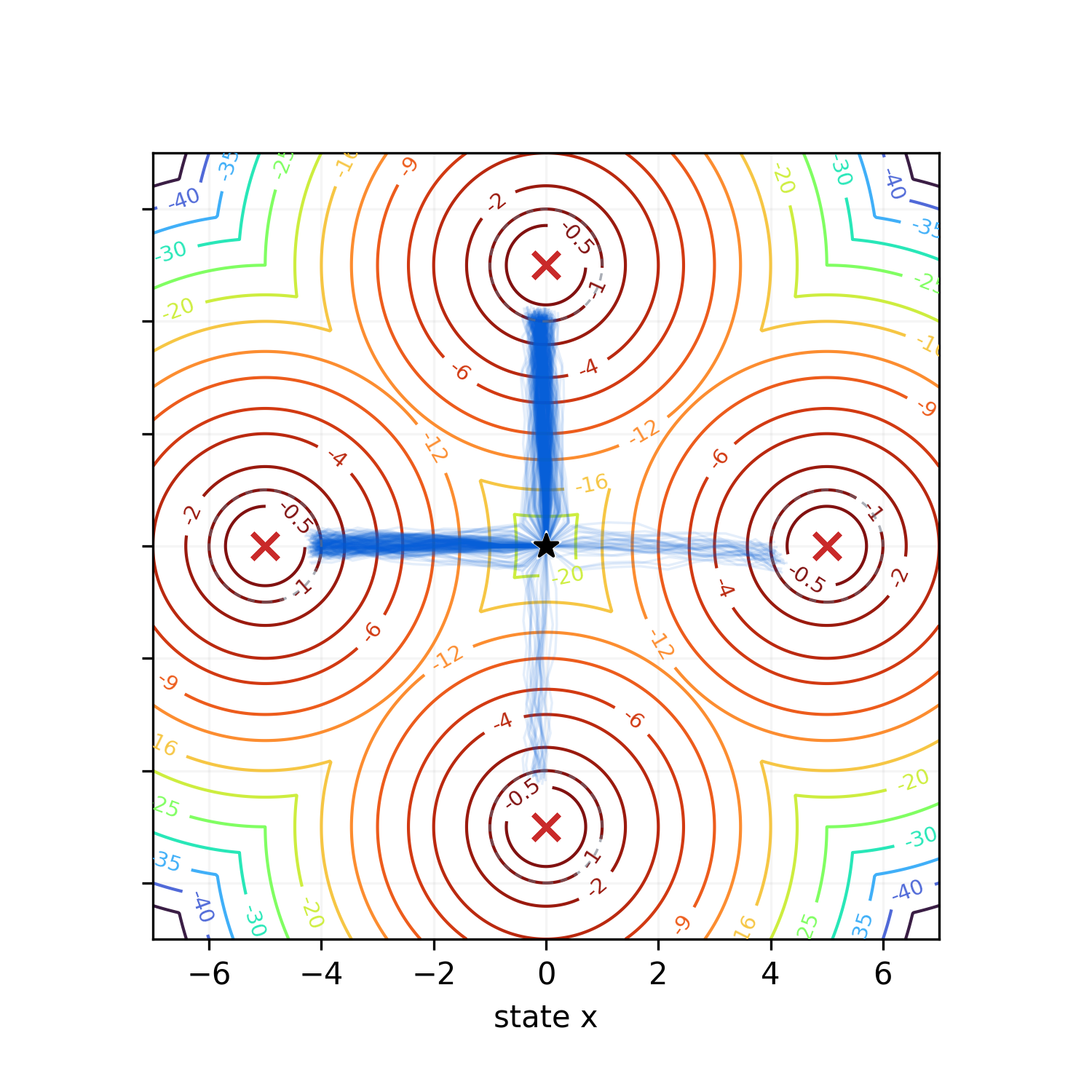}
    \caption{FQL ($\alpha=1$)}
\end{subfigure}\hfill
\begin{subfigure}[t]{\panelw}
    \centering
    \includegraphics[width=\linewidth]{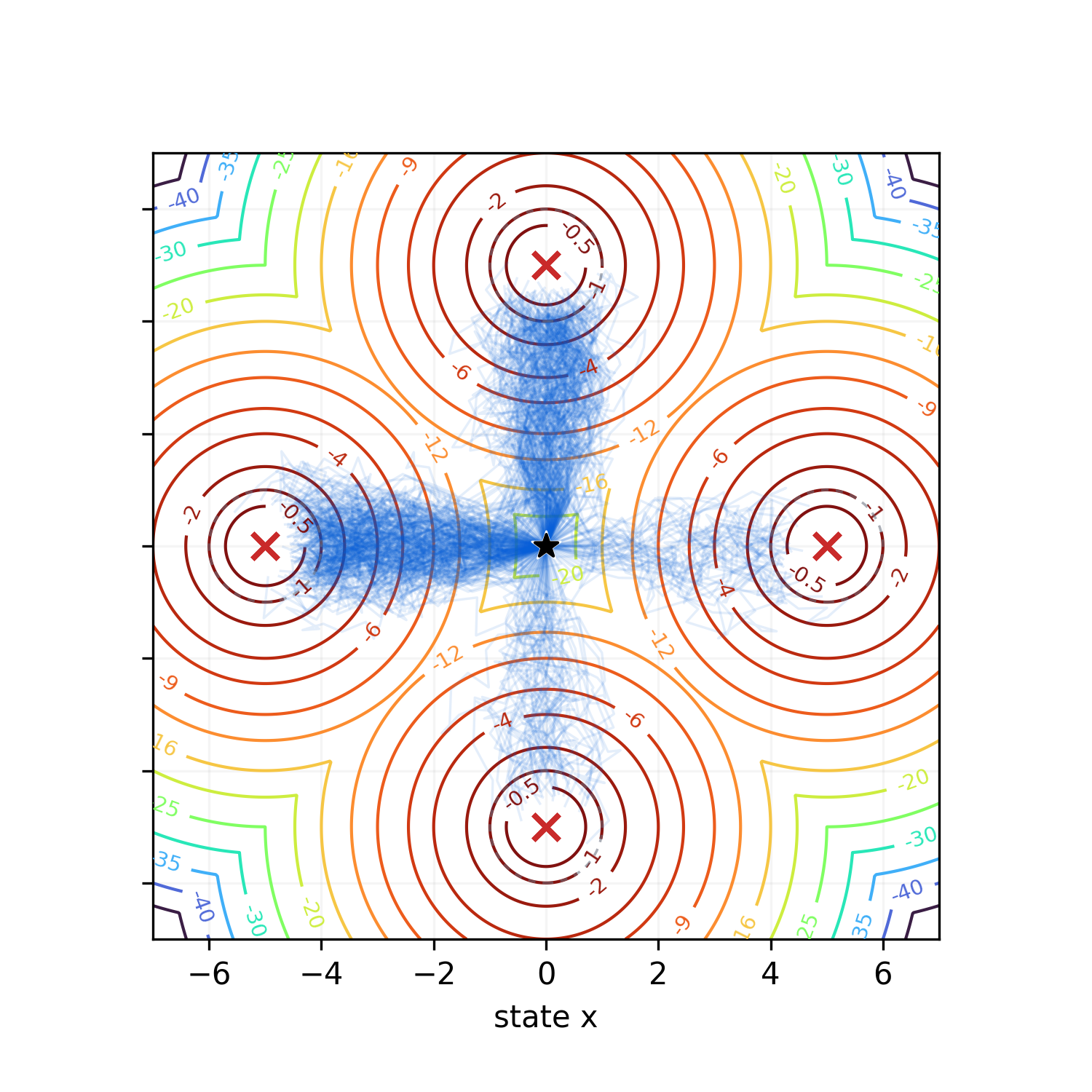}
    \caption{FQL ($\alpha=10$)}
\end{subfigure}\hfill
\begin{subfigure}[t]{\panelw}
    \centering
    \includegraphics[width=\linewidth]{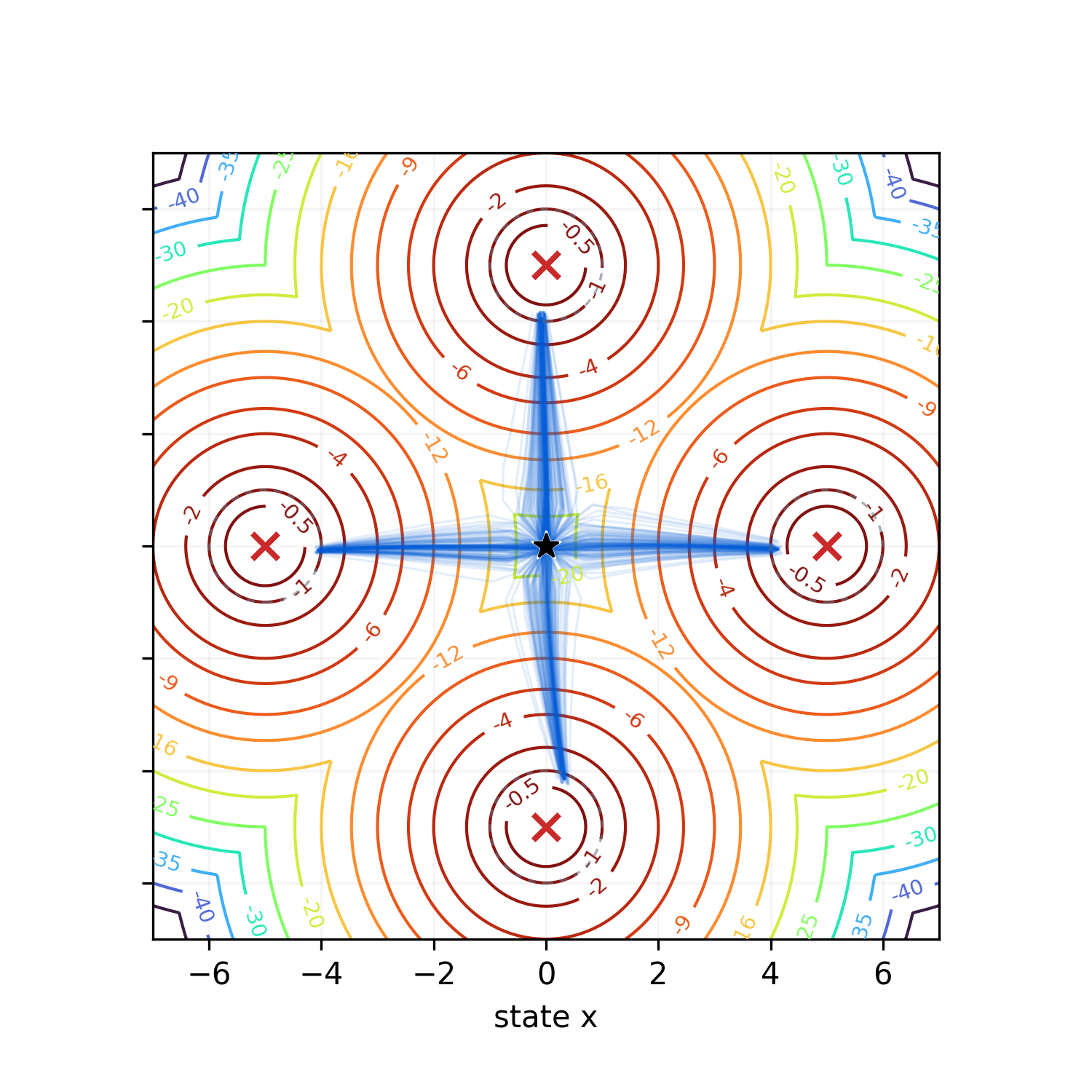}
    \caption{\textbf{OptiFlow-BC}}
\end{subfigure}

\caption{
\textbf{Four-mode environment for evaluating multimodal behavior.}
(a) Offline dataset distribution with intentionally uneven coverage across the four goal directions. (b–e) Learned policy distributions overlaid on the reward landscape. (b–d) FQL with increasing distillation strength \(\alpha\), exhibiting a diversity–quality trade-off. (e) OptiFlow-BC, maintaining balanced coverage across the four equally optimal modes.
}
\label{fig:mode_coverage_env}
\end{figure}

\paragraph{\textcolor{black}{Beyond the bandit setting.}}
\textcolor{black}{
We further examine the preservation of multimodal behavior in a four-goal MDP introduced by~\citep{haarnoja2017reinforcement}, as illustrated in Figure~\ref{fig:mode_coverage_env}, where the four goals constitute equally optimal modes but are represented unevenly in the offline dataset (Figure~\ref{fig:mode_coverage_env}a). This setting tests whether the one-step policy extraction preserves multiple high-value modes despite imbalanced data coverage. To isolate the effect of transport-guided distillation from value-aware reference-policy training, we use OptiFlow-BC, a controlled variant of OptiFlow that replaces the VaBC reference policy with a standard BC reference while retaining the same value-weighted OT coupling and transport-guided distillation. OptiFlow-BC (Figure~\ref{fig:mode_coverage_env}e) achieves balanced coverage of the four optimal modes while attaining the highest return, {demonstrating that the policy extraction mechanism can preserve multimodal behavior despite imbalanced mode frequencies in the offline dataset.  In contrast, FQL exhibits a clear diversity–quality trade-off (Figure~\ref{fig:mode_coverage_env}b–d): stronger distillation improves multimodal coverage but progressively degrades policy return.}
}

\textcolor{black}{
We also observe that the mode-level transport assignments remain consistent across repeated resampling of the reference actions, suggesting that the stochastic candidate set does not induce frequent assignment switches across distinct modes. The environment and reference-resampling protocols for assignment stability experiment are detailed in Appendix~\ref{app:4way_diagnostic_setting}.
}


\subsection{Benchmark Experimental Setup}

\paragraph{Benchmarks.}
We evaluate OptiFlow on reward-based single-task variants of OGBench~\citep{park2025ogbench} and on D4RL AntMaze and Adroit~\citep{fu2020d4rl}. OGBench includes navigation and manipulation tasks with complex and often multimodal action distributions, while D4RL provides sparse-reward navigation and dexterous manipulation benchmarks. Full per-task results are provided in Appendix~\ref{app:results}.

\begin{table}[t]
\centering
\caption{\textbf{Aggregate performance across OGBench families and D4RL domains.}
For state-based OGBench, each family row reports the mean $\pm$ standard error over the five single-task variants in that family. Visual OGBench rows report the corresponding visual single-task environments. D4RL rows aggregate over all tasks in the corresponding domain. Values within 95\% of the best score in each row are in \textbf{bold}. Dashes denote unreported entries. Italicized entries indicate results taken from prior work~\citep{park2025flow}, with complete per-task result tables in Appendix~\ref{app:results}.}
\label{tab:main_summary}
\scriptsize
\setlength{\tabcolsep}{3pt}
\renewcommand{\arraystretch}{1.2}
\begin{tabular}{@{}>{\raggedright\arraybackslash}p{3.0cm}*{7}{c}!{\hskip 3pt\vrule\hskip 1.5pt\vrule\hskip 3pt}c@{}}
\toprule
\textbf{Family / Domain} & BC & IQL & ReBRAC & IDQL & IFQL & FQL & GFP & \textbf{\textcolor[rgb]{0.05,0.35,0.75}{OptiFlow}} \\
\midrule
\multicolumn{9}{@{}l}{\textbf{OGBench --- Navigation}} \\
\midrule
AntMaze-Large \{1--5\}      & \val{1.7}{0.2} & \val{53.0}{2.3} & \val{86.0}{1.5} & \val{26.8}{1.4} & \val{28.0}{2.5} & \val{79.6}{1.0} & \val{\textbf{93.9}}{0.5} & \val{\textbf{91.9}}{0.5} \\
AntMaze-Giant \{1--5\}      & \val{0.0}{0.0} & \val{4.3}{0.8}  & \val{\textbf{32.4}}{3.3} & \val{0.0}{0.0} & \val{0.9}{0.4} & \val{9.5}{2.3} & \val{18.9}{2.5} & \val{26.5}{1.5} \\
Humanoid-Medium \{1--5\}    & \val{1.4}{0.2} & \val{33.8}{1.4} & \val{19.1}{2.4} & \val{3.0}{0.3} & \val{58.0}{4.2} & \val{57.4}{1.6} & \val{\textbf{71.8}}{1.0} & \val{\textbf{74.4}}{1.2} \\
Humanoid-Large \{1--5\}     & \val{0.0}{0.0} & \val{2.3}{0.3} & \val{2.7}{0.5} & \val{0.4}{0.1} & \val{12.7}{0.9} & \val{3.6}{0.7} & \val{11.0}{1.8} & \val{\textbf{19.1}}{1.1} \\
AntSoccer-Arena \{1--5\}    & \val{0.2}{0.1} & \val{9.4}{0.6} & \val{0.1}{0.1} & \val{3.5}{0.5} & \val{29.1}{3.1} & \val{\textbf{61.4}}{1.5} & \val{\textbf{61.1}}{1.3} & \val{52.7}{1.6} \\
\textit{Navigation Avg.}    & \val{0.7}{0.1} & \val{20.6}{0.6} & \val{28.1}{0.9} & \val{6.7}{0.3} & \val{25.8}{1.2} & \val{42.3}{0.7} & \val{\textbf{51.3}}{0.7} & \val{\textbf{52.9}}{0.5} \\
\midrule
\multicolumn{9}{@{}l}{\textbf{OGBench --- Manipulation}} \\
\midrule
Cube-Single \{1--5\}        & \val{2.8}{1.0} & \val{50.1}{2.5} & \val{87.8}{1.2} & \val{91.9}{0.5} & \val{83.9}{0.8} & \val{91.0}{1.1} & \val{\textbf{96.3}}{0.5} & \val{\textbf{97.6}}{0.3} \\
Cube-Double \{1--5\}        & \val{0.0}{0.0} & \val{2.4}{0.2} & \val{8.1}{1.3} & \val{2.8}{0.7} & \val{10.5}{0.7} & \val{25.2}{1.5} & \val{43.3}{1.6} & \val{\textbf{46.1}}{1.3} \\
Cube-Triple \{1--5\}        & \val{0.0}{0.0} & \val{0.5}{0.2} & \val{2.2}{0.4} & \val{0.4}{0.2} & \val{0.2}{0.1} & \val{1.1}{0.4} & \val{1.8}{0.3} & \val{\textbf{3.6}}{0.2} \\
Puzzle-3$\times$3 \{1--5\}  & \val{3.9}{0.6} & \val{4.3}{0.5} & \val{22.3}{0.5} & \val{14.0}{0.7} & \val{19.3}{0.4} & \val{\textbf{29.5}}{0.8} & \val{24.4}{1.1} & \val{\textbf{28.1}}{2.3} \\
Puzzle-4$\times$4 \{1--5\}  & \val{0.0}{0.0} & \val{1.7}{0.2} & \val{14.0}{0.5} & \val{22.1}{1.2} & \val{26.8}{1.1} & \val{15.0}{0.8} & \val{25.8}{1.1} & \val{\textbf{32.1}}{1.5} \\
Scene-Play \{1--5\}         & \val{0.4}{0.1} & \val{20.0}{0.8} & \val{42.0}{1.5} & \val{39.1}{0.3} & \val{44.7}{1.7} & \val{55.8}{1.4} & \val{53.1}{1.6} & \val{\textbf{59.4}}{0.3} \\
\textit{Manipulation Avg.}  & \val{1.2}{0.2} & \val{13.2}{0.5} & \val{29.4}{0.4} & \val{28.4}{0.3} & \val{30.9}{0.4} & \val{36.3}{0.4} & \val{40.8}{0.5} & \val{\textbf{44.5}}{0.5} \\
\midrule
\multicolumn{9}{@{}l}{\textbf{OGBench --- Visual}} \\
\midrule
Visual-Cube-Single          & -- & \itres{\val{70}{6}} & \itres{\val{83}{3}} & -- & \itres{\val{49}{3.5}} & \itres{\val{81}{6}} & \val{72.2}{3.6} & \val{\textbf{86.6}}{1.3} \\
Visual-Cube-Double          & -- & \itres{\val{34}{11.5}} & \itres{\val{4}{2}} & -- & \itres{\val{8}{3}} & \itres{\val{21}{5.5}} & \val{12.0}{2.6} & \val{\textbf{66.7}}{3.5} \\
Visual-Scene-Play           & -- & \textit{\val{\textbf{96}}{1}} & \textit{\val{\textbf{98}}{2}} & -- & \itres{\val{86}{5}} & \textit{\val{\textbf{98}}{1.5}} & \val{\textbf{98.7}}{0.3} & \val{92.8}{0.8} \\
Visual-Puzzle-3$\times$3    & -- & \itres{\val{7}{7.5}} & \itres{\val{88}{2}} & -- & \textit{\val{\textbf{100}}{0}} & \itres{\val{94}{0.5}} & \val{86.5}{1.1} & \val{93.8}{1.2} \\
Visual-Puzzle-4$\times$4    & -- & \itres{\val{0}{0}} & \itres{\val{26}{3}} & -- & \itres{\val{8}{7.5}} & \itres{\val{33}{3}} & \val{19.3}{1.4} & \val{\textbf{39.7}}{4.8} \\
\midrule\midrule
\multicolumn{9}{@{}l}{\textbf{Overall}} \\
\midrule
OGBench Visual Avg.         & -- & \itres{\val{41.4}{3.0}} & \itres{\val{59.8}{1.1}} & -- & \itres{\val{50.2}{2.0}} & \itres{\val{65.4}{1.8}} & \val{57.7}{1.0} & \val{\textbf{75.9}}{1.3} \\
OGBench State-Based Avg.    & \val{0.9}{0.1} & \val{16.5}{0.4} & \val{28.8}{0.5} & \val{18.5}{0.2} & \val{28.6}{0.6} & \val{39.0}{0.4} & \val{45.6}{0.4} & \val{\textbf{48.3}}{0.4} \\
AntMaze (D4RL) Avg.         & \itres{16.7} & \itres{58.4} & \itres{76.9} & \itres{79.1} & \val{61.0}{2.5} & \val{\textbf{82.3}}{1.3} & \val{\textbf{82.6}}{1.5} & \val{\textbf{86.6}}{0.7} \\
Adroit (D4RL) Avg.          & \itres{36.7} & \itres{53.5} & \textit{\textbf{58.6}} & \val{50.2}{0.4} & \val{48.2}{0.5} & \val{50.7}{0.5} & \val{52.1}{0.5} & \val{50.5}{0.4} \\
\bottomrule
\end{tabular}
\end{table}

\paragraph{Baselines.}
We compare against both conventional offline RL baselines and expressive generative-policy baselines. \textcolor{black}{BC, IQL~\citep{kostrikov2022offline}, ReBRAC~\citep{tarasov2023revisiting}, serve as standard offline RL references, while IDQL~\citep{hansenestruch2023idql}, IFQL, FQL~\citep{park2025flow} and GFP~\citep{tiofack2026guided} represent the most closely related baselines based on expressive generative policies.}

\paragraph{Training and evaluation.}
For state-based OGBench tasks, all methods are trained for 1M gradient steps. For D4RL and OGBench visual tasks, all methods are trained for 500K gradient steps. We report final-checkpoint performance averaged over 8 random seeds unless otherwise noted. For visual tasks, due to high computational cost, we report performance averaged over 4 random seeds and over checkpoints at 300K, 400K, and 500K steps, following the evaluation protocol of FQL~\citep{park2025flow}. Full training and evaluation details are provided in Appendix~\ref{app:impl} and the result-table captions.

\paragraph{Method details.}
OptiFlow uses a value-aware reference flow policy and a one-step flow policy. For each state, we sample $N=16$ one-step-policy actions and $M=64$ reference-policy actions, construct a value-weighted reference-policy marginal, compute an entropic Optimal Transport plan with 30 Sinkhorn iterations, \textcolor{black}{select a reference action for each one-step-policy sample via the maximum transport assignment in the corresponding row, and distill the resulting assignments into the one-step policy.} Additional implementation details and full hyperparameters are given in Appendix~\ref{app:impl}.

\subsection{Main Results}

\paragraph{Aggregate benchmark performance.}
Table~\ref{tab:main_summary} summarizes performance across OGBench families and D4RL domains. OptiFlow achieves the best overall state-based OGBench average, with $48.3\pm0.4$ compared to $45.6\pm0.4$ for GFP and $39.0\pm0.4$ for FQL. The gains are most pronounced on OGBench manipulation, where OptiFlow obtains the strongest aggregate performance, improving the manipulation average to $44.5\pm0.5$ compared to $40.8\pm0.5$ for GFP and $36.3\pm0.4$ for FQL. These results show that value-weighted transport-guided distillation improves over the closest flow-based baselines in the setting most aligned with OptiFlow's motivation: learning efficient one-step policies from complex and multimodal action distributions.

\paragraph{OGBench manipulation and navigation.}
OptiFlow obtains the best aggregate score on five of the six OGBench manipulation families: Cube-Single, Cube-Double, Cube-Triple, Puzzle-4$\times$4, and Scene-Play. On OGBench navigation, OptiFlow remains competitive: it improves over FQL on AntMaze-Giant, Humanoid-Medium, and Humanoid-Large. Overall, OptiFlow's strongest gains appear in tasks where multimodal reference-policy structure is useful for one-step policy learning.

\paragraph{D4RL results.}
On D4RL, OptiFlow achieves the best AntMaze average, improving over GFP by $4.0$ points and FQL by $4.3$ points. On Adroit, OptiFlow is competitive with flow-based baselines. These results suggest that OptiFlow transfers beyond OGBench, with its clearest advantage in settings where useful multimodal reference actions can be transferred to a one-step policy through structured assignment.

\paragraph{Sensitivity analysis.}
We provide additional sensitivity analyses of OptiFlow with respect to the value-weighting temperatures \(\tau\) and \(\eta\) (Appendix~\ref{app:sensitivity}), the entropic regularization coefficient and the number of Sinkhorn iterations (Appendix~\ref{app:sinkhorn_param_abal}), and the numbers of one-step and reference-policy action samples (Appendix~\ref{app:sampling_abal}).


\paragraph{Additional ablations and diagnostics.}
We first control for OptiFlow’s use of multiple action samples per state and show that increased sampling alone does not explain its gains (Appendix~\ref{app:matched_sampling}). We then disentangle the contributions of the reference policy and transport-guided policy extraction (Appendix~\ref{app:disentangle_vabc_OT}) and compare our default VaBC weighting with an AWR-style alternative (Appendix~\ref{app:awr}). Finally, we examine whether direct critic maximization provides benefits beyond transport-based supervision (Appendix~\ref{app:direct_q_max}) and demonstrate \textbf{strong offline-to-online fine-tuning performance} (Appendix~\ref{app:online}).

\section{Conclusion}

We presented \textbf{One-step Flow Policy via Optimal Transport (OptiFlow)}, a value-weighted optimal-transport method to learn efficient one-step flow policies for offline reinforcement learning. OptiFlow uses a value-aware reference flow policy to represent multimodal, data-supported behavior, and trains a one-step flow policy through transport-guided distillation. By constructing an entropic OT coupling between one-step-policy and reference-policy samples, OptiFlow separates value and geometry: critic-estimated values determine how much supervision mass each reference action receives, while action-space distance determines how one-step samples are assigned. This provides a structured alternative to independent noise-wise distillation and direct critic maximization.

Experimental results show that OptiFlow preserves multimodal action structure in controlled diagnostics and achieves strong performance across offline RL benchmarks, with especially clear gains on OGBench manipulation tasks. These results support the central claim that value-guided one-step policy learning benefits from transport-based distillation.

More broadly, OptiFlow suggests that critic information in offline RL need not be used only through direct critic maximization. It can instead shape structured supervision for policy learning, opening a path toward value-guided generative policy learning methods that preserve multimodal behavior while retaining efficient one-step policy deployment.

OptiFlow also has several limitations. Compared to standard offline RL methods, especially existing flow-based policies, it introduces additional computational overhead from repeated policy sampling and Sinkhorn-based optimal transport updates. However, these operations mainly consist of matrix-based computations that can be efficiently parallelized on modern GPUs (see Appendix~\ref{app:compute} for detailed analysis of computational overhead of OptiFlow).

\newpage
\begin{ack}
This work was supported by
the Institute of Information \& communications Technology Planning \& Evaluation (IITP) grants funded by the Korea government (MSIT) (No. RS-2020-II201361, Artificial Intelligence Graduate School Program (Yonsei University), RS-2024-00457882, AI Research Hub Project)
and the National Research Foundation of Korea (NRF) grants funded by the Korea government(MSIT) (RS-2026-25505230, RS-2026-25619452).
This research was also supported by a grant from the Institute for AI and Social Innovation at Yonsei University and the Yonsei University Research Fund of 2026-22-0204.

\end{ack}

\bibliographystyle{unsrtnat}
\bibliography{refs}

\newpage
\appendix
\startcontents[appendix]

\section*{Appendix Contents}
\begingroup
\small
\setcounter{tocdepth}{2}
\printcontents[appendix]{}{1}{}
\endgroup
\section{Discussion}
\label{app:discusison}
In this section, we provide additional analysis and discussion of OptiFlow. We first discuss its limitations and then examine two key aspects of its transport-based formulation. We show why value information should be encoded in the reference marginal rather than solely in the transport cost, and analyze alternative distillation strategies to clarify the role of global transport-based matching in preserving multimodality. Finally, we characterize the computational cost of OptiFlow, including its scaling with the number of Sinkhorn iterations and policy samples.

\subsection{Limitations}
\label{sec:limitations}

OptiFlow introduces additional training-time computation because it solves a state-wise entropic optimal transport problem over sampled one-step-policy and reference-policy actions. This requires Sinkhorn iterations during training, making OptiFlow more expensive than purely pointwise distillation or direct actor updates. However, this overhead does not affect deployment: after training, OptiFlow uses only the one-step flow policy for action generation, so inference remains as efficient as other one-step policies. Appendix~\ref{app:compute} discusses the trade-off between Sinkhorn iterations, training cost, and policy performance.

OptiFlow is most beneficial when the reference flow policy provides multiple useful candidate actions and the supervision mass get allocated to advantage. When the reference action distribution is effectively simple, or when a single local target is already sufficient for policy extraction, the advantage over simpler one-step distillation methods may be smaller.

\subsection{Why use a value-aware reference marginal in optimal transport?}\label{app:uniform_marginal}

We construct a simple $2\times 2$ example showing that, under a uniform reference marginal, incorporating value information solely through the transport cost is insufficient: regardless of the cost design, the induced transport plan cannot guarantee policy improvement.

\paragraph{Counterexample.}
Consider a \(2 \times 2\) transport problem with uniform student and teacher
marginals in certain state $s\in\cS$,
\[
    p = \left(\frac12, \frac12\right),
    \qquad
    q = \left(\frac12, \frac12\right).
\]
The transport plan \(P\) must satisfy
\[
    \sum_j P_{ij} = p_i,
    \qquad
    \sum_i P_{ij} = q_j.
\]
Suppose both target policy's actions have the same value,
\[
    Q^\pi(s,a_{\theta,1}^s) = Q^\pi(s,a_{\theta,2}^s) = 7,
\]
while the reference policy's actions have highly asymmetric values,
\[
    Q^\pi(s,\tilde a_{\omega,1}^s) = 10,
    \qquad
    Q^\pi(s,\tilde a_{\omega,2}^s) = 1.
\]
Even if the cost function is designed so that both target policy actions prefer the first reference policy's action, e.g.,
\[
    c_{11} < c_{12},
    \qquad
    c_{21} < c_{22},
\]
the uniform reference-side marginal constraint still imposes
\[
    P_{11} + P_{21} = \frac12,
    \qquad
    P_{12} + P_{22} = \frac12.
\]
Therefore, at least half of the total mass must be transported to the
low-value reference action \(a_2^\omega\). This is independent of how the cost is
constructed, as long as the reference-side marginal follows uniform distribution.

Indeed, the expected value induced by any feasible transport plan
is fixed by the reference-side marginal:
\[
    \sum_{i,j} P_{ij} Q^\pi(s,\tilde a_{\omega,j}^s)
    =
    \sum_j q_j Q^\pi(s,\tilde a_{\omega,j}^s)
    =
    \frac12 \cdot 10 + \frac12 \cdot 1
    =
    5.5.
\]
Thus, any cost formulation cannot improve the target policy in this example, unless the uniform marginal constraint can be mitigated or changed.

In contrast, our softmax-marginal formulation sets the reference-side marginal
according to the value,
\[
    w_j
    =
    \frac{\exp(Q^\pi(s,\tilde a_{\omega,j}^s) / \tau)}
    {\sum_k \exp(Q^\pi(s,\tilde a_{\omega,k}^s) / \tau)}.
\]
When \(\tau\) is small enough, the marginal concentrates on the high-value reference action:
\[
    q^{\tau^*} \approx (1,0).
\]
Then the transport plan
\[
    P \approx
    \begin{pmatrix}
    \frac 12 & 0 \\
    \frac 12 & 0
    \end{pmatrix}
\]
is feasible, and both student actions can be guided toward the optimal teacher
action. The induced teacher-side value becomes approximately
\[
    \sum_j q^{\tau^*}_j Q^\pi(s,\tilde a_{\omega,j}^s) \approx 10 > 7,
\]
rather than the uniform-marginal value \(5.5\). This counterexample supports why the value information should be encoded in the reference-side marginal, not in the transport cost.

\textcolor{black}{
\paragraph{Generalization beyond the $2\times2$ example.}
\label{app:generalization_of_example}
The observation above extends to an arbitrary number of sampled actions.
For any state $s$, let
$\{\tilde a_{\omega,j}^s\}_{j=1}^{M}$ denote the reference actions and
$q^s$ their prescribed marginal. For any feasible coupling
$P \in \Pi(p^s,q^s)$, we have
\[
    \sum_{i,j} P_{ij}
    Q^\pi(s,\tilde a_{\omega,j}^s)
    =
    \sum_j q_j^s
    Q^\pi(s,\tilde a_{\omega,j}^s),
\]
since $\sum_i P_{ij}=q_j^s$.
Thus, the expected value of the reference actions assigned through the
coupling is determined solely by the reference-side marginal $q^s$,
regardless of the transport cost.
}
\textcolor{black}{
Under a uniform reference marginal, \textbf{this quantity reduces to the average
value of the sampled reference actions, regardless of the cost matrix.}
}

\textcolor{black}{
In contrast, the value-weighted marginal assigns more mass to higher-value
reference actions, where the policy improvement is controlled by $\tau$: the expected assigned value increases monotonically as $\tau$ decreases, from the candidate average at $\tau\rightarrow \infty$ to the best candidate value at $\tau\rightarrow0$, so a sufficiently small $\tau$ guarantees improvement whenever some candidiates beats the current policy. We use softmax rater than an argmax for $q^s$ since the latter concentrates all supervision on one candidate and collapses the multimodal structure, so $\tau$ trades assigned value against mode coverage.
}
\textcolor{black}{
\paragraph{Empirical effect of the reference marginal.}
We further evaluate whether the choice of reference marginal has a
corresponding empirical effect. We compare full OptiFlow, which combines
VaBC with the value-weighted marginal, against (i) a variant that retains
VaBC but replaces the value-weighted marginal with a uniform marginal, and
(ii) OptiFlow-BC, which replaces VaBC with standard BC while retaining the
value-weighted marginal.
}

\begin{table}[H]
\centering
\caption{\textbf{Effect of the reference marginal.}
\label{app:reference_marginal}
We compare full OptiFlow against using a uniform reference marginal and
against replacing VaBC with standard BC while retaining the value-weighted
marginal. We follow the same evaluation protocol as
Table~\ref{tab:ogbench_table_part1}.}
\label{tab:reference_marginal}

\small
\setlength{\tabcolsep}{3pt}
\renewcommand{\arraystretch}{1.15}

\begin{tabularx}{\textwidth}{
@{}>{\raggedright\arraybackslash}p{4.2cm}
>{\centering\arraybackslash}X
>{\centering\arraybackslash}X
>{\centering\arraybackslash}X@{}
}
\toprule
\textbf{Task}
& \shortstack{
    \textbf{\textcolor[rgb]{0.05,0.35,0.75}{OptiFlow}}
  }
& \shortstack{
    \textbf{VaBC + Uniform}
  }
& \shortstack{
    \textbf{BC + Value-weighted}
  } \\
\midrule

{\fontsize{7.5}{8.3}\selectfont antmaze-large-navigate\{1\}}
& \val{\textbf{94.0}}{0.9}
& \val{0.2}{0.2}
& \val{\textbf{91.0}}{2.0} \\

{\fontsize{7.5}{8.3}\selectfont antmaze-giant-navigate\{1\}}
& \val{\textbf{5.4}}{2.4}
& \val{0.0}{0.0}
& \val{0.9}{0.3} \\

{\fontsize{7.5}{8.3}\selectfont humanoidmaze-medium-navigate\{1\}}
& \val{\textbf{82.4}}{1.2}
& \val{3.6}{0.5}
& \val{50.5}{8.3} \\

{\fontsize{7.5}{8.3}\selectfont humanoidmaze-large-navigate\{1\}}
& \val{\textbf{48.3}}{2.2}
& \val{0.0}{0.0}
& \val{43.4}{4.1} \\

{\fontsize{7.5}{8.3}\selectfont antsoccer-arena-navigate\{4\}}
& \val{\textbf{46.8}}{4.1}
& \val{11.6}{2.2}
& \val{43.6}{3.4} \\

{\fontsize{7.5}{8.3}\selectfont cube-single-play\{2\}}
& \val{\textbf{98.5}}{0.5}
& \val{25.9}{3.6}
& \val{\textbf{97.5}}{1.3} \\

{\fontsize{7.5}{8.3}\selectfont cube-double-play\{2\}}
& \val{\textbf{52.4}}{2.9}
& \val{0.6}{0.2}
& \val{39.1}{3.5} \\

{\fontsize{7.5}{8.3}\selectfont scene-play\{2\}}
& \val{\textbf{97.5}}{1.1}
& \val{47.4}{6.3}
& \val{\textbf{98.2}}{0.8} \\

{\fontsize{7.5}{8.3}\selectfont puzzle-3x3-play\{4\}}
& \val{\textbf{25.3}}{6.8}
& \val{1.8}{0.4}
& \val{1.1}{0.7} \\

{\fontsize{7.5}{8.3}\selectfont puzzle-4x4-play\{4\}}
& \val{\textbf{22.8}}{2.9}
& \val{5.5}{0.9}
& \val{20.9}{4.2} \\

\midrule
\textbf{Average}
& \val{\textbf{57.3}}{1.0}
& \val{9.7}{0.7}
& \val{48.6}{1.3} \\

\bottomrule
\end{tabularx}
\end{table}

\textcolor{black}{
Table~\ref{tab:reference_marginal} shows that the value-weighted marginal is
critical to the performance of OptiFlow. Replacing it with a uniform marginal
while retaining VaBC reduces the average performance from 57.3 to 9.7.
In contrast, replacing VaBC with standard BC while retaining the
value-weighted marginal results in a substantially smaller decrease, from
57.3 to 48.6. These results are consistent with the analysis above: the
reference marginal directly controls how much transport mass is allocated to
high-value reference actions, whereas the transport cost alone cannot alter
the total mass assigned to each reference action.
}

\subsection{Alternative distillation methods and why OptiFlow?}
\label{app:swd-tkd}

\begin{figure}[t]
\centering
\newcommand{\panelw}{0.20\linewidth}

\begin{subfigure}[t]{\panelw}
    \centering
    \includegraphics[width=\linewidth]{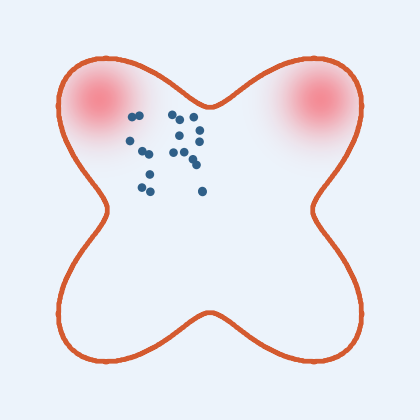}
    \caption{Initialization}
\end{subfigure}\hfill
\begin{subfigure}[t]{\panelw}
    \centering
    \includegraphics[width=\linewidth]{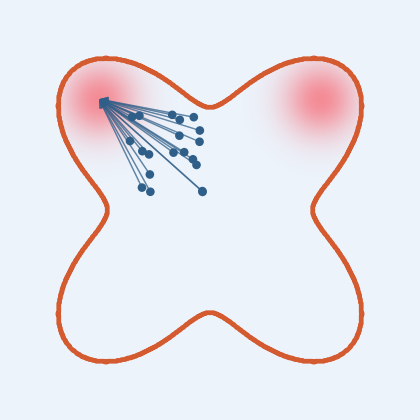}
    \caption{VDWM}
\end{subfigure}\hfill
\begin{subfigure}[t]{\panelw}
    \centering
    \includegraphics[width=\linewidth]{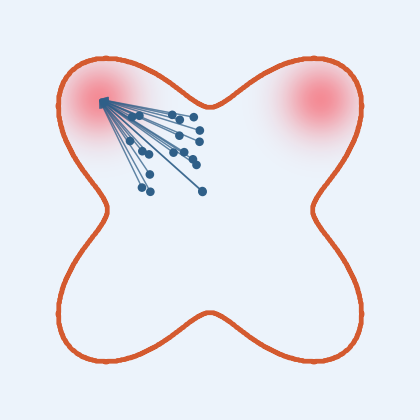}
    \caption{Top-2 NN}
\end{subfigure}\hfill
\begin{subfigure}[t]{\panelw}
    \centering
    \includegraphics[width=\linewidth]{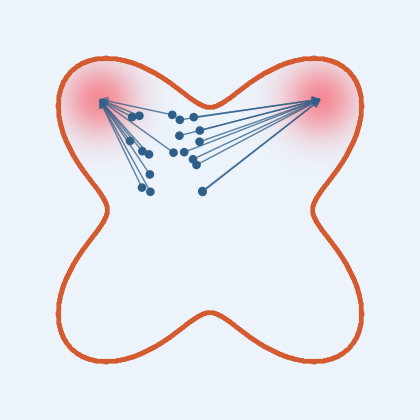}
    \caption{\textbf{OptiFlow (ours)}}
\end{subfigure}
\caption{
Initialization-dependent matching in an equal-value bimodal bandit.
The reference set contains two equal-value optimal anchors, while all generated
particles are initialized in the left half of the support.
Value-distance weighted matching (VDWM) selects an anchor independently for each
particle and therefore assigns all particles to the closer left mode.
Top-2 nearest-neighbor matching still assigns each particle independently to its nearest anchor, again collapsing to the left mode.
In contrast, OptiFlow preserves multimodality since the marginal constraint assigns equal
mass to both modes, preserving the bimodal allocation.
}
\label{fig:init-dependent-matching}
\end{figure}

Here, we analyze alternative distillation strategies such as value-weighted and nearest-neighbor matching, and discuss their failure modes illustrated in Figure~\ref{fig:init-dependent-matching}.

Our method can be viewed as combining ideas from value-weighted distillation
(VWD) and Top-K nearest-neighbor (Top-K~NN) matching, while addressing
their limitations. Define reference-policy actions
\(\{\tilde a_{\omega,j}^s\}_{j=1}^N\), where \(\tilde a_{\omega,j}^s=\mu_{\omega}(s,z_j)\), and
one-step policy actions \(\{a_{\theta,i}^s\}_{i=1}^N\), where
\(a_{\theta,i}^s=\mu_\theta(s,z_i)\). We use the same noise bank for both policies,
i.e., \(z_i=z_j\) when \(i=j\), with \(z_i\sim\mathcal{N}(0,I)\). VWD assigns a value-based weight to each reference action,
\[
    w_j
    =
    \frac{f(Q(s,\tilde a_{\omega,j}^s))}
    {\sum_{\ell=1}^N f(Q(s,\tilde a_{\omega,\ell}^s))} .
\]
It then minimizes
\[
    \mathcal{L}_{\mathrm{VWD}}
    =
    \sum_{j=1}^N
    w_j
    \left\|a_j^\theta - \tilde a_{\omega,j}^s\right\|_2^2 .
\]
In practice, we consider $w_j$ as:
\[
    w_j
    =
    \frac{\exp(Q(s,\tilde a_{\omega,j}^s)/\tau)}
    {\sum_{\ell=1}^N \exp(Q(s,\tilde a_{\omega,\ell}^s)/\tau)} .
\]
Top-K~NN first selects the $K$ highest-value reference samples,
\[
    \mathcal{T}_K(s)
    =
    \operatorname{TopK}_{j \in \{1,\dots,M\}}
    Q(s,\tilde a_{\omega,j}^s),
\]
and then assigns each one-step policy's action to its nearest reference action within this
set,
\[
    j^*_i
    =
    \arg\min_{j \in \mathcal{T}_K(s)}
    \left\|a_{\theta,i}^s - \tilde a_{\omega,j}^s\right\|_2^2 .
\]
The corresponding distillation loss is
\[
    \mathcal{L}_{\mathrm{TopK}}
    =
    \sum_{i=1}^N
    \left\|a_{\theta,i}^s - \tilde a_{\omega,j^*_i}^s\right\|_2^2 .
\]

In the bandit example in Figure~\ref{fig:bandit_toy}, VWD collapses to the middle
point, a well-known failure mode when distilling multi-step flow models into a
one-step policy with naive \(\ell_2\) regression~\citep{frans2024one}. Top-K~NN is also sensitive to the choice of $K$. When \(K=1\), the policy collapses
to a single selected mode. When \(K=3\), it can preserve some multimodality, but
it still cannot suppress probability mass in the intermediate region between
modes. Although Top-K~NN shows potential for capturing multimodality, it
relies on hard selection and does not specify how much mass each selected
reference action should receive, making it overly restrictive and sensitive to
the choice of $K$.

Now let's consider Value-Distance Weighted Matching(VDWM) distillation, more general formulation which is closer to our algorithm. Define a score
\[
\xi_{ij} = f(d_{ij}, Q(s,a_{\theta,i}^s), Q(s,\tilde a_{\omega,j}^s)),
\]
and perform weighted distillation:
\[
\mathcal{L}_{\mathrm{VDWM}} = \sum_i \xi_{i j_i^*} \, \|a_{\theta,i}^s - \tilde a_{\omega,j_i^*}^s\|^2,
\]
where $j_i^* = \arg\max_j s_{ij}$. For instance, one may choose
\[
\xi_{ij} = \frac{\exp\big(-d_{ij} + \alpha (Q(s,\tilde a_{\omega,j}^s) - Q(s,a_{\theta,i}^s))\big)}{\sum_{k=1}^M \exp\big(-d_{ik} + \alpha (Q(s,\tilde a_{\omega,k}^s) - Q(s,a_{\theta,i}^s))\big)}
\]

However, this approach performs independent selection for each $i$. As a result, it does not impose any global mass-allocation constraint across reference actions, and therefore has probability of collapse onto a subset of modes. In particular, when multiple target modes have comparable value, small differences in initialization, distance, or estimated value can break the symmetry arbitrarily, causing many samples to select the same mode.

In contrast, our method constructs a global coupling with value-conditioned
marginals. This induces a structured allocation of mass across reference
actions, so multiple high-value modes can be represented proportionally rather
than selected independently. As a result, our method mitigates mode collapse and
better preserves multimodality. We illustrate this behavior with the bandit
example in Figure~\ref{fig:init-dependent-matching}.

\subsection{Computational cost}
\label{app:compute}

OptiFlow introduces additional computational cost from three sources: (i) per-state Sinkhorn updates for optimal transport, (ii) action generation from the $K$-step reference flow model, and (iii) action generation from the one-step policy. We analyze how these factors contribute to the computation cost.

First, we identify the training-time bottleneck by varying the number of Sinkhorn iterations and analyzing the computational overhead introduced by OT updates.

Second, we analyze how the computational cost scales with the number of one-step flow policy and reference $K$-step flow policy samples $(N, M)$, which affect reference action generation and transport matching. The corresponding experiments are reported in Appendix~\ref{app:sampling_abal}.

Finally, we compare OptiFlow against alternative baselines in terms of both training and inference runtime. 

\paragraph{Effect of Sinkhorn iterations on runtime.} \textcolor{black}{We use {30} Sinkhorn iterations as the default setting. To assess whether this choice introduces unnecessary computational overhead, we vary the number of iterations and measure the average computation time per training step. As shown in Table~\ref{tab:n_ot_iter_abl}, reducing the number of iterations from {30} to {20} or {10} yields only marginal runtime improvements, suggesting that Sinkhorn updates are not a dominant training-time bottleneck. We analyze the corresponding performance behavior in Table~\ref{tab:sinkhorn_param_abal}b.
} 

\begin{table}[H]
\centering
\scriptsize
\setlength{\tabcolsep}{8pt}
\renewcommand{\arraystretch}{1.1}
\begin{tabular}{@{}ccccc@{}}
\toprule
\textbf{Sinkhorn Iterations} & 10 & 20 & \textbf{30} (default) & 40 \\
\midrule
\textbf{ms/step} & \val{5.14}{0.04} & \val{5.26}{0.04} & \val{5.37}{0.04} & \val{5.47}{0.04} \\
\bottomrule
\end{tabular}
\caption{
\textbf{Effect of the number of Sinkhorn iterations on runtime.}
We report the average computation time per training step (ms/step) over 15 tasks measured on a single NVIDIA H100 GPU.}
\label{tab:n_ot_iter_abl}
\end{table}

\paragraph{Computational scaling with the number of samples $(N, M)$.}
We analyze how the computational cost scales with the number of one-step flow
policy samples and reference $K$-step flow policy samples $(N, M)$, which jointly affect both reference action generation and transport matching.
As shown in Table~\ref{tab:nm_timing_all}, the computational cost is more sensitive to the number of reference-policy samples $M$ than to the number of one-step-policy samples $N$. In particular, increasing $M$ introduces substantially higher runtime overhead due to the increased cost of reference action generation and transport matching, whereas increasing $N$ results in comparatively moderate overhead. We analyze the corresponding performance behavior in Appendix~\ref{app:sampling_abal}.

\begin{table}[H]
\centering
\footnotesize
\setlength{\tabcolsep}{8pt}
\renewcommand{\arraystretch}{1.3}
\begin{tabular}{@{}c|cccc|c@{}}
\toprule
$N \backslash M$ & 4 & 16 & \textbf{64} (default) & 256 & Avg($M$) \\
\midrule
4 & \val{3.73}{0.13} & \val{3.53}{0.01} & \val{4.99}{0.04} & \val{16.74}{0.10} & \val{7.25}{0.93} \\
8 & \val{3.39}{0.07} & \val{3.65}{0.01} & \val{5.10}{0.04} & \val{16.57}{0.09} & \val{7.18}{0.92} \\
\textbf{16} (default) & \val{3.42}{0.01} & \val{3.74}{0.02} & \val{5.36}{0.04} & \val{16.83}{0.09} & \val{7.34}{0.94} \\
32 & \val{3.50}{0.02} & \val{3.85}{0.01} & \val{5.76}{0.04} & \val{17.42}{0.09} & \val{7.63}{0.97} \\
\midrule
Avg($N$) & \val{3.51}{0.04} & \val{3.69}{0.02} & \val{5.30}{0.05} & \val{16.89}{0.07} & \\
\bottomrule
\end{tabular}
\caption{
\textbf{Effect of the number of samples $(N,M)$ on runtime.} We report the average computation time per training step (ms/step) over 15 tasks measured on a single NVIDIA H100 GPU.
}
\label{tab:nm_timing_all}
\end{table}

\paragraph{Training and inference runtime comparison.}

\begin{figure}[H]
    \centering
    \includegraphics[width=0.48\linewidth]{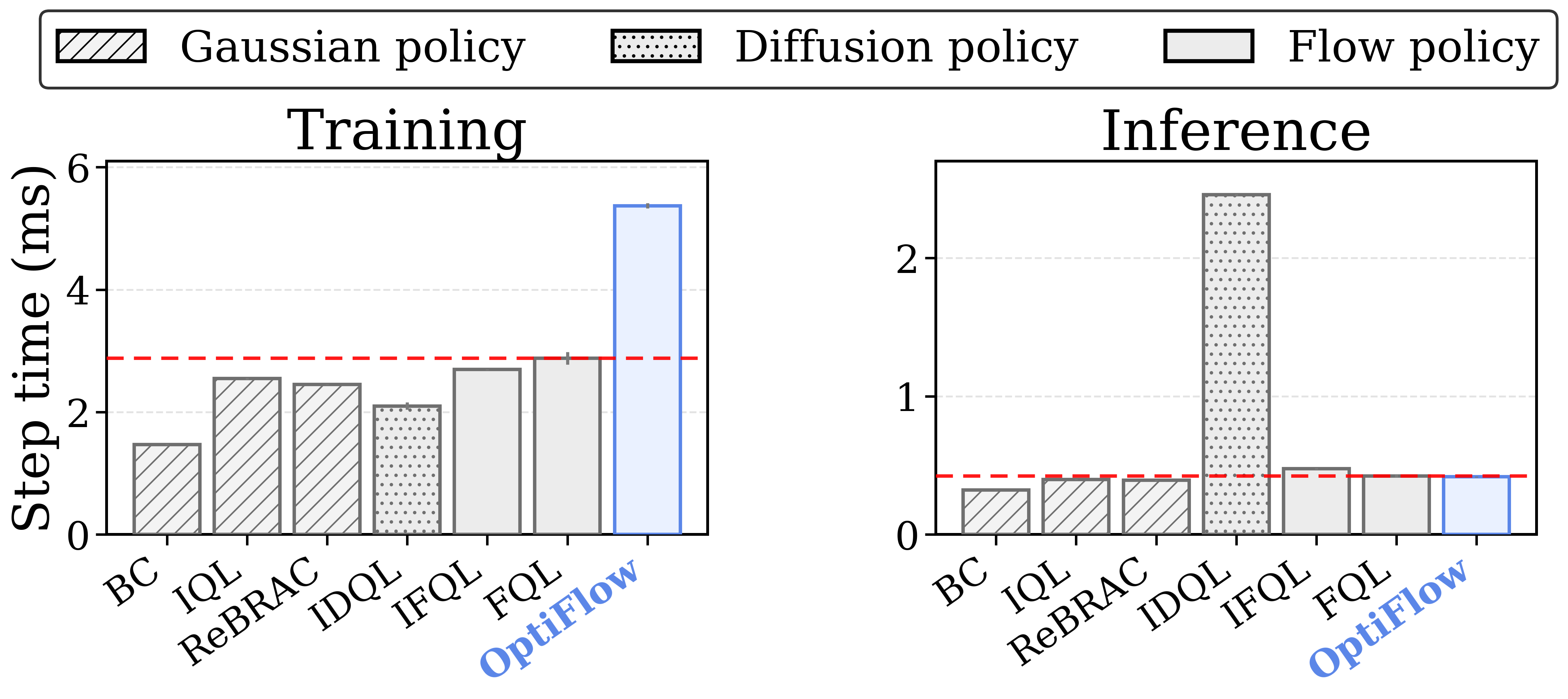}
    \caption{\textbf{Comparison of training and inference time across methods.} We report the average training time per gradient step (ms/step), averaged over 15 tasks and measured on a single NVIDIA H100 GPU.}
    \label{fig:training_inference_runtime}
\end{figure}

Finally, we compare the training and inference runtime of OptiFlow with alternative baselines. As shown in Figure~\ref{fig:training_inference_runtime}, OptiFlow introduces additional training-time overhead compared to standard one-step methods due to the need to generate and match multiple samples $(N,M)$ from the one-step and reference flow policies. In contrast, inference remains efficient since the learned policy requires only a single forward pass at test time.
Combined with our Sinkhorn iteration analysis, these results suggest that the dominant computational cost arises from sample generation and matching rather than from the OT updates themselves.
Reducing the sampling cost required for effective transport-based distillation is an important direction for future work.

\section{Broader Impacts}
\label{app:broad}
This work studies offline reinforcement learning in simulated control benchmarks. The main potential benefit is improved policy learning from fixed datasets, which may reduce the need for unsafe or expensive online exploration when applying RL to physical systems. OptiFlow also produces a one-step policy at deployment time, which can make generative-policy methods more practical in latency-sensitive control settings.

The main risk is the usual risk of offline RL: a learned policy may behave poorly when deployed outside the coverage of the dataset, especially if the critic assigns inaccurate values to unsupported actions. While OptiFlow is designed to reduce direct critic exploitation by using value information through transport-based target construction, it does not remove the need for careful validation before deployment. In safety-critical applications, the method should be combined with dataset auditing, environment-specific safety constraints, and out-of-distribution monitoring.

\newpage

\section{Sinkhorn Algorithm}
\label{app:sinkhorn}

\subsection{Algorithm}
\begin{algorithm}[H]
\caption{Sinkhorn Algorithm}
\label{alg:sinkhorn}
\small
\begin{algorithmic}[1]
\STATE Initialize $K = \exp(-C / \varepsilon)$
\STATE Initialize $u = \mathbf{1}_N$, $v = \mathbf{1}_M$
\FOR{$t = 1, \dots, T$}
    \STATE $u_i \leftarrow {p_i}/{(K v)_i}$
    \STATE $v_j \leftarrow {q_j}/{(K^\top u)_j}$
\ENDFOR
\STATE $P = \mathrm{diag}(u) K \mathrm{diag}(v)$
\end{algorithmic}
\end{algorithm}

For completeness, we provide the Sinkhorn procedure~\citep{cuturi2013sinkhorn} used to solve the entropy-regularized optimal transport problem in Step~\hyperref[step:ot]{3} of our method. Given the cost matrix $C$, reference-side marginal $p$, target-side marginal $q$, and entropic regularization parameter $\varepsilon$, the algorithm alternates row and column scaling updates to compute the transport plan.

In all experiments, we use a fixed number of $T=30$ Sinkhorn iterations.




\section{Implementation Details}
\label{app:impl}

\paragraph{Architectures.}
The critic, reference flow policy, and one-step policy all use $[512,512,512,512]$ MLPs with GeLU~\citep{hendrycks2016gaussian} activations. LayerNorm~\citep{ba2016layer} is applied to the critic only. The critic is a 2-ensemble; we aggregate with the mean by default, unless otherwise stated as minimum. The one-step policy takes concat$(s,z)$ through the MLP and a linear head (init scale $10^{-2}$); policy outputs are clipped to $[-1,1]$ at use time. All methods are implemented in JAX~\citep{jax2018github}.

\paragraph{Image processing.}
For pixel-based OGBench environments, we use a smaller variant of the IMPALA encoder~\citep{espeholt2018impalascalabledistributeddeeprl} and apply random-shift augmentation with probability $0.5$, following the official implementation of~\citep{park2025flow}. We use 3-frame stacking, prestacked at dataset load time.

\paragraph{Flow teacher.}
We use linear-interpolant flow matching with uniform time sampling and integrate with $K=10$ Euler steps. Time $t$ is concatenated directly into the MLP input.

\paragraph{Value-aware BC.}
The reference action in the VaBC weight $g_\eta$ is the stop-graded one-step policy action, and the minibatch scale $\lambda=(B^{-1}\sum_i|Q_\phi(s_i,a_i)|)^{-1}$ is also stop-graded. In practice, $g_\eta$ is implemented as a 2-class softmax between $Q_\phi(s,a)$ and $Q_\phi(s,\mu_\theta(s,z))$. Gradients from this loss flow only into the reference policy.

\paragraph{Transport and distillation.}
Per state, we sample $N=16$ one-step actions and $M=64$ reference actions. The cost is squared Euclidean distance normalized by its batchwise mean. The one-step marginal is uniform, and the reference marginal is a softmax with temperature $\tau$ applied to reference critic values. We run balanced Sinkhorn in the log domain for $T=30$ iterations, with entropic regularization fixed at $\varepsilon=0.05$ across all tasks. We use hard row-wise anchors $j_i^*=\arg\max_j P^*_{ij}$, with distillation weights $P^*_{i,j_i^*}$. The actor objective combines the VaBC and distillation losses with equal weights.

\paragraph{Training and evaluation.}
We train for 1M gradient steps on state-based OGBench tasks and 500K steps on D4RL. Every 100K steps we evaluate with $100$ episodes and reported scores are taken at the final training step and averaged over $8$ seeds unless otherwise stated. For OGBench visual tasks, we report the average success rates across the last three evaluation epochs (300K, 400K, 500K), each evaluated over 50 episodes, following the protocol of FQL~\citep{park2025flow} due to computational constraints.

\textcolor{black}{
\paragraph{Bandit multimodal diagnostic.}
\label{app:bandit_diagnostic_setup}
For Figure~\ref{fig:bandit_toy}, FQL-D uses $\alpha=15$ and FQL-Q uses $\alpha=1$, where larger $\alpha$ places greater relative weight on the distillation term. For reference, at $\alpha=1000$, the resulting policy becomes nearly indistinguishable from BC. The reward landscape contains two high-value modes with maximum reward $+20$ and a suboptimal band across the
middle with reward $+10$ (the faint red region).
}

\paragraph{Four-mode environment diagnostic}
\label{app:4way_diagnostic_setting}
We consider a two-dimensional point-mass environment adapted from the multi-goal environment of~\citep{haarnoja2017reinforcement}. The state and action spaces are bounded by $[-7,7]^2$ and $[-1,1]^2$, respectively, and the dynamics are given by $s'=\operatorname{clip}(s+a)$. Four symmetric goals are located at
$\mathcal{G}=\{(5,0),(-5,0),(0,5),(0,-5)\}$.
The transition reward is
\[
r(s,a,s')
=
-30\lVert a\rVert_2^2
-\min_{g\in\mathcal{G}}\lVert s'-g\rVert_2^2
+10\ \cdot\mathbb{I}
\!\left[
\min_{g\in\mathcal{G}}\lVert s'-g\rVert_2<1
\right],
\]
and an episode terminates upon reaching any goal or after 30 steps. The reward contours in Figure~\ref{fig:mode_coverage_env} visualize the state-dependent term
$-\min_{g\in\mathcal{G}}\lVert s-g\rVert_2^2$; consequently, all four goals are equally optimal from the symmetric initial region.

We first collect 100,000 transitions using a uniform-random behavior policy. To introduce asymmetric coverage while retaining all four directional modes, we independently remove $80\%$ of transition tuples $(s,a,s')$ for which either $s$ or $s'$ lies below the diagonal $y=x$. This leaves 55,044 transitions and changes the initial action proportions from approximately uniform to
$\mathrm{R/L/T/B}=16.2/35.4/34.7/13.7\%$.
For FQL, we sweep
$\alpha\in\{0.01,0.1,0.3,0.5,1,3,5,10\}$,
where larger $\alpha$ places greater relative weight on flow-policy distillation. The figure presents representative results for
$\alpha\in\{0.1,1,10\}$.
For OptiFlow-BC, we use a standard flow-BC reference policy and set the value-weighted marginal temperature to $\tau=0.1$; no VaBC temperature $\eta$ is used.

\paragraph{OT assignment stability diagnostic}
\label{app:assignment_stability_setting}
Using the same four-mode environment, we examine whether OT assignments remain directionally consistent when the reference actions are resampled. With all networks frozen, we fix four one-step actions at a given state, with one action pointing toward each of the right, left, top, and bottom goals. We then independently resample $M=64$ reference actions 100 times and recompute the value-weighted marginal, Sinkhorn transport plan, and hard assignments at every round. Each assigned reference action is labeled according to the goal direction toward which it moves. We perform this diagnostic at the symmetric origin, where all four directions are equally valuable, and at nearby states where the values are no longer tied.

\paragraph{Online fine-tuning.}
For offline-to-online RL, we extend offline OptiFlow training with 1M additional gradient steps starting from the 1M-step offline checkpoint, adding online transitions to a unified replay buffer that already contains the offline dataset. Online data are collected using the deterministic one-step policy with no exploration noise, and we continue to train all components of OptiFlow with the same objective as in offline training.

\paragraph{Hyperparameters.}
We provide complete lists of hyperparameters used across all tasks and methods in Tables~\ref{tab:OptiFlow_shared_hparams} to \ref{tab:hparams_d4rl}. We use the hyperparameters setup shown in~\citep{park2025flow} for the baseline experiments and those presented in~\citep{tiofack2026guided}.

\paragraph{Modified  TD target.}
Standard TD target takes the form $y = r + \gamma Q(s', a')$. \citep{tiofack2026guided} presents a variant of the TD target $y^{\text{VaBC}}(s, r, s') = r + \frac{\gamma}{2}\left(Q_{\bar{\phi}}(s', \mu_\theta(s', z)) + Q_{\bar{\phi}}(s', \mu_\omega(s', z))\right)$ where $\quad z \sim \mathcal{N}(0, I_d)$. Here $\mu_\theta(s', z)$ denotes the action from the one-step policy and $\mu_\omega(s', z)$ denotes the action from the reference policy. This TD target is formulated as the average of the two estimates of Q-value. This is a more conservative variant of the TD target, and we report results of OptiFlow and GFP on the specified tasks in Table~\ref{tab:OptiFlow_hparams_override}
\newpage

\begin{table}[H]
\centering
\caption{\textbf{Shared hyperparameters used across all tasks and methods} unless otherwise noted in Table~\ref{tab:OptiFlow_hparams_override}. We evaluated across all tasks and methods equally according to Table~\ref{tab:OptiFlow_shared_hparams} and \ref{tab:OptiFlow_hparams_override} where relevant.}
\label{tab:OptiFlow_shared_hparams}
\small
\setlength{\tabcolsep}{2pt}
\renewcommand{\arraystretch}{1.3}
\begin{tabular}{@{}l@{\hspace{2em}}>{\raggedright\arraybackslash}p{8.5cm}@{}}
\toprule
\textbf{Hyperparameter} & \textbf{Value} \\
\midrule
MLP hidden dims                            & $[512, 512, 512, 512]$ \\
Activation                                 & GeLU~\citep{hendrycks2016gaussian} \\
LayerNorm~\citep{ba2016layer}              & Critic only \\
Clipped double Q-learning                  & False (default), True (antmaze-\{large, giant\}-navigate, adroit) \\
\midrule
Optimizer                                  & Adam~\citep{kingma2014adam} \\
Learning rate                              & $3\times 10^{-4}$ \\
Discount $\gamma$                          & $0.99$ (default), $0.995$ (antmaze-giant, humanoidmaze, antsoccer) \\
TD target                                  & Standard \\
Target network smoothing coefficient       & $0.005$ \\
Batch size                                 & $256$ \\
Gradient steps                             & 1M (OGBench), 500k (OGBench visual tasks, D4RL) \\
Critic update                              & Every step (default), every 5 steps (antsoccer for OptiFlow only) \\
\midrule
Euler integration steps $K$                & $10$ \\
\midrule
Student samples $N$                        & $16$ \\
Teacher samples $M$                        & $64$ \\
Sinkhorn iterations $T$                    & $30$ \\
Entropic regularization $\varepsilon$      & $0.05$ \\
\bottomrule
\end{tabular}
\end{table}

\begin{table}[H]
\centering
\caption{\textbf{Per-task overrides to the shared defaults} in
Table~\ref{tab:OptiFlow_shared_hparams} for OptiFlow and GFP only. We bring these overrides exactly from GFP setup. All other baselines and tasks not listed use the shared defaults. We make these overrides in order to maintain consistency with all baselines following the convention, and to provide accurate comparison against GFP which makes the following exact overrides due to apparently significant performance improvements. OptiFlow also demonstrates slightly improved performance with this setting. We report OptiFlow and GFP results on the following tasks with these overrides and all other tasks not listed follow the shared defaults in Table~\ref{tab:OptiFlow_shared_hparams}.}
\label{tab:OptiFlow_hparams_override}
\small
\setlength{\tabcolsep}{4pt}
\renewcommand{\arraystretch}{1.4}
\begin{tabular}{@{}>{\raggedright\arraybackslash}p{8cm}cc@{}}
\toprule
\textbf{Task} & \textbf{Discount $\gamma$} & \textbf{TD target} \\
\midrule
antmaze-large-navigate-singletask-task\{1-5\}-v0        & 0.995 & -- \\
humanoidmaze-large-navigate-singletask-task\{1-5\}-v0   & 0.999 & -- \\
humanoidmaze-medium-navigate-singletask-task\{1-5\}-v0  & --    & $y^{\text{VaBC}}$ \\
cube-single-play-singletask-task\{1-5\}-v0              & --    & $y^{\text{VaBC}}$ \\
cube-double-play-singletask-task\{1-5\}-v0              & --    & $y^{\text{VaBC}}$ \\
cube-triple-play-singletask-task\{1-5\}-v0              & --    & $y^{\text{VaBC}}$ \\
\bottomrule
\end{tabular}
\end{table}

\newpage
\begin{table}[H]
\centering
\caption{\textbf{Task-specific hyperparameters for offline RL on OGBench.} We use the hyperparameters used by For OptiFlow, $\tau$ is the teacher-marginal softmax temperature and $\eta$ is the value-aware BC temperature. "-" indicates that the experimental result is taken from prior work (or does not exist).}
\label{tab:hparams_ogbench}
\scriptsize
\setlength{\tabcolsep}{4pt}
\renewcommand{\arraystretch}{1.1}
\begin{tabular}{@{}>{\raggedright\arraybackslash}p{5.4cm}cccccccc@{}}
\toprule
\textbf{Task} & BC & IQL & ReBRAC & IDQL & IFQL & FQL & GFP & \textbf{\textcolor[rgb]{0.05,0.35,0.75}{OptiFlow}} \\
              &    & $\alpha$ & $(\alpha_1, \alpha_2)$ & $N$ & $N$ & $\alpha$ & $(\alpha, \eta)$ & $(\tau, \eta)$ \\
\midrule
antmaze-large-navigate-singletask\{1-5\}-v0       & -- & 10 & (1e-3, 1e-2) & 32 & 32 & 10 & (0.3,1e-4) & (2.0, 1e-2) \\
antmaze-giant-navigate-singletask\{1-5\}-v0       & -- & 10 & (1e-3, 1e-2) & 32 & 32 & 10 & (0.1,1e-1) & (1.0, 1e-3) \\
\midrule
humanoidmaze-medium-navigate-singletask\{1-5\}-v0 & -- & 10 & (1e-2,1e-2) & 32 & 32 & 30 & (0.3,1e-3) & (1.0, 1e-3) \\
humanoidmaze-large-navigate-singletask\{1-5\}-v0  & -- & 10 & (1e-2,1e-2) & 32 & 32 & 30 & (0.3,1e-4) & (2.0, 1e-1) \\
\midrule
antsoccer-arena-navigate-singletask\{1-5\}-v0     & -- & 1 & (1e-2,1e-2) & 32 & 64 & 10 & (0.1,1e-2) & (1.0, 1e-2) \\
\midrule
cube-single-play-singletask\{1-5\}-v0             & -- & 1 & (1,0) & 32 & 32 & 300 & (10,1e-1) & (2.0, 1e-1) \\
cube-double-play-singletask\{1-5\}-v0             & -- & 0.3 & (1e-1,0) & 32 & 32 & 300 & (1.0,1e-2) & (2.0, 1e-2) \\
cube-triple-play-singletask\{1-5\}-v0             & -- & 1 & (1e-1,0) & 64 & 128 & 300 & (0.1,1e-5) & (1.0, 1e-5) \\
\midrule
scene-play-singletask\{1-5\}-v0                   & -- & 10 & (1e-1,1e-2) & 32 & 32 & 300 & (10,1e-3) & (1.0, 1e-2) \\
\midrule
puzzle-3x3-play-singletask\{1-5\}-v0              & -- & 10 & (3e-1,1e-2) & 32 & 32 & 1000 & (3.0,1e-3) & (1.0, 1e-3) \\ 
puzzle-4x4-play-singletask\{1-5\}-v0              & -- & 3 & (3e-1,1e-2) & 32 & 32 & 1000 & (3.0,1e-5) & (2.0,1e-2) \\
\midrule
visual-cube-single-play-singletask1-v0            & -- & -- & -- & -- & -- & -- & (10,1e-1) & (1.0,1e-1) \\
visual-cube-double-play-singletask1-v0            & -- & -- & -- & -- & -- & -- & (0.3,1e-2) & (2.0,1e-2) \\
visual-scene-play-singletask1-v0                  & -- & -- & -- & -- & -- & -- & (10,1e-3) & (1.0,1e-1) \\
visual-puzzle-3x3-play-singletask1-v0             & -- & -- & -- & -- & -- & -- & (3.0, 1e-2) & (1.0,1e-2) \\
visual-puzzle-4x4-play-singletask1-v0             & -- & -- & -- & -- & -- & -- & (1.0, 1e-4) & (1.0,1e-1) \\
\bottomrule
\end{tabular}
\end{table}

\begin{table}[H]
\centering
\caption{\textbf{Task-specific hyperparameters for offline RL on D4RL.} For OptiFlow, $\tau$ is the teacher-marginal softmax temperature and $\eta$ is the value-aware BC temperature. "-" indicates that the experimental result is taken from prior work (or does not exist).}
\label{tab:hparams_d4rl}
\scriptsize
\setlength{\tabcolsep}{4pt}
\renewcommand{\arraystretch}{1.1}
\begin{tabular}{@{}>{\raggedright\arraybackslash}p{2.7cm}cccccccc@{}}
\toprule
\textbf{Task} & BC & IQL & ReBRAC & IDQL & IFQL & FQL & GFP & \textbf{\textcolor[rgb]{0.05,0.35,0.75}{OptiFlow}} \\
              &    & $\alpha$ & $(\alpha_1, \alpha_2)$ & $N$ & $N$ & $\alpha$ & $(\alpha, \eta)$ & $(\tau, \eta)$ \\
\midrule
antmaze-umaze          & -- & -- & -- & -- & 32 & 10 & (0.1,1e-3) & (1.0, 1e-3) \\
antmaze-umaze-diverse  & -- & -- & -- & -- & 32 & 10 & (0.1,1e-3) & (1.5, 1e-6) \\
antmaze-medium-play    & -- & -- & -- & -- & 32 & 10 & (0.03,1e-3) & (0.8, 1e-3) \\
antmaze-medium-diverse & -- & -- & -- & -- & 32 & 10 & (0.03,1e-3) & (2.0, 1e-4) \\
antmaze-large-play     & -- & -- & -- & -- & 32 & 3 & (0.03,1e-5) & (0.5, 1e-5) \\
antmaze-large-diverse  & -- & -- & -- & -- & 32 & 3 & (0.03,1e-5) & (0.8, 1e-4) \\
\midrule
pen-human-v1           & -- & -- & -- & 32 & 32 & 10000 & (3.0,1e-4) & (5.0, 1e-3) \\
pen-cloned-v1          & -- & -- & -- & 32 & 32 & 10000 & (3.0,1e-5) & (5.0, 1e-3) \\
pen-expert-v1          & -- & -- & -- & 32 & 32 & 3000 & (1.0,1e-3) & (3.0, 1e-3) \\
door-human-v1          & -- & -- & -- & 32 & 32 & 30000 & (10,1e-2) & (3.0, 1e-5) \\
door-cloned-v1         & -- & -- & -- & 32 & 128 & 30000 & (10,1e-2) & (1.0, 1e-6) \\
door-expert-v1         & -- & -- & -- & 32 & 32 & 30000 & (10,1e-2) & (10,  1e-3) \\
hammer-human-v1        & -- & -- & -- & 128 & 32 & 30000 & (10,1e-5) & (5.0, 1e-5) \\
hammer-cloned-v1       & -- & -- & -- & 32 & 32 & 10000 & (10,1e-5) & (50,  1e-3) \\
hammer-expert-v1       & -- & -- & -- & 32 & 32 & 30000 & (10,1e-2) & (5.0, 1e-2) \\
relocate-human-v1      & -- & -- & -- & 32 & 128 & 10000 & (10,1e-4) & (10,  1e-3) \\
relocate-cloned-v1     & -- & -- & -- & 64 & 32 & 30000 & (10,1e-4) & (5.0, 1e-4) \\
relocate-expert-v1     & -- & -- & -- & 32 & 32 & 30000 & (10,1e-4) & (5.0, 1e-4) \\
\bottomrule
\end{tabular}
\end{table}

\newpage
\section{Complete Experimental Results}
\label{app:results}

This section presents experimental results spanning diverse OGBench (Table~\ref{tab:ogbench_table_part1}, ~\ref{tab:ogbench_table_part2}) and D4RL (Table~\ref{tab:d4rl_table}) benchmarks tasks. Table~\ref{tab:OptiFlow_shared_hparams} and Table~\ref{tab:OptiFlow_hparams_override} indicate the hyperparameters used across all tasks and methods. All results are averaged over 8 random seeds except for the visual tasks which are averaged over 4 random seeds due to high computational cost.

\begin{table}[H]
\centering
\caption{\textbf{OGBench results part 1/2.} All methods are trained for 1M gradient steps (500K for visual tasks) and evaluated at the final step over 100 episodes. We report success rate as mean $\pm$ standard error across 8 random seeds for state-based tasks and 4 random seeds for visual tasks. \textbf{Bold} indicates results within 95\% of the best score on each task. \textit{Italics} indicate results taken from prior work~\citep{park2025flow}.}
\label{tab:ogbench_table_part1}
\scriptsize
\setlength{\tabcolsep}{2pt}
\renewcommand{\arraystretch}{1.05}
\begin{tabularx}{\textwidth}{@{}>{\raggedright\arraybackslash}p{1.5cm}*{7}{>{\centering\arraybackslash}X}!{\hskip 4pt\vrule\hskip 2pt\vrule\hskip 4pt}>{\centering\arraybackslash}X@{\hspace{6pt}}}
\toprule
\textbf{Task} & BC & IQL & ReBRAC & IDQL & IFQL & FQL & GFP & \textbf{\textcolor[rgb]{0.05,0.35,0.75}{OptiFlow}} \\
\midrule

\grouprownine{antmaze-large-navigate-singletask}
task1-v0 & \val{0.0}{0.0} & \val{51.4}{4.3} & \val{\textbf{93.3}}{1.5} & \val{58.1}{3.1} & \val{31.9}{6.0} & \val{79.5}{2.1} & \val{\textbf{94.5}}{1.4} & \val{\textbf{94.0}}{0.9} \\
task2-v0 & \val{0.0}{0.0} & \val{40.3}{5.3} & \val{\textbf{87.7}}{1.1} & \val{0.0}{0.0} & \val{8}{1.9} & \val{61.4}{3.6} & \val{\textbf{91.1}}{1.4} & \val{\textbf{87.6}}{1.1} \\
task3-v0 & \val{0.9}{0.4} & \val{71.9}{3.1} & \val{75}{6.4} & \val{27.5}{4.9} & \val{34.5}{8.4} & \val{\textbf{94.6}}{1.2} & \val{\textbf{97.2}}{0.7} & \val{\textbf{93.8}}{1.0} \\
task4-v0 & \val{6.3}{0.7} & \val{48.5}{3.9} & \val{\textbf{87.4}}{1.6} & \val{0.0}{0.0} & \val{18.9}{5.2} & \val{81.4}{1.9} & \val{\textbf{91.3}}{0.8} & \val{\textbf{88.8}}{1.1} \\
task5-v0 & \val{1.3}{0.5} & \val{53.1}{7.9} & \val{86.5}{2.4} & \val{48.5}{3.8} & \val{46.6}{4.6} & \val{81.3}{1.2} & \val{\textbf{95.6}}{0.9} & \val{\textbf{95.1}}{1.0} \\

\grouprownine{antmaze-giant-navigate-singletask}
task1-v0 & \val{0.0}{0.0} & \val{0.0}{0.0} & \val{\textbf{39.4}}{9.9} & \val{0.0}{0.0} & \val{0.0}{0.0} & \val{10.5}{2.9} & \val{8.6}{2.5} & \val{5.4}{2.4} \\
task2-v0 & \val{0.0}{0.0} & \val{0.1}{0.1} & \val{16.6}{9.0} & \val{0.0}{0.0} & \val{0.0}{0.0} & \val{16.4}{4.9} & \val{\textbf{36.6}}{4.6} & \val{25.9}{3.4} \\
task3-v0 & \val{0.0}{0.0} & \val{0.1}{0.1} & \val{\textbf{53.5}}{8.8} & \val{0.0}{0.0} & \val{0.0}{0.0} & \val{0.0}{0.0} & \val{3.3}{0.7} & \val{0.4}{0.2} \\
task4-v0 & \val{0.0}{0.0} & \val{0.0}{0.0} & \val{0.0}{0.0} & \val{0.0}{0.0} & \val{0.0}{0.0} & \val{0.0}{0.0} & \val{28.1}{7.8} & \val{\textbf{43.0}}{4.3} \\
task5-v0 & \val{0.1}{0.1} & \val{21.4}{4.1} & \val{52.6}{4.8} & \val{0.0}{0.0} & \val{4.5}{2.2} & \val{20.6}{9.8} & \val{17.9}{8.3} & \val{\textbf{57.9}}{4.6} \\

\grouprownine{humanoidmaze-medium-navigate-singletask}
task1-v0 & \val{0.1}{0.1} & \val{32}{3.9} & \val{13.9}{6.5} & \val{2.8}{0.5} & \val{61.9}{8.4} & \val{19}{4.2} & \val{\textbf{84.8}}{2.1} & \val{\textbf{82.4}}{1.2} \\
task2-v0 & \val{0.4}{0.3} & \val{39.5}{2.9} & \val{10.6}{3.2} & \val{3.9}{0.4} & \val{84.3}{8.8} & \val{\textbf{94}}{1.1} & \val{\textbf{92.3}}{1.6} & \val{\textbf{92.9}}{1.7} \\
task3-v0 & \val{4.5}{0.3} & \val{30.1}{1.9} & \val{47.6}{6.2} & \val{1.6}{0.7} & \val{45.9}{17.0} & \val{74}{6.4} & \val{\textbf{85.1}}{4.0} & \val{\textbf{82.1}}{1.8} \\
task4-v0 & \val{0.6}{0.2} & \val{0.3}{0.2} & \val{10.5}{4.4} & \val{0.6}{0.3} & \val{0.0}{0.0} & \val{3}{1.4} & \val{0.1}{0.1} & \val{\textbf{19.1}}{5.4} \\
task5-v0 & \val{1.4}{0.6} & \val{66.9}{4.7} & \val{12.8}{5.6} & \val{5.9}{0.9} & \val{\textbf{98.1}}{0.4} & \val{\textbf{97}}{0.7} & \val{\textbf{96.5}}{1.4} & \val{\textbf{95.5}}{0.7} \\

\grouprownine{humanoidmaze-large-navigate-singletask}
task1-v0 & \val{0.0}{0.0} & \val{2.3}{0.6} & \val{0.8}{0.4} & \val{0.1}{0.1} & \val{12}{2.3} & \val{1}{0.5} & \val{36}{5.8} & \val{\textbf{48.3}}{2.2} \\
task2-v0 & \val{0.1}{0.1} & \val{0.0}{0.0} & \val{0.0}{0.0} & \val{0.0}{0.0} & \val{0.0}{0.0} & \val{0.0}{0.0} & \val{\textbf{0.6}}{0.6} & \val{0.0}{0.0} \\
task3-v0 & \val{0.1}{0.1} & \val{6.5}{1.5} & \val{9.9}{1.6} & \val{1.9}{0.5} & \val{\textbf{51.1}}{3.7} & \val{15.4}{3.0} & \val{8.1}{2.3} & \val{20.9}{1.8} \\
task4-v0 & \val{0.0}{0.0} & \val{1}{0.3} & \val{1.8}{1.5} & \val{0.0}{0.0} & \val{0.3}{0.2} & \val{0.1}{0.1} & \val{\textbf{3.8}}{3.7} & \val{0.1}{0.1} \\
task5-v0 & \val{0.0}{0.0} & \val{1.9}{0.5} & \val{0.9}{0.5} & \val{0.0}{0.0} & \val{0.0}{0.0} & \val{1.5}{1.2} & \val{6.7}{4.9} & \val{\textbf{26.3}}{4.4} \\

\grouprownine{antsoccer-arena-navigate-singletask}
task1-v0 & \val{0.3}{0.2} & \val{15.6}{0.8} & \val{0.1}{0.1} & \val{15.8}{2.6} & \val{66.9}{7.3} & \val{\textbf{79.3}}{2.0} & \val{\textbf{79.1}}{2.5} & \val{67.9}{3.3} \\
task2-v0 & \val{0.5}{0.2} & \val{17.3}{1.6} & \val{0.3}{0.3} & \val{0.0}{0.0} & \val{46.3}{12.7} & \val{\textbf{92.8}}{1.5} & \val{\textbf{94.1}}{1.9} & \val{81.5}{1.6} \\
task3-v0 & \val{0.0}{0.0} & \val{5.8}{1.9} & \val{0.0}{0.0} & \val{1.8}{0.6} & \val{5.8}{3.7} & \val{\textbf{54.8}}{2.8} & \val{\textbf{52.5}}{2.4} & \val{43.3}{3.9} \\
task4-v0 & \val{0.0}{0.0} & \val{3.1}{0.9} & \val{0.3}{0.3} & \val{0.0}{0.0} & \val{26.4}{2.8} & \val{38.1}{4.4} & \val{42.5}{3.8} & \val{\textbf{46.8}}{4.1} \\
task5-v0 & \val{0.0}{0.0} & \val{5.4}{1.4} & \val{0.0}{0.0} & \val{0.0}{0.0} & \val{0.3}{0.3} & \val{\textbf{42.0}}{5.1} & \val{37.2}{3.0} & \val{23.8}{4.2} \\

\grouprownine{cube-single-play-singletask}
task1-v0 & \val{2.5}{1.3} & \val{56.3}{4.3} & \val{92.1}{1.3} & \val{92}{0.8} & \val{79.1}{1.5} & \val{93.5}{1.8} & \val{\textbf{96}}{0.9} & \val{\textbf{99.1}}{0.3} \\
task2-v0 & \val{6.1}{4.3} & \val{53.5}{6.1} & \val{86.7}{4.2} & \val{\textbf{98.5}}{0.5} & \val{89.5}{0.8} & \val{\textbf{96.1}}{0.6} & \val{\textbf{97.7}}{1.1} & \val{\textbf{98.5}}{0.5} \\
task3-v0 & \val{4.9}{1.9} & \val{57.9}{6.2} & \val{89.7}{2.1} & \val{\textbf{99.6}}{0.2} & \val{89.3}{0.9} & \val{\textbf{95.1}}{1.8} & \val{\textbf{98.8}}{0.6} & \val{\textbf{99.0}}{0.6} \\
task4-v0 & \val{0.4}{0.4} & \val{42}{6.1} & \val{87.3}{2.5} & \val{86.5}{0.5} & \val{79}{2.1} & \val{84.5}{3.4} & \val{\textbf{95.7}}{1.0} & \val{\textbf{95.8}}{0.9} \\
task5-v0 & \val{0.0}{0.0} & \val{40.8}{5.5} & \val{83.1}{2.7} & \val{82.9}{2.1} & \val{82.5}{2.7} & \val{85.8}{3.4} & \val{\textbf{93.5}}{1.7} & \val{\textbf{95.4}}{1.2} \\

\grouprownine{cube-double-play-singletask}
task1-v0 & \val{0.0}{0.0} & \val{11.6}{1.1} & \val{27.2}{6.3} & \val{3.9}{1.2} & \val{27.5}{2.8} & \val{50.1}{4.1} & \val{73.5}{3.6} & \val{\textbf{80.4}}{2.3} \\
task2-v0 & \val{0.0}{0.0} & \val{0.0}{0.0} & \val{6.4}{1.2} & \val{4.4}{1.9} & \val{9.8}{1.1} & \val{33.8}{5.1} & \val{48.3}{3.7} & \val{\textbf{52.4}}{2.9} \\
task3-v0 & \val{0.0}{0.0} & \val{0.0}{0.0} & \val{3.3}{0.8} & \val{0.5}{0.2} & \val{5.3}{0.8} & \val{21.1}{2.2} & \val{40.6}{3.7} & \val{\textbf{46.3}}{3.3} \\
task4-v0 & \val{0.0}{0.0} & \val{0.0}{0.0} & \val{1.6}{0.7} & \val{0.0}{0.0} & \val{0.8}{0.3} & \val{4.6}{0.9} & \val{5.2}{1.2} & \val{\textbf{16.8}}{1.3} \\
task5-v0 & \val{0.0}{0.0} & \val{0.4}{0.3} & \val{1.9}{0.6} & \val{5.4}{2.5} & \val{8.9}{1.0} & \val{16.6}{3.1} & \val{\textbf{48.7}}{4.3} & \val{34.6}{3.9} \\

\grouprownine{cube-triple-play-singletask}
task1-v0 & \val{0.2}{0.1} & \val{2.7}{1.0} & \val{10.8}{1.8} & \val{2.1}{1.0} & \val{1.2}{0.6} & \val{5.7}{2.1} & \val{9.0}{1.6} & \val{\textbf{17.9}}{0.8} \\
task2-v0 & \val{0.0}{0.0} & \val{0.0}{0.0} & \val{0.0}{0.0} & \val{0.0}{0.0} & \val{0.0}{0.0} & \val{0.0}{0.0} & \val{0.0}{0.0} & \val{0.0}{0.0} \\
task3-v0 & \val{0.0}{0.0} & \val{0.0}{0.0} & \val{\textbf{0.3}}{0.1} & \val{0.0}{0.0} & \val{0.0}{0.0} & \val{0.0}{0.0} & \val{0.0}{0.0} & \val{0.0}{0.0} \\
task4-v0 & \val{0.0}{0.0} & \val{0.0}{0.0} & \val{0.0}{0.0} & \val{0.0}{0.0} & \val{0.0}{0.0} & \val{0.0}{0.0} & \val{0.0}{0.0} & \val{0.0}{0.0} \\
task5-v0 & \val{0.0}{0.0} & \val{0.0}{0.0} & \val{0.0}{0.0} & \val{0.0}{0.0} & \val{0.0}{0.0} & \val{0.0}{0.0} & \val{0.0}{0.0} & \val{0.0}{0.0} \\

\bottomrule
\end{tabularx}
\end{table}

\newpage
\begin{table}[H]
\centering
\caption{\textbf{OGBench results 2/2.} We follow the same evaluation protocol as Table~\ref{tab:ogbench_table_part1}. Dashes denote unreported entries.}
\label{tab:ogbench_table_part2}
\scriptsize
\setlength{\tabcolsep}{2pt}
\renewcommand{\arraystretch}{1.05}
\begin{tabularx}{\textwidth}{@{}>{\raggedright\arraybackslash}p{1.5cm}*{7}{>{\centering\arraybackslash}X}!{\hskip 4pt\vrule\hskip 2pt\vrule\hskip 4pt}>{\centering\arraybackslash}X@{\hspace{6pt}}}
\toprule
\textbf{Task} & BC & IQL & ReBRAC & IDQL & IFQL & FQL & GFP & \textbf{\textcolor[rgb]{0.05,0.35,0.75}{OptiFlow}} \\
\midrule
\grouprownine{puzzle-3x3-play-singletask}
task1-v0 & \val{19.1}{2.9} & \val{12.1}{2.4} & \val{\textbf{96.9}}{0.9} & \val{69.9}{3.7} & \val{92}{1.3} & \val{90.4}{1.4} & \val{\textbf{96.1}}{1.2} & \val{\textbf{95.3}}{1.5} \\
task2-v0 & \val{0.1}{0.1} & \val{3.3}{0.6} & \val{1.8}{0.9} & \val{0.0}{0.0} & \val{2.4}{1.1} & \val{\textbf{16.9}}{2.3} & \val{0.2}{0.1} & \val{0.0}{0.0} \\
task3-v0 & \val{0.0}{0.0} & \val{1.4}{0.4} & \val{1.9}{0.6} & \val{0.0}{0.0} & \val{1}{0.3} & \val{\textbf{12.0}}{1.3} & \val{0.7}{0.2} & \val{0.0}{0.0} \\
task4-v0 & \val{0.0}{0.0} & \val{2.5}{0.5} & \val{5.0}{1.8} & \val{0.0}{0.0} & \val{0.1}{0.1} & \val{11.1}{1.8} & \val{5.3}{1.2} & \val{\textbf{25.3}}{6.8} \\
task5-v0 & \val{0.3}{0.3} & \val{2.3}{0.7} & \val{6.1}{1.3} & \val{0.0}{0.0} & \val{0.9}{0.3} & \val{17.3}{1.8} & \val{\textbf{19.6}}{5.3} & \val{\textbf{20.1}}{8.9} \\
\grouprownine{puzzle-4x4-play-singletask}
task1-v0 & \val{0.0}{0.0} & \val{3.3}{0.9} & \val{23.8}{1.1} & \val{26.1}{2.5} & \val{43.9}{2.5} & \val{28.9}{2.7} & \val{\textbf{53.3}}{3.1} & \val{\textbf{51.8}}{3.6} \\
task2-v0 & \val{0.0}{0.0} & \val{1.6}{0.4} & \val{10.0}{0.6} & \val{3.5}{1.4} & \val{\textbf{16.9}}{3.5} & \val{13.3}{1.3} & \val{7.2}{1.2} & \val{11.6}{3.7} \\
task3-v0 & \val{0.0}{0.0} & \val{1.3}{0.3} & \val{16.8}{1.1} & \val{42.9}{3.2} & \val{40}{2.2} & \val{16.3}{1.6} & \val{46.0}{3.9} & \val{\textbf{63.1}}{3.6} \\
task4-v0 & \val{0.0}{0.0} & \val{1.0}{0.5} & \val{11.0}{1.6} & \val{\textbf{33.3}}{4.2} & \val{22.9}{2.5} & \val{8.8}{1.1} & \val{16.2}{1.7} & \val{22.8}{2.9} \\
task5-v0 & \val{0.0}{0.0} & \val{1.3}{0.4} & \val{8.6}{0.7} & \val{4.5}{0.4} & \val{10.5}{1.1} & \val{7.8}{1.2} & \val{6.3}{1.2} & \val{\textbf{11.1}}{2.7} \\
\grouprownine{scene-play-singletask}
task1-v0 & \val{1.8}{0.7} & \val{65}{2.6} & \val{\textbf{95.6}}{1.4} & \val{\textbf{99.8}}{0.2} & \val{\textbf{99.4}}{0.4} & \val{\textbf{100}}{0} & \val{\textbf{99.6}}{0.2} & \val{\textbf{100}}{0} \\
task2-v0 & \val{0.0}{0.0} & \val{10.9}{1.9} & \val{46.1}{5.5} & \val{0.3}{0.2} & \val{43.4}{8.0} & \val{78.1}{6.4} & \val{91.0}{2.9} & \val{\textbf{97.5}}{1.1} \\
task3-v0 & \val{0.0}{0.0} & \val{23}{2.1} & \val{67.5}{5.2} & \val{\textbf{95.3}}{1.3} & \val{80.8}{2.4} & \val{\textbf{95.1}}{0.8} & \val{75.0}{7.6} & \val{\textbf{98.0}}{0.6} \\
task4-v0 & \val{0.0}{0.0} & \val{1.1}{0.5} & \val{0.9}{0.4} & \val{0.0}{0.0} & \val{0.1}{0.1} & \val{\textbf{6.0}}{2.3} & \val{0.0}{0.0} & \val{1.5}{0.8} \\
task5-v0 & \val{0.0}{0.0} & \val{0.0}{0.0} & \val{0.0}{0.0} & \val{0.0}{0.0} & \val{0.0}{0.0} & \val{0.0}{0.0} & \val{0.0}{0.0} & \val{0.0}{0.0} \\
\grouprownine{visual-\{--\}-play-singletask-task1-v0}
cube-single & - & \itres{\val{70}{6}} & {\val{\textbf{\textit{83}}}{3}} & - & \itres{\val{49}{3.5}} & \itres{\val{81}{6}} & \val{72.2}{3.6} & \val{\textbf{86.6}}{1.3} \\
cube-double & - & \itres{\val{34}{11.5}} & \itres{\val{4}{2}} & - & \itres{\val{8}{3}} & \itres{\val{21}{5.5}} & \val{12.0}{2.6} & \val{\textbf{66.7}}{3.5} \\
scene & - & {\val{\textbf{\textit{96}}}{1}} & {\val{\textbf{\textit{98}}}{2}} & - & \itres{\val{86}{5}} & {\val{\textbf{\textit{98}}}{1.5}} & \val{\textbf{98.7}}{0.3} & \val{92.8}{0.8} \\
puzzle-3x3 & - & \itres{\val{7}{7.5}} & \itres{\val{88}{2}} & - & {\val{\textbf{\textit{100}}}{0}} & \itres{\val{94}{0.5}} & \val{86.5}{1.1} & \val{93.8}{1.2} \\
puzzle-4x4 & - & \itres{\val{0.0}{0.0}} & \itres{\val{26}{3}} & - & \itres{\val{8}{7.5}} & \itres{\val{33}{3}} & \val{19.3}{1.4} & \val{\textbf{39.7}}{4.8} \\
\bottomrule
\end{tabularx}
\end{table}

\begin{table}[H]
\centering
\caption{\textbf{D4RL results} across AntMaze navigation and Adroit dexterous manipulation tasks. All methods are trained for 500K gradient steps and evaluated at the final step over 100 episodes. We report scores as mean $\pm$ standard error across 8 random seeds, following D4RL conventions: success rate for AntMaze and normalized return for Adroit. \textbf{Bold} indicates results within 95\% of the best score on each task. \textit{Italics} indicate results taken from prior work~\citep{tarasov2023revisiting, hansenestruch2023idql}.}
\label{tab:d4rl_table}
\scriptsize
\setlength{\tabcolsep}{2pt}
\renewcommand{\arraystretch}{1.05}
\begin{tabularx}{\textwidth}{@{}>{\raggedright\arraybackslash}p{2.8cm}*{7}{>{\centering\arraybackslash}X}!{\hskip 4pt\vrule\hskip 2pt\vrule\hskip 4pt}>{\centering\arraybackslash}X@{\hspace{6pt}}}
\toprule
\textbf{Task} & BC & IQL & ReBRAC & IDQL & IFQL & FQL & GFP & \textbf{\textcolor[rgb]{0.05,0.35,0.75}{OptiFlow}} \\
\midrule
antmaze-umaze & \itres{54.6} & \itres{83.3} & \textit{\textbf{97.8}} & \textit{\textbf{94.0}} & \val{86.5}{2.1} & \val{\textbf{95.0}}{1.8} & \val{\textbf{96.5}}{0.5} & \val{\textbf{96.0}}{0.9} \\
antmaze-umaze-diverse & \itres{45.6} & \itres{70.6} & \textit{\textbf{88.3}} & \itres{80.2} & \val{58.5}{5.3} & \val{83.3}{4.8} & \val{\textbf{91.8}}{1.6} & \val{84.9}{1.6} \\
antmaze-medium-play & \itres{0.0} & \itres{64.6} & \textit{\textbf{84.0}} & \textit{\textbf{84.2}} & \val{54.8}{6.5} & \val{76.5}{2.7} & \val{83.2}{3.2} & \val{\textbf{88.0}}{1.4} \\
antmaze-medium-diverse & \itres{0.0} & \itres{61.7} & \itres{76.3} & \textit{\textbf{84.8}} & \val{60.3}{8.8} & \val{70.0}{4.2} & \val{57.5}{8.2} & \val{\textbf{87.4}}{1.6} \\
antmaze-large-play & \itres{0.0} & \itres{42.5} & \itres{60.4} & \itres{63.5} & \val{46.3}{4.8} & \val{\textbf{87.0}}{1.7} & \val{82.1}{1.8} & \val{82.6}{2.4} \\
antmaze-large-diverse & \itres{0.0} & \itres{27.6} & \itres{54.4} & \itres{67.9} & \val{59.5}{7.4} & \val{\textbf{81.75}}{1.4} & \val{\textbf{84.5}}{1.2} & \val{\textbf{80.6}}{1.8} \\
\midrule
pen-human-v1 & \itres{34.4} & \itres{81.5} & \textit{\textbf{103.5}} & \val{56.1}{2.3} & \val{69.8}{2.4} & \val{40.0}{3.5} & \val{59.1}{3.6} & \val{61.1}{2.3} \\
pen-cloned-v1 & \itres{56.9} & \itres{77.2} & \textit{\textbf{91.8}} & \val{74.2}{1.9} & \val{76.9}{2.8} & \val{74.9}{4.5} & \val{78.7}{2.7} & \val{67.4}{2.5} \\
pen-expert-v1 & \itres{85.1} & \itres{133.6} & \textit{\textbf{154.1}} & \val{133.7}{1.2} & \val{135.7}{1.9} & \val{144.8}{1.2} & \val{141.2}{2.0} & \val{127.6}{2.2} \\
door-human-v1 & \itres{0.5} & \itres{3.1} & \itres{0.0} & \val{\textbf{3.7}}{0.6} & \val{2.9}{0.6} & \val{0.1}{0.1} & \val{0.1}{0.0} & \val{2.0}{0.4} \\
door-cloned-v1 & \itres{-0.1} & \itres{0.8} & \itres{1.1} & \val{\textbf{6.1}}{1.3} & \val{0.2}{0.2} & \val{1.7}{0.5} & \val{0.3}{0.2} & \val{3.0}{1.6} \\
door-expert-v1 & \itres{34.9} & \textit{\textbf{105.3}} & \textit{\textbf{104.6}} & \val{\textbf{103.9}}{0.3} & \val{97.9}{2.6} & \val{\textbf{104.0}}{0.3} & \val{\textbf{104.2}}{0.2} & \val{\textbf{104.3}}{0.4} \\
hammer-human-v1 & \itres{1.5} & \itres{2.5} & \itres{0.2} & \val{1.4}{0.3} & \val{1.5}{0.3} & \val{2.3}{0.6} & \val{\textbf{5.7}}{2.2} & \val{5.1}{1.0} \\
hammer-cloned-v1 & \itres{0.8} & \itres{1.1} & \itres{6.7} & \val{1.8}{0.3} & \val{0.2}{0.2} & \val{\textbf{7.5}}{2.7} & \val{\textbf{7.4}}{1.5} & \val{2.1}{0.5} \\
hammer-expert-v1 & \itres{125.6} & \textit{\textbf{129.6}} & \textit{\textbf{133.8}} & \val{113.7}{3.4} & \val{97.9}{2.6} & \val{125.7}{0.4} & \val{122.4}{1.2} & \val{\textbf{128.8}}{0.2} \\
relocate-human-v1 & \itres{0.0} & \itres{0.1} & \itres{0.0} & \val{0.0}{0.0} & \val{0.2}{0.1} & \val{0.0}{0.0} & \val{\textbf{0.6}}{0.2} & \val{0.0}{0.0} \\
relocate-cloned-v1 & \itres{-0.1} & \itres{0.2} & \itres{0.9} & \val{0.0}{0.0} & \val{-0.0}{0.0} & \val{0.2}{0.1} & \val{\textbf{1.5}}{0.3} & \val{0.0}{0.0} \\
relocate-expert-v1 & \itres{101.3} & \textit{\textbf{106.5}} & \textit{\textbf{106.6}} & \val{\textbf{107.5}}{0.3} & \val{95.6}{2.8} & \val{\textbf{106.8}}{0.8} & \val{\textbf{104.2}}{2.1} & \val{\textbf{104.4}}{1.3} \\
\bottomrule
\end{tabularx}
\end{table}

\newpage
\section{Ablation Studies}
\label{app:appendix}


We conduct a comprehensive set of ablation studies to validate the design choices and probe the behavior of OptiFlow. We first analyze sensitivity to the task-specific temperatures $\tau$ and $\eta$, as well as to the Sinkhorn regularization coefficient and iteration count. We then study the effect of the reference and one-step action sample counts $(M,N)$ and show that the gains of OptiFlow cannot be explained by a larger action-sampling budget alone. Next, we disentangle the contributions of reference-policy quality and transport-based policy extraction, and compare our default value-aware behavior cloning objective with an AWR-style weighting alternative. Finally, we examine whether direct Q-maximization provides additional benefits beyond transport-based supervision and evaluate OptiFlow under offline-to-online fine-tuning.

\subsection{Sensitivity to value-weighting temperatures}
\label{app:sensitivity}
OptiFlow introduces two task-specific hyperparameters: the reference-marginal
softmax temperature $\tau$, which controls how sharply the transport plan is
biased toward high-value reference actions, and the value-aware BC temperature
$\eta$, which controls the strength of value weighting in reference policy training.

\begin{figure}[H]
\centering

\begin{minipage}[t]{0.31\textwidth}
\centering
\begin{tikzpicture}[x=0.8cm,y=0.8cm]
\foreach \x/\y/\v/\e in {
    1/3/92/2, 2/3/92/1, 3/3/94/1,
    1/2/91/2, 2/2/94/1, 3/2/84/3,
    1/1/93/1, 2/1/88/2, 3/1/85/2
}{
    \scorecell{\x}{\y}{\v}{\e}
}
\OptiFlowaxes
\end{tikzpicture}

\vspace{0.35em}
\parbox{\linewidth}{\centering\small\bfseries (a) OptiFlow antmaze-large}
\end{minipage}
\hfill
\begin{minipage}[t]{0.31\textwidth}
\centering
\begin{tikzpicture}[x=0.8cm,y=0.8cm]
\foreach \x/\y/\v/\e in {
    1/3/33/3, 2/3/43/3, 3/3/40/3,
    1/2/56/3, 2/2/52/3, 3/2/45/3,
    1/1/56/5, 2/1/45/3, 3/1/37/4
}{
    \scorecell{\x}{\y}{\v}{\e}
}
\OptiFlowaxes
\end{tikzpicture}

\vspace{0.35em}
\parbox{\linewidth}{\centering\small\bfseries (b) OptiFlow cube-double}
\end{minipage}
\hfill
\begin{minipage}[t]{0.31\textwidth}
\centering
\begin{tikzpicture}[x=0.8cm,y=0.8cm]
\foreach \x/\y/\v/\e in {
    1/3/42/10, 2/3/62/10, 3/3/88/7,
    1/2/90/3, 2/2/98/1, 3/2/97/2,
    1/1/95/1, 2/1/98/1, 3/1/93/2
}{
    \scorecell{\x}{\y}{\v}{\e}
}
\OptiFlowaxes
\end{tikzpicture}

\vspace{0.35em}
\parbox{\linewidth}{\centering\small\bfseries (c) OptiFlow scene}
\end{minipage}

\vspace{0.8em}
\caption{
\textbf{Sensitivity analysis to the reference-marginal temperature $\tau$, and to the value-aware BC temperature $\eta$.}
We evaluate the sensitivity by testing variations around task-specific hyperparameters $(\tau^*, \eta^*)$ from Table~\ref{tab:hparams_ogbench}, using 8 random seeds.
Figures (a), (b), and (c) report OptiFlow's sensitivity to $\tau^*$ and $\eta^*$ on \textit{antmaze-large-navigate-task1}, \textit{cube-double-play-task2}, and \textit{scene-play-task2}, respectively.
}
\label{fig:tau_eta_sensitivity}
\end{figure}

Result figures show that OptiFlow's performance varies smoothly around the tuned settings $(\tau^*, \eta^*)$, with no single hyperparameter dominating the result. It is simple to find appropriate task-specific $(\tau^*, \eta^*)$ values via minor tuning with an understanding of how these parameters affect the learning curve depending on the difficulty and the nature of tasks.

\subsection{Sensitivity to Sinkhorn hyperparameters}
\label{app:sinkhorn_param_abal}
\textcolor{black}{
We normalize each per-state cost matrix by its mean (Eq.~\ref{eq:cost}), so the entropic regularization coefficient $\epsilon$ has a comparable scale across tasks. We use $\epsilon=0.05$ and 30 Sinkhorn iterations by default, and sweep both choices on the three tasks from Figure~\ref{fig:tau_eta_sensitivity}.
}

\begin{table}[H]
\centering
\begin{minipage}[t]{0.485\textwidth}
\centering
\small\textbf{(a) Regularization coefficient $\epsilon$}\par\vspace{3pt}
\scriptsize
\setlength{\tabcolsep}{2.5pt}
\renewcommand{\arraystretch}{1.1}
\begin{tabular}{@{}lcccc@{}}
\toprule
\textbf{Task} & .01 & \textbf{.05} & .1 & .5 \\
\midrule
Antmaze & \val{91.6}{2.7} & \val{\textbf{94.0}}{0.9} & \val{93.8}{1.2} & \val{93.9}{0.9} \\
Cube & \val{\textbf{62.1}}{3.7} & \val{52.4}{2.9} & \val{57.0}{5.0} & \val{49.9}{2.9} \\
Scene & \val{93.7}{2.4} & \val{\textbf{97.5}}{1.1} & \val{96.0}{1.2} & \val{82.9}{3.6} \\
\bottomrule
\end{tabular}
\end{minipage}\hfill
\begin{minipage}[t]{0.485\textwidth}
\centering
\small\textbf{(b) Sinkhorn iterations}\par\vspace{3pt}
\scriptsize
\setlength{\tabcolsep}{3pt}
\renewcommand{\arraystretch}{1.1}
\begin{tabular}{@{}lcccc@{}}
\toprule
\textbf{Task} & 10 & 20 & \textbf{30} & 40 \\
\midrule
Antmaze & \val{91.8}{1.9} & \val{92.9}{1.9} & \val{\textbf{94.0}}{0.9} & \val{93.3}{1.9} \\
Cube & \val{50.3}{3.7} & \val{48.9}{3.2} & \val{\textbf{52.4}}{2.9} & \val{51.8}{3.9} \\
Scene & \val{96.4}{1.5} & \val{94.8}{2.2} & \val{\textbf{97.5}}{1.1} & \val{95.9}{1.4} \\
\bottomrule
\end{tabular}
\end{minipage}
\caption{\textbf{Sensitivity to Sinkhorn hyperparameters.} Antmaze, Cube, and Scene denote antmaze-large-navigate-task1, cube-double-play-task2, and scene-play-task2. Evaluation protocol follows Table~\ref{tab:ogbench_table_part1}.}
\label{tab:sinkhorn_param_abal}
\end{table}

\textcolor{black}{
The sweeps show a mild task-dependent trade-off. The fixed $\epsilon=0.05$ is a robust intermediate point on the normalized cost scale, while 30 iterations performs best on all three tasks and adds little overhead (Appendix~\ref{app:compute}).
}

\subsection{Reference and one-step action sampling}
\label{app:sampling_abal}

The transport plan in OptiFlow is constructed over $N$ one-step actions and $M$ reference actions sampled per state, and the quality of the resulting supervision depends on whether these sample sets adequately cover the conditional action distribution. We sweep $(N, M)$ over $\{4, 8, 16, 32\} \times \{4, 16, 64, 256\}$, holding all other hyperparameters fixed, to identify how many samples are needed for stable transport-based distillation and show that we only need an adequate number of samples.

\begin{table}[H]
\centering
\caption{\textbf{Antmaze-large-navigate-task1.}
We follow the same evaluation protocol as Table~\ref{tab:ogbench_table_part1}, varying one-step samples $N$
and reference samples $M$.}
\label{tab:nm_antmaze_large}
\normalsize
\setlength{\tabcolsep}{18pt}
\renewcommand{\arraystretch}{1.3}
\begin{tabular}{@{}c|cccc@{}}
\toprule
$N \backslash M$ & 4 & 16 & \textbf{64} (default) & 256 \\
\midrule
4  & \val{2.8}{0.8} & \val{80}{2.3} & \val{92.2}{2.2} & \val{\textbf{97.2}}{0.3} \\
8  & \val{7.8}{2.8} & \val{38.7}{4.0} & \val{90.7}{0.7} & \val{\textbf{96.7}}{0.6} \\
\textbf{16} (default) & \val{18.5}{4.3} & \val{39.8}{3.3} & \val{\textbf{94.0}}{0.9} & \val{\textbf{94.2}}{0.6} \\
32 & \val{4.8}{0.7} & \val{29.8}{3.1} & \val{81.8}{3.0} & \val{83.0}{2.2} \\
\bottomrule
\end{tabular}
\end{table}

It can seen that performance depends mainly on \(M\) and is largely insensitive to \(N\). The default \((N,M)=(16,64)\) matches or achieves the best result on each task: smaller \(M\) degrades performance sharply, while \(M=256\) gives consistent gain as can be expected. Sampling larger number of reference actions may be advantageous for distill target actions selection. Due to performance vs computation trade off, we use \((N,M)=(16,64)\) by default, concluding that OptiFlow only needs sufficient number of one-step and reference action samples, and excessive number of samples are not critical.

\textcolor{black}{
\subsection{Controlling for the action-sampling budget}
\label{app:matched_sampling}
OptiFlow samples multiple actions per state to construct the transport coupling, raising the possibility that its gains arise simply from a larger action-sampling budget rather than from transport-based supervision.
To control for this factor, we modify FQL and GFP to draw either $N=16$ or $N=64$ one-step actions per state and average their actor gradients over the sampled actions, while leaving the remaining training procedure unchanged.
}
\begin{table}[H]
\centering
\caption{\textbf{Controlling for the action-sampling budget.}
FQL and GFP with $N=16$ or $N=64$ draw multiple actions per state and average
the actor gradient over the sampled actions. OptiFlow uses $N=16$ one-step
policy samples per state. We follow the same evaluation protocol as
Table~\ref{tab:ogbench_table_part1}.}
\label{tab:matched_sampling}
\scriptsize
\setlength{\tabcolsep}{2.5pt}
\renewcommand{\arraystretch}{1.15}

\begin{tabularx}{\textwidth}{
@{}>{\raggedright\arraybackslash}p{3.6cm}
*{7}{>{\centering\arraybackslash}X}@{}
}
\toprule
\textbf{Task}
& \shortstack{\textbf{FQL}\\\textbf{(original)}}
& \shortstack{\textbf{FQL}\\\textbf{($N=16$)}}
& \shortstack{\textbf{FQL}\\\textbf{($N=64$)}}
& \shortstack{\textbf{GFP}\\\textbf{(original)}}
& \shortstack{\textbf{GFP}\\\textbf{($N=16$)}}
& \shortstack{\textbf{GFP}\\\textbf{($N=64$)}}
& \textbf{\textcolor[rgb]{0.05,0.35,0.75}{OptiFlow}} \\
\midrule

antmaze-large-navigate\{1\}
& \val{79.5}{2.1} & \val{79.5}{1.8} & \val{76.3}{4.0}
& \val{\textbf{94.5}}{1.4} & \val{\textbf{95.2}}{0.6} & \val{\textbf{94.4}}{0.7}
& \val{\textbf{94.0}}{0.9} \\

antmaze-giant-navigate\{1\}
& \val{10.5}{2.9} & \val{8.0}{2.5} & \val{12.3}{3.5}
& \val{8.6}{2.5} & \val{{15.9}}{7.0} & \val{\textbf{17.5}}{6.8}
& \val{5.4}{2.4} \\

humanoidmaze-medium-navigate\{1\}
& \val{19.0}{4.2} & \val{25.0}{10.6} & \val{10.3}{4.1}
& \val{\textbf{84.8}}{2.1} & \val{\textbf{83.0}}{1.8} & \val{\textbf{84.7}}{1.0}
& \val{\textbf{82.4}}{1.2} \\

humanoidmaze-large-navigate\{1\}
& \val{1.0}{0.5} & \val{2.0}{0.7} & \val{3.3}{1.8}
& \val{36.0}{5.8} & \val{22.5}{7.4} & \val{17.5}{7.4}
& \val{\textbf{48.3}}{2.2} \\

antsoccer-arena-navigate\{4\}
& \val{38.1}{4.4} & \val{34.8}{1.7} & \val{43.5}{2.4}
& \val{42.5}{3.8} & \val{42.5}{1.8} & \val{\textbf{44.6}}{1.4}
& \val{\textbf
{46.8}}{4.1} \\

cube-single-play\{2\}
& \val{\textbf{96.1}}{0.6} & \val{\textbf{97.8}}{1.0} & \val{\textbf{98.0}}{0.8}
& \val{\textbf{97.7}}{1.1} & \val{\textbf{99.3}}{0.2} & \val{\textbf{98.8}}{0.3}
& \val{\textbf{98.5}}{0.5} \\

cube-double-play\{2\}
& \val{33.8}{5.1} & \val{36.0}{5.2} & \val{36.3}{4.0}
& \val{48.3}{3.7} & \val{40.8}{2.0} & \val{41.8}{1.9}
& \val{\textbf{52.4}}{2.9} \\

scene-play\{2\}
& \val{78.1}{6.4} & \val{79.8}{3.5} & \val{78.8}{5.6}
& \val{91.0}{2.9} & \val{90.6}{2.2} & \val{89.7}{1.5}
& \val{\textbf{97.5}}{1.1} \\

puzzle-3x3-play\{4\}
& \val{11.1}{1.8} & \val{12.5}{1.2} & \val{13.3}{1.9}
& \val{5.3}{1.2} & \val{6.2}{0.8} & \val{7.9}{1.6}
& \val{\textbf{25.3}}{6.8} \\

puzzle-4x4-play\{4\}
& \val{8.8}{1.1} & \val{9.8}{1.2} & \val{9.3}{1.2}
& \val{16.2}{1.7} & \val{17.1}{1.4} & \val{15.4}{1.2}
& \val{\textbf{22.8}}{2.9} \\

\midrule
\textbf{Average}
& \val{37.6}{1.1} & \val{38.5}{1.5} & \val{38.1}{1.3}
& \val{52.5}{0.9} & \val{51.3}{0.8} & \val{51.2}{1.2}
& \val{\textbf{57.3}}{1.0} \\

\bottomrule
\end{tabularx}
\end{table}

\textcolor{black}{
Increasing the number of sampled actions does not yield a consistent improvement for either baseline. FQL obtains average scores of $37.6$,
$38.5$, and $38.1$ with $N=1$, $16$, and $64$, respectively, while GFP
obtains $52.5$, $51.3$, and $51.2$. All variants remain below OptiFlow's $57.3$. Notably, the $N=64$ baseline variants use more one-step action
samples per state than OptiFlow ($N=16$), yet do not close the performance gap. Thus, the performance improvement cannot be explained by the number of one-step actions sampled during training alone.
}

\textcolor{black}{
\subsection{Disentangling reference policy quality and distillation}
\label{app:disentangle_vabc_OT}
To disentangle whether OptiFlow’s performance gains arise from a stronger reference policy or from the distillation procedure itself, we separately evaluate the reference and one-step policies and introduce OptiFlow-BC, which replaces the VaBC reference policy with a standard BC reference while keeping the OT distillation procedure unchanged.
}

\begin{table}[H]
\centering
\caption{\textbf{Reference and one-step policy performance.}
We report the performance of the reference (Ref.) and distilled one-step policies for FQL, OptiFlow-BC, GFP, and OptiFlow. We follow the same evaluation protocol as Table~\ref{tab:ogbench_table_part1}.}
\label{tab:reference_policy_performance}

\fontsize{7.2}{8.3}\selectfont
\setlength{\tabcolsep}{1.2pt}
\renewcommand{\arraystretch}{1.15}

\begin{tabularx}{\textwidth}{
@{}>{\raggedright\arraybackslash}p{4.1cm}
*{8}{>{\centering\arraybackslash}X}@{}
}
\toprule
& \multicolumn{2}{c}{\textbf{FQL}}
& \multicolumn{2}{c}{\textbf{OptiFlow-BC}}
& \multicolumn{2}{c}{\textbf{GFP}}
& \multicolumn{2}{c}{\textbf{\textcolor[rgb]{0.05,0.35,0.75}{OptiFlow}}} \\
\cmidrule(lr){2-3}
\cmidrule(lr){4-5}
\cmidrule(lr){6-7}
\cmidrule(lr){8-9}

\textbf{Task}
& \textbf{BC Ref.} & \textbf{one-step}
& \textbf{BC Ref.} & \textbf{one-step}
& \textbf{VaBC Ref.} & \textbf{one-step}
& \textbf{VaBC Ref.} & \textbf{one-step} \\
\midrule

{\fontsize{6.5}{7.2}\selectfont antmaze-large-navigate\{1\}}
& \val{1.5}{1.0} & \val{79.5}{2.1}
& \val{1.1}{0.5} & \val{\textbf{91.0}}{2.0}
& \val{88.5}{2.3} & \val{\textbf{94.5}}{1.4}
& \val{18.4}{3.9} & \val{\textbf{94.0}}{0.9} \\

{\fontsize{6.5}{7.2}\selectfont antmaze-giant-navigate\{1\}}
& \val{0.0}{0.0} & \val{\textbf{10.5}}{2.9}
& \val{0.0}{0.0} & \val{0.9}{0.3}
& \val{0.3}{0.2} & \val{8.6}{2.5}
& \val{0.0}{0.0} & \val{5.4}{2.4} \\

{\fontsize{6.5}{7.2}\selectfont humanoidmaze-medium-navigate\{1\}}
& \val{0.5}{0.3} & \val{19.0}{4.2}
& \val{0.8}{0.3} & \val{50.5}{8.3}
& \val{28.8}{2.4} & \val{\textbf{84.8}}{2.1}
& \val{46.1}{1.1} & \val{\textbf{82.4}}{1.2} \\

{\fontsize{6.5}{7.2}\selectfont humanoidmaze-large-navigate\{1\}}
& \val{0.0}{0.0} & \val{1.0}{0.5}
& \val{0.0}{0.0} & \val{43.4}{4.1}
& \val{2.1}{1.1} & \val{36.0}{5.8}
& \val{0.0}{0.0} & \val{\textbf{48.3}}{2.2} \\

{\fontsize{6.5}{7.2}\selectfont antsoccer-arena-navigate\{4\}}
& \val{1.4}{0.2} & \val{38.1}{4.4}
& \val{1.4}{0.5} & \val{43.6}{3.4}
& \val{7.8}{1.6} & \val{42.5}{3.8}
& \val{11.8}{2.2} & \val{\textbf{46.8}}{4.1} \\

{\fontsize{6.5}{7.2}\selectfont cube-single-play\{2\}}
& \val{10.4}{2.6} & \val{\textbf{96.1}}{0.6}
& \val{8.6}{1.5} & \val{\textbf{97.5}}{1.3}
& \val{44.4}{6.0} & \val{\textbf{97.7}}{1.1}
& \val{45.1}{7.1} & \val{\textbf{98.5}}{0.5} \\

{\fontsize{6.5}{7.2}\selectfont cube-double-play\{2\}}
& \val{0.3}{0.2} & \val{33.8}{5.1}
& \val{0.0}{0.0} & \val{39.1}{3.5}
& \val{2.3}{0.6} & \val{48.3}{3.7}
& \val{1.4}{0.4} & \val{\textbf{52.4}}{2.9} \\

{\fontsize{6.5}{7.2}\selectfont scene-play\{2\}}
& \val{2.4}{0.8} & \val{78.1}{6.4}
& \val{2.8}{0.9} & \val{\textbf{98.2}}{0.8}
& \val{94.0}{2.5} & \val{91.0}{2.9}
& \val{69.8}{3.7} & \val{\textbf{97.5}}{1.1} \\

{\fontsize{6.5}{7.2}\selectfont puzzle-3x3-play\{4\}}
& \val{0.4}{0.2} & \val{11.1}{1.8}
& \val{0.1}{0.1} & \val{1.1}{0.7}
& \val{10.3}{1.6} & \val{5.3}{1.2}
& \val{4.3}{0.5} & \val{\textbf{25.3}}{6.8} \\

{\fontsize{6.5}{7.2}\selectfont puzzle-4x4-play\{4\}}
& \val{0.3}{0.2} & \val{8.8}{1.1}
& \val{0.0}{0.0} & \val{20.9}{4.2}
& \val{5.1}{1.6} & \val{16.2}{1.7}
& \val{1.5}{0.3} & \val{\textbf{22.8}}{2.9} \\

\midrule

\textbf{Average}
& \val{1.7}{0.3} & \val{37.6}{1.1}
& \val{1.5}{0.2} & \val{48.6}{1.3}
& \val{28.4}{3.4} & \val{52.5}{0.9}
& \val{19.8}{2.8} & \val{\textbf{57.3}}{1.0} \\

\bottomrule
\end{tabularx}
\end{table}

\textcolor{black}{
As shown in the Table~\ref{tab:reference_policy_performance}, FQL and OptiFlow-BC show comparable reference performance, yet OptiFlow-BC yields a better one-step policy on 8 of 10 tasks, improving the average from 37.6 to 48.6. Since the reference policy is matched by construction, this indicates that a large part of OptiFlow’s gain comes from the distillation process itself. Comparing OptiFlow-BC with full OptiFlow demonstrates that the Value-aware BC reference policy and the OT distillation are complementary: replacing the BC reference policy with a VaBC one improves the one-step average further, from 48.6 to 57.3.    VaBC would become unnecessary only with a very large $M$, since the marginal can only select within the sampled pool. However, larget $M$ increases compute cost, so VaBC is what makes a moderate $M$ sufficient, by improving the quality of the candidate pool. 
}

\textcolor{black}{
Notably, OptiFlow’s reference policy is weaker than GFP’s on average (19.8 vs 28.4) while its one-step policy is stronger (57.3 vs 52.5), which further indicates that reference performance alone does not determine one-step performance. 
}

\textcolor{black}{
\subsection{AWR-style weighting for value-aware behavior cloning}
\label{app:awr}
Our reference flow policy is trained using the value-aware behavior cloning
(VaBC) objective in Eq.~\eqref{eq:ref_loss}, where the weight $g_\eta$
is implemented as a two-class softmax between the value of the dataset action
and that of the current one-step policy action. As an alternative, we consider
the AWR-style exponential weighting used in GFP~\citep{tiofack2026guided}.
Specifically, OptiFlow-AWR replaces the default weight $g_\eta$ with
\begin{align}
g_\eta^{\mathrm{AWR}}(s,a,z)
=
\exp\left(
\frac{\lambda}{\eta}
\left[
Q_\phi(s,a)
-
Q_\phi(s,\mu_\theta(s,z))
\right]
\right),
\end{align}
while leaving all other components of OptiFlow unchanged.
}
\begin{table}[H]
\centering
\caption{\textbf{AWR-style weighting for value-aware behavior cloning.}
OptiFlow-AWR replaces the default softmax weight $g_\eta$ in the
reference-policy objective with an AWR-style exponential advantage weight,
while leaving all other components unchanged. We follow the same evaluation
protocol as Table~\ref{tab:ogbench_table_part1}.}
\label{tab:awr_ablation}

\small
\setlength{\tabcolsep}{12pt}
\renewcommand{\arraystretch}{1.15}

\begin{tabular}{@{}lcc@{}}
\toprule
\textbf{Task}
& \shortstack{\textbf{\textcolor[rgb]{0.05,0.35,0.75}{OptiFlow}}\\
              \textbf{(default)}}
& \textbf{OptiFlow-AWR} \\
\midrule

antmaze-large-navigate\{1\}
& \val{94.0}{0.9} & \val{97.6}{0.8} \\

antmaze-giant-navigate\{1\}
& \val{5.4}{2.4} & \val{4.4}{1.3} \\

humanoidmaze-medium-navigate\{1\}
& \val{82.4}{1.2} & \val{88.6}{2.4} \\

humanoidmaze-large-navigate\{1\}
& \val{48.3}{2.2} & \val{34.2}{9.2} \\

antsoccer-arena-navigate\{4\}
& \val{46.8}{4.1} & \val{36.8}{3.9} \\

cube-single-play\{2\}
& \val{98.5}{0.5} & \val{98.8}{0.6} \\

cube-double-play\{2\}
& \val{52.4}{2.9} & \val{48.8}{6.3} \\

scene-play\{2\}
& \val{97.5}{1.1} & \val{93.4}{3.4} \\

puzzle-3x3-play\{4\}
& \val{25.3}{6.8} & \val{1.8}{1.8} \\

puzzle-4x4-play\{4\}
& \val{22.8}{2.9} & \val{20.6}{3.9} \\

\midrule
\textbf{Average}
& \val{\textbf{57.3}}{1.0}
& \val{52.5}{1.4} \\

\bottomrule
\end{tabular}
\end{table}
\textcolor{black}{
As shown in Table~\ref{tab:awr_ablation}, AWR-style weighting improves
performance on some tasks, including antmaze-large and humanoidmaze-medium,
but does not provide a consistent improvement across tasks. Averaged over the
10 evaluated tasks, OptiFlow-AWR achieves $52.5\pm1.4$, compared with
$57.3\pm1.0$ for the default weighting. We therefore retain the default
two-class softmax weighting for the reference-policy objective.
}

\subsection{Direct Q-maximization}
\label{app:direct_q_max}

A standard alternative to transport-based supervision is to optimize the
one-step policy directly against the critic. We add a Q-maximization term
weighted by $\lambda_q$ (default $\lambda_q = 0$, recovering pure OptiFlow) to
the actor objective and sweep its weight to test whether direct critic
guidance complements or substitutes for transport-based distillation.

\begin{table}[H]
\centering
\caption{\textbf{Direct Q-maximization ablation.} We follow the same evaluation protocol as Table~\ref{tab:ogbench_table_part1}. $\lambda_q=0$ is the OptiFlow default reported in our main results.}
\label{tab:qmax_ablation}
\renewcommand{\arraystretch}{1.5}
\resizebox{\textwidth}{!}{%
\begin{tabular}{@{}>
{\raggedright\arraybackslash}p{5.8cm}cccc@{}}
\toprule
\textbf{Task} & \makecell{$\boldsymbol{\lambda_q=0}$ \\ \textbf{(default)}} & $\lambda_q=10^{-3}$ & $\lambda_q=10^{-2}$ & $\lambda_q=10^{-1}$ \\
\midrule
humanoidmaze-medium-navigate-task1 & \val{\textbf{84.4}}{1.2} & \val{81.6}{1.7} & \val{81.4}{2.4} & \val{\textbf{88.8}}{1.2} \\
antsoccer-arena-navigate-task1     & \val{67.9}{2.4} & \val{42.0}{3.0} & \val{45.3}{2.1} & \val{\textbf{76.4}}{3.3} \\
cube-double-play-task2             & \val{\textbf{52.4}}{2.9} & \val{\textbf{54.3}}{3.3} & \val{\textbf{52.9}}{3.9} & \val{7.5}{1.2} \\
puzzle-4x4-play-task1              & \val{\textbf{51.8}}{3.6} & \val{39.4}{5.2} & \val{10.4}{2.1} & \val{0.9}{0.4} \\
\bottomrule
\end{tabular}%
}
\end{table}

The effect of direct critic maximization is mixed and task-dependent, likely because $\lambda_q$ interacts with $\tau$ and $\eta$, which control the sharpness of value weighting in the transport plan and reference-policy training. Although $\lambda_q=0.1$ improves humanoidmaze-medium and antsoccer-arena under our setting, it sharply degrades cube-double and puzzle-4$\times$4, and larger values can collapse performance. Jointly tuning $\lambda_q$ with $\tau$ and $\eta$ may improve some tasks, but it introduces another sensitive hyperparameter.

OptiFlow already achieves strong performance across diverse benchmarks with $\lambda_q=0$, showing that value-aware transport-based distillation is sufficient in our setting. Together with the diagnostics in Figure.~\ref{fig:bandit_toy} from the main experiment section, this supports using the critic primarily for transport-based target construction rather than direct one-step policy maximization. Accordingly, all main OptiFlow results use $\lambda_q=0$, isolating the effect of value-weighted OT distillation and avoiding an additional task-sensitive tuning parameter. This matches our design choice of avoiding direct critic maximization to address mode collapse and overestimation bias in out-of-distribution regions.

\section{Offline-to-Online Fine-tuning}
\label{app:online}
While OptiFlow is designed for the offline setting, the structure of OT coupling between the reference policy and one-step target policy is naturally compatible with continued learning from newly collected data. We initialize from an offline OptiFlow checkpoint at 1M training steps and continue
training with online interaction, retaining the same reference policy, transport-based target construction, and one-step policy distillation. This study examines whether OptiFlow can convert offline gains into further improvements
online without algorithmic modification.

\begin{table}[H]
\centering
\caption{\textbf{Task-specific OptiFlow hyperparameters for offline-to-online RL.} We individually tune these hyperparameters for each task. ``--'' indicates that the default values from Table~\ref{tab:OptiFlow_shared_hparams} were used.}
\label{tab:o2o_hparams}
\scriptsize
\setlength{\tabcolsep}{6pt}
\renewcommand{\arraystretch}{1.0}
\begin{tabular}{@{}>{\raggedright\arraybackslash}p{7.5cm}cc@{}}
\toprule
\textbf{Task} & \textbf{Q-agg} & $\boldsymbol{(\tau,\, \eta)}$ \\
\midrule
humanoidmaze-medium-navigate-singletask-task1-v0  & -- & -- \\
antsoccer-arena-navigate-singletask-task4-v0      & -- & -- \\
cube-double-play-singletask-task2-v0              & -- & -- \\
scene-play-singletask-task2-v0                    & -- & -- \\
puzzle-4x4-play-singletask-task4-v0               & -- & -- \\
\midrule
antmaze-umaze-v2          & -- & -- \\
antmaze-umaze-diverse-v2  & -- & -- \\
antmaze-medium-play-v2    & -- & -- \\
antmaze-medium-diverse-v2 & -- & -- \\
antmaze-large-play-v2     & -- & -- \\
antmaze-large-diverse-v2  & -- & -- \\
\midrule
pen-cloned-v1      & qmean & (1.0,1e-2) \\
door-cloned-v1     & qmean & (1.0,1e-5) \\
hammer-cloned-v1   & qmean & (2.0,1e-4) \\
relocate-cloned-v1 & qmean & (1.0,1e-4) \\
\bottomrule
\end{tabular}
\end{table}

\begin{table}[H]
\centering
\caption{\textbf{Offline-to-online RL results.} Each cell reports the offline $\to$ online performance, and we follow the same evaluation protocol as Table~\ref{tab:ogbench_table_part1}. \textbf{Bold} indicates online results within 95\% of the best online score on each task. \textit{Italics} indicate results taken from FQL~\citep{park2025flow}.}
\label{tab:o2o_table}
\setlength{\tabcolsep}{4pt}
\renewcommand{\arraystretch}{1.05}
\resizebox{\textwidth}{!}{%
\begin{tabular}{@{}lcccc!{\hskip 4pt\vrule\hskip 2pt\vrule\hskip 4pt}c@{}}
\toprule
\textbf{Task} & IQL & ReBRAC & IFQL & FQL & \textbf{\textcolor[rgb]{0.05,0.35,0.75}{OptiFlow}} \\
\midrule
\grouprowsix{OGBench}
humanoidmaze-medium-navigate\{1\} & \val{\mathit{21}}{\mathit{13}}~$\to$~\val{\mathit{16}}{\mathit{8}}   & \val{\mathit{16}}{\mathit{20}}~$\to$~\val{\mathit{1}}{\mathit{1}}    & \val{70.5}{2.3}~$\to$~\val{43.0}{3.4} & \val{18.1}{1.4}~$\to$~\val{24.6}{3.6} & \val{79.0}{2.0}~$\to$~\val{\textbf{97.1}}{0.7} \\
antsoccer-arena-navigate\{4\}     & \val{\mathit{2}}{\mathit{1}}~$\to$~\val{\mathit{0}}{\mathit{0}}      & \val{\mathit{0}}{\mathit{0}}~$\to$~\val{\mathit{0}}{\mathit{0}}      & \val{40.8}{3.5}~$\to$~\val{36.0}{5.3} & \val{33.8}{2.8}~$\to$~\val{\textbf{82.8}}{2.1} & \val{41.0}{5.1}~$\to$~\val{71.5}{3.5} \\
cube-double-play\{2\}             & \val{\mathit{0}}{\mathit{1}}~$\to$~\val{\mathit{0}}{\mathit{0}}      & \val{\mathit{6}}{\mathit{5}}~$\to$~\val{\mathit{28}}{\mathit{28}}    & \val{11.8}{1.7}~$\to$~\val{55.2}{5.1} & \val{35.1}{3.2}~$\to$~\val{\textbf{94.0}}{1.1} & \val{49.1}{5.1}~$\to$~\val{\textbf{94.9}}{1.3} \\
scene-play\{2\}                   & \val{\mathit{14}}{\mathit{11}}~$\to$~\val{\mathit{10}}{\mathit{9}}   & \val{\mathit{55}}{\mathit{10}}~$\to$~\val{\mathit{\textbf{100}}}{\mathit{0}}  & \val{29.0}{4.8}~$\to$~\val{94.2}{1.7} & \val{79.5}{4.2}~$\to$~\val{\textbf{99.8}}{0.2} & \val{94.8}{1.8}~$\to$~\val{\textbf{100.0}}{0.0} \\
puzzle-4x4-play\{4\}              & \val{\mathit{5}}{\mathit{2}}~$\to$~\val{\mathit{1}}{\mathit{1}}      & \val{\mathit{8}}{\mathit{4}}~$\to$~\val{\mathit{14}}{\mathit{35}}    & \val{23.5}{2.3}~$\to$~\val{25.0}{16.4} & \val{9.2}{1.0}~$\to$~\val{\textbf{100.0}}{0.0} & \val{13.0}{2.5}~$\to$~\val{64.1}{17.5} \\
\grouprowsix{AntMaze (D4RL)}
antmaze-umaze          & $\mathit{77 \to \mathbf{96}}$    & $\mathit{98 \to 75}$    & \val{91.5}{1.5}~$\to$~\val{\textbf{96.0}}{0.8} & \val{96.8}{0.6}~$\to$~\val{\textbf{99.5}}{0.3} & \val{95.8}{1.0}~$\to$~\val{\textbf{99.5}}{0.3} \\
antmaze-umaze-diverse  & $\mathit{60 \to 64}$    & $\mathit{74 \to \mathbf{98}}$    & \val{61.0}{9.8}~$\to$~\val{\textbf{95.2}}{1.0} & \val{89.0}{1.9}~$\to$~\val{\textbf{99.4}}{0.3} & \val{80.1}{2.3}~$\to$~\val{\textbf{99.0}}{0.4} \\
antmaze-medium-play    & $\mathit{72 \to 90}$    & $\mathit{88 \to \mathbf{98}}$    & \val{36.2}{9.8}~$\to$~\val{90.5}{1.9} & \val{75.8}{2.5}~$\to$~\val{\textbf{97.5}}{0.3} & \val{82.8}{2.0}~$\to$~\val{\textbf{98.1}}{0.5} \\
antmaze-medium-diverse & $\mathit{64 \to 92}$    & $\mathit{85 \to \mathbf{99}}$    & \val{59.5}{8.9}~$\to$~\val{92.5}{1.3} & \val{65.6}{5.7}~$\to$~\val{\textbf{96.0}}{0.7} & \val{76.5}{4.5}~$\to$~\val{\textbf{97.6}}{0.7} \\
antmaze-large-play     & $\mathit{38 \to 64}$    & $\mathit{68 \to 32}$    & \val{67.5}{2.5}~$\to$~\val{86.5}{1.6} & \val{84.0}{2.4}~$\to$~\val{\textbf{93.2}}{0.6} & \val{83.0}{1.6}~$\to$~\val{\textbf{94.2}}{0.6} \\
antmaze-large-diverse  & $\mathit{27 \to 64}$    & $\mathit{67 \to 72}$    & \val{66.5}{3.0}~$\to$~\val{83.8}{2.1} & \val{87.0}{1.9}~$\to$~\val{\textbf{95.1}}{0.7} & \val{83.0}{1.7}~$\to$~\val{\textbf{95.8}}{0.9} \\
\grouprowsix{Adroit (D4RL)}
pen-cloned      & $\mathit{84 \to 102}$    & $\mathit{74 \to \mathbf{138}}$    & \val{71.6}{2.6}~$\to$~\val{106.0}{3.4} & \val{54.4}{4.9}~$\to$~\val{124.0}{1.9} & \val{45.2}{2.7}~$\to$~\val{\textbf{132.0}}{1.9} \\
door-cloned     & $\mathit{1 \to 20}$      & $\mathit{0 \to \mathbf{102}}$     & \val{0.3}{0.2}~$\to$~\val{35.2}{3.1} & \val{0.3}{0.3}~$\to$~\val{72.7}{4.4} & \val{{-}0.1}{0.1}~$\to$~\val{78.0}{3.2} \\
hammer-cloned   & $\mathit{1 \to 57}$      & $\mathit{7 \to \mathbf{125}}$     & \val{2.0}{0.7}~$\to$~\val{52.1}{5.3} & \val{1.2}{0.5}~$\to$~\val{110.0}{9.8} & \val{2.2}{1.0}~$\to$~\val{97.6}{9.9} \\
relocate-cloned & $\mathit{0 \to 0}$       & $\mathit{1 \to 7}$       & \val{{-}0.1}{0.0}~$\to$~\val{3.2}{2.1} & \val{0.5}{0.2}~$\to$~\val{\textbf{73.6}}{3.8} & \val{{-}0.1}{0.0}~$\to$~\val{{-}0.0}{0.1} \\
\bottomrule
\end{tabular}%
}
\end{table}

\newpage
\begin{figure}[H]
\centering
\includegraphics[width=0.32\textwidth]{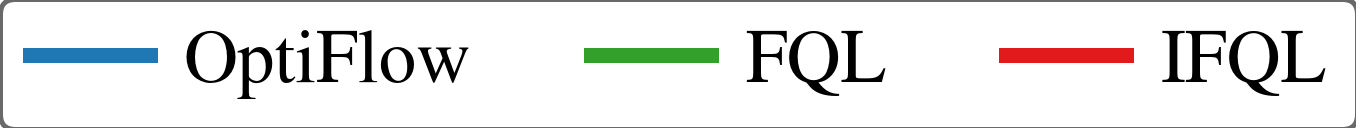}\\[4pt]
\includegraphics[width=0.195\textwidth]{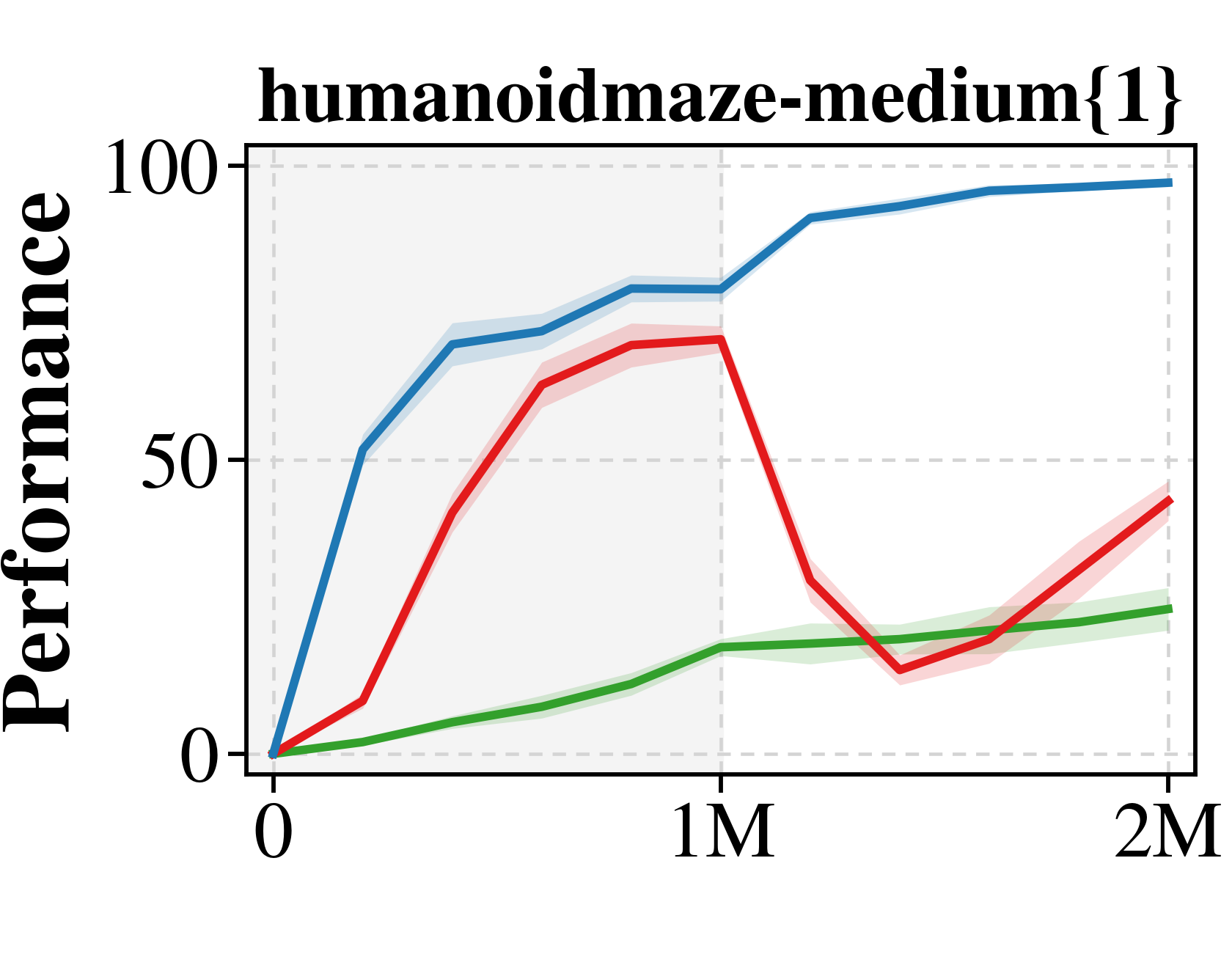}\hfill
\includegraphics[width=0.195\textwidth]{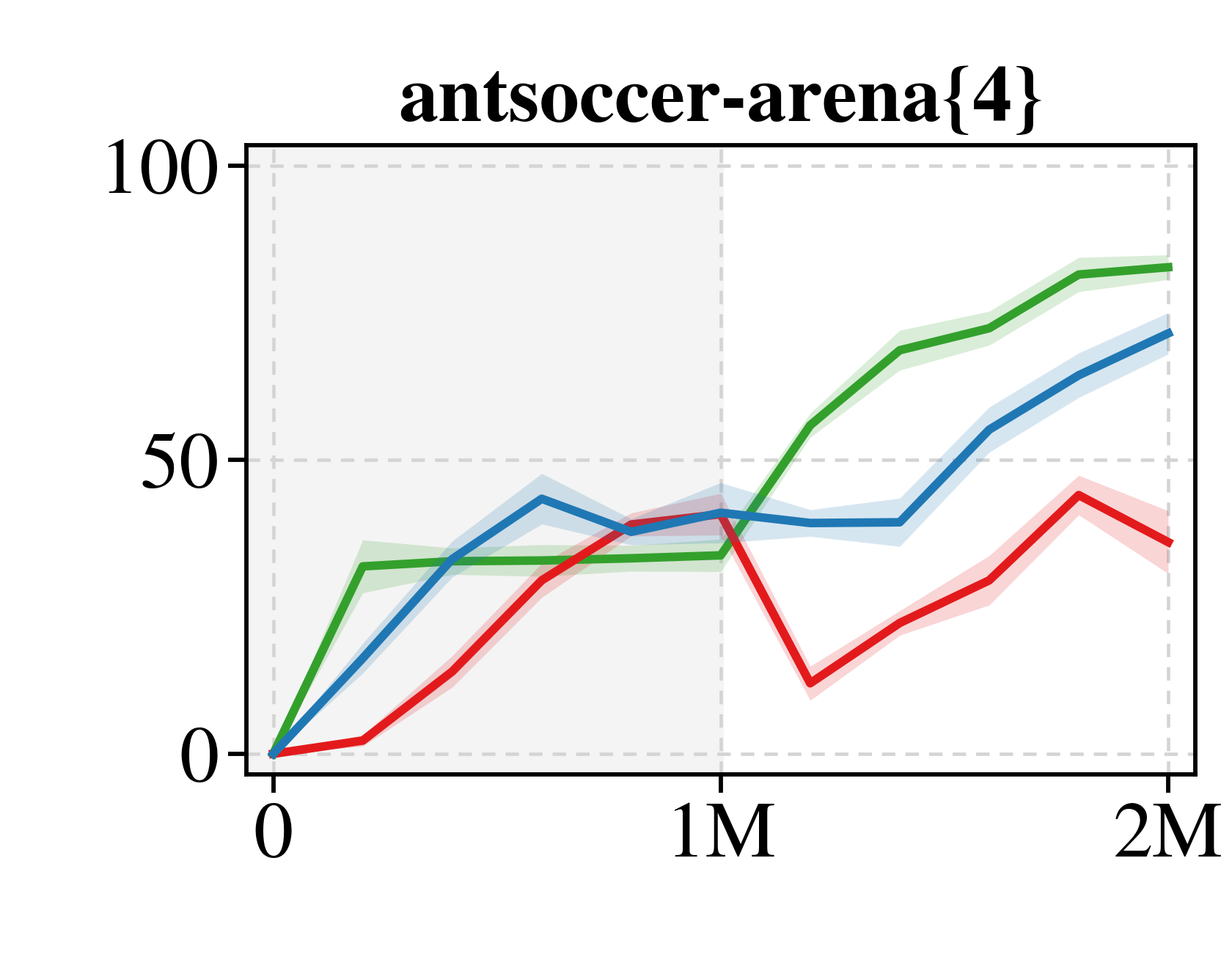}\hfill
\includegraphics[width=0.195\textwidth]{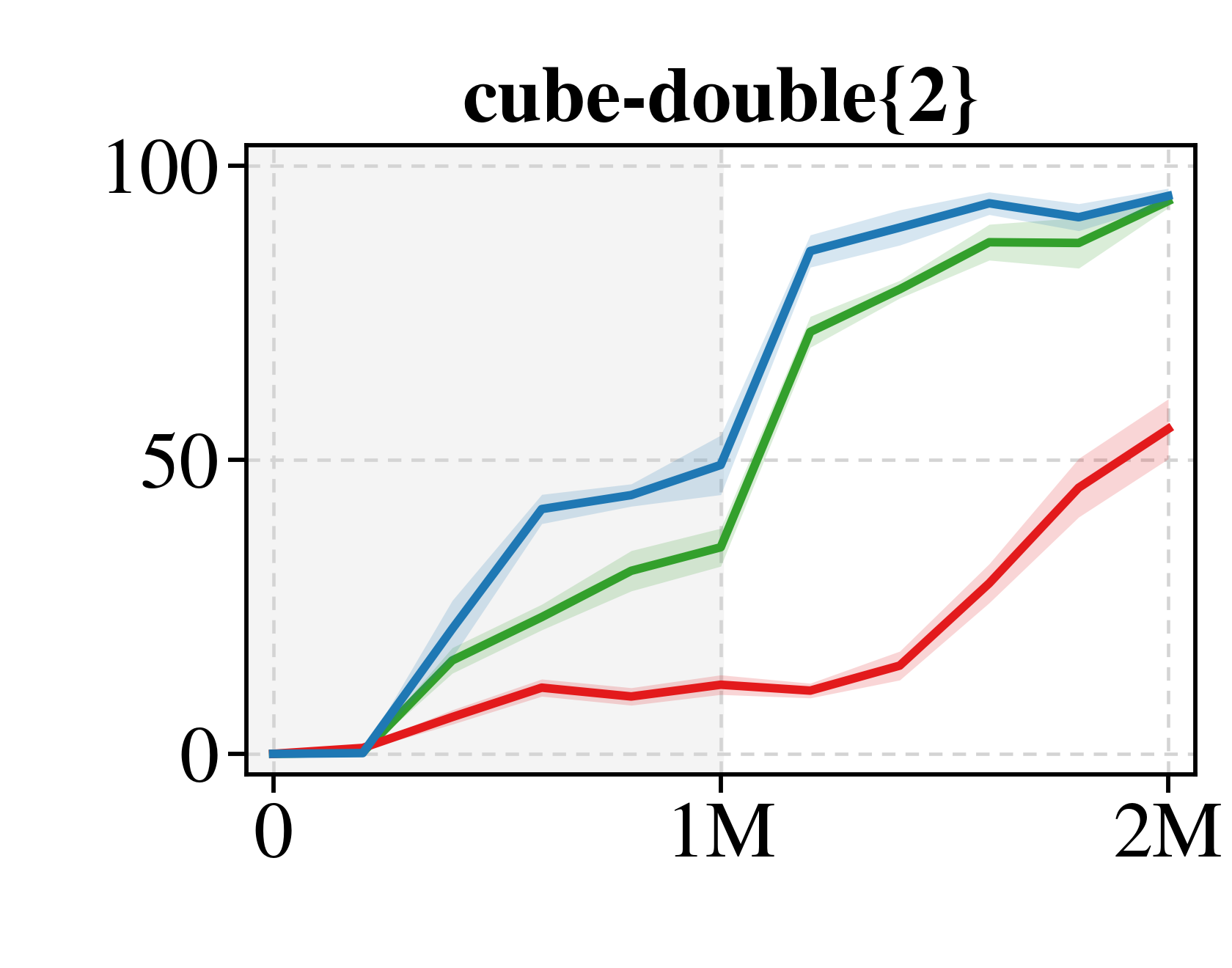}\hfill
\includegraphics[width=0.195\textwidth]{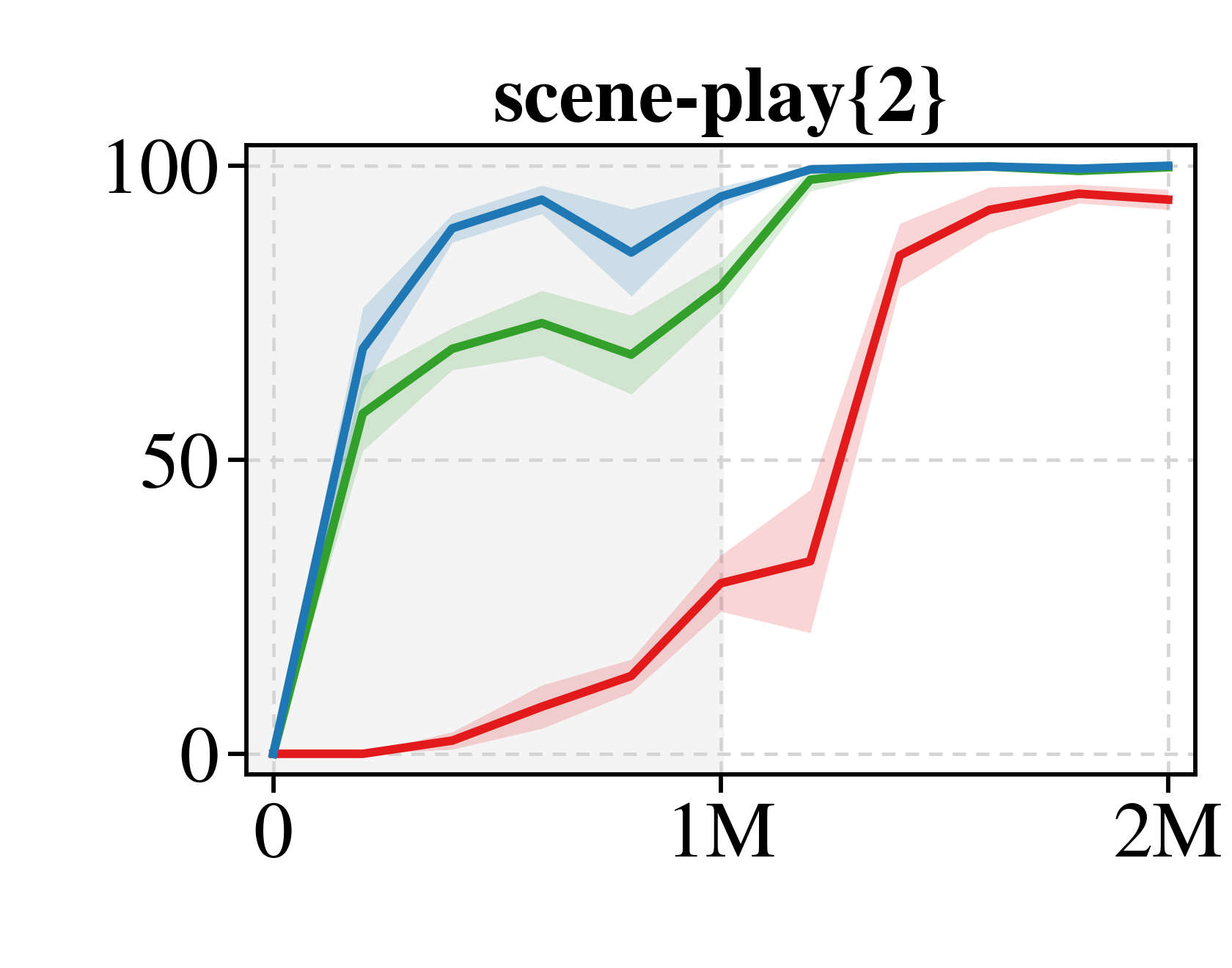}\hfill
\includegraphics[width=0.195\textwidth]{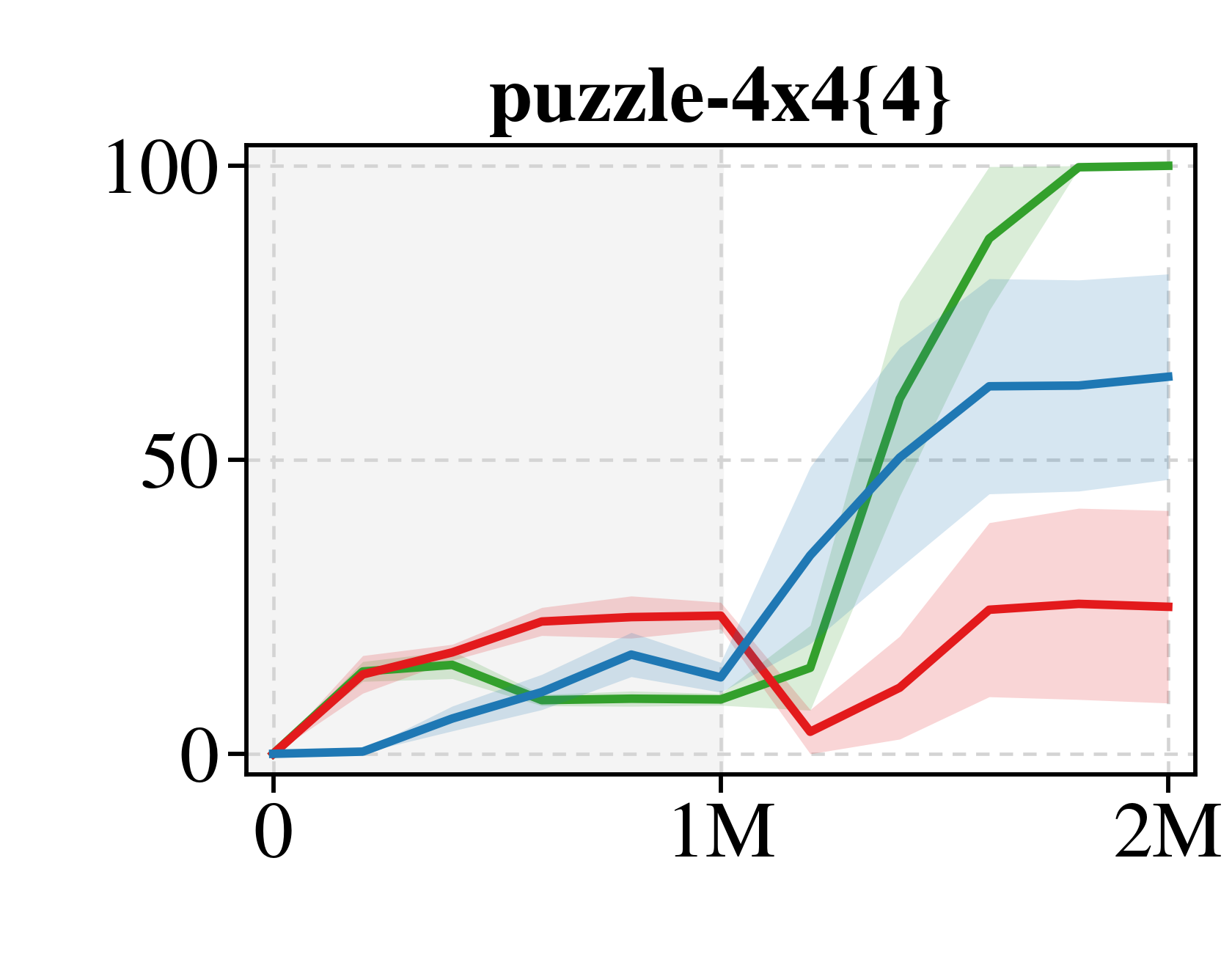}\\[2pt]
\includegraphics[width=0.195\textwidth]{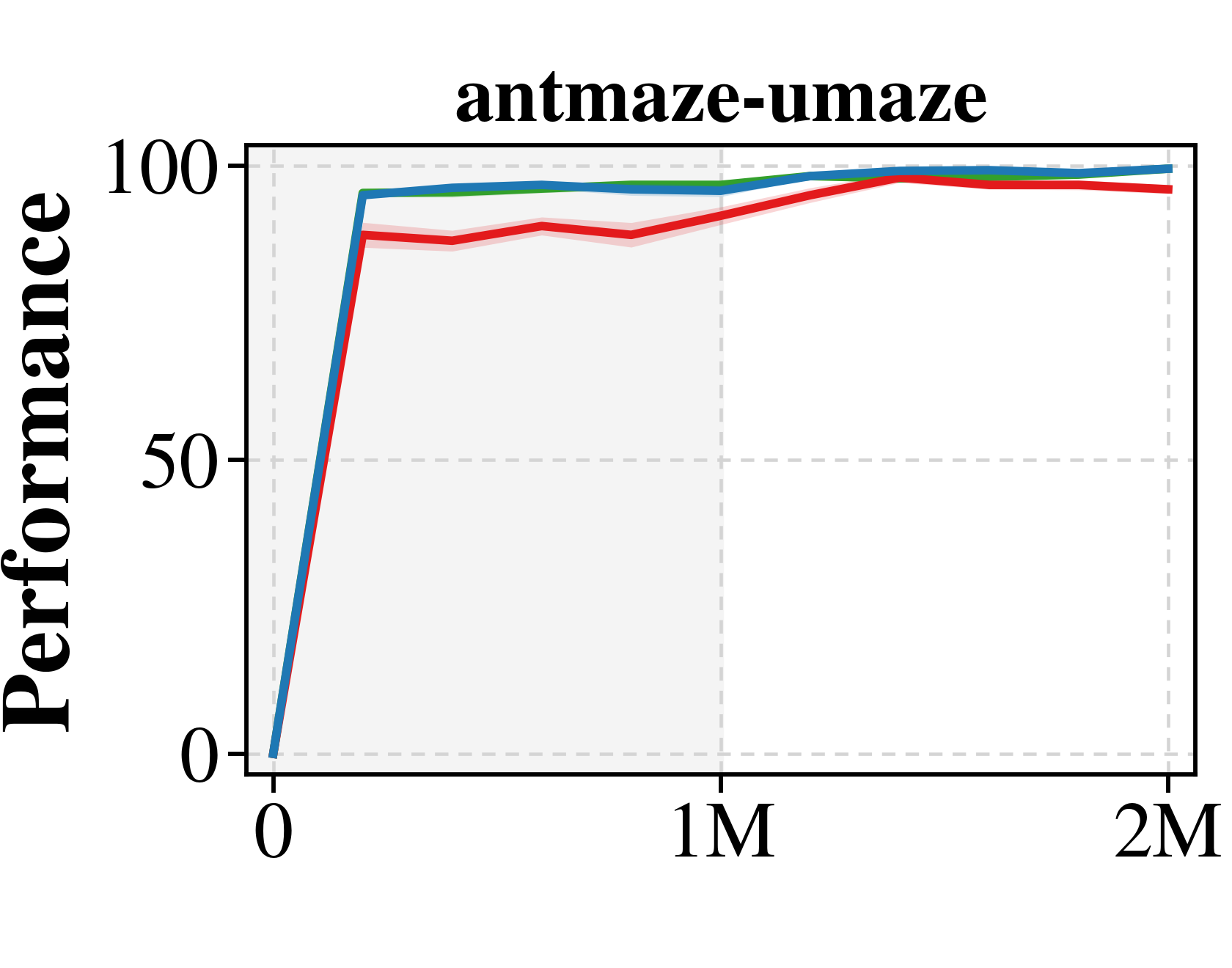}\hfill
\includegraphics[width=0.195\textwidth]{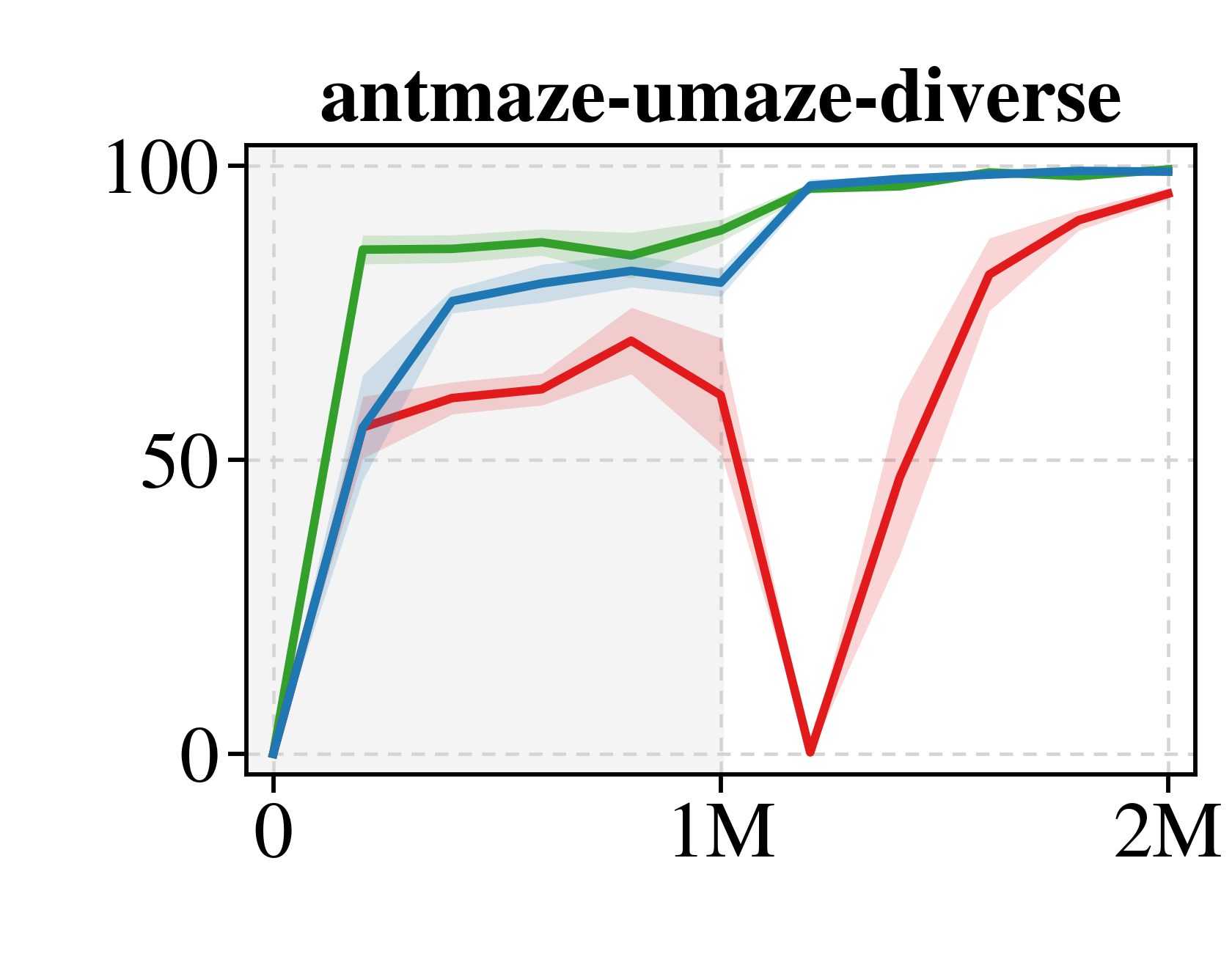}\hfill
\includegraphics[width=0.195\textwidth]{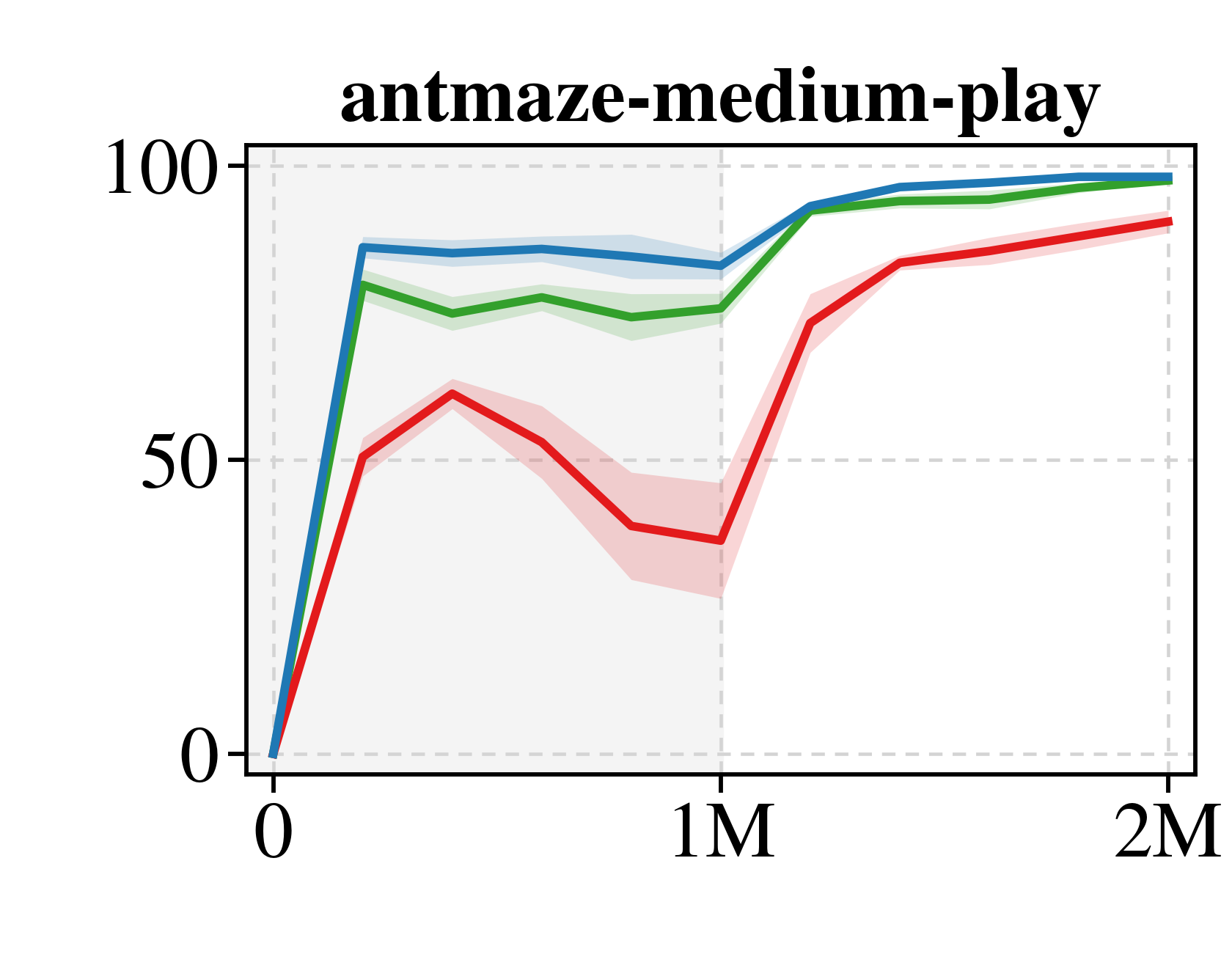}\hfill
\includegraphics[width=0.195\textwidth]{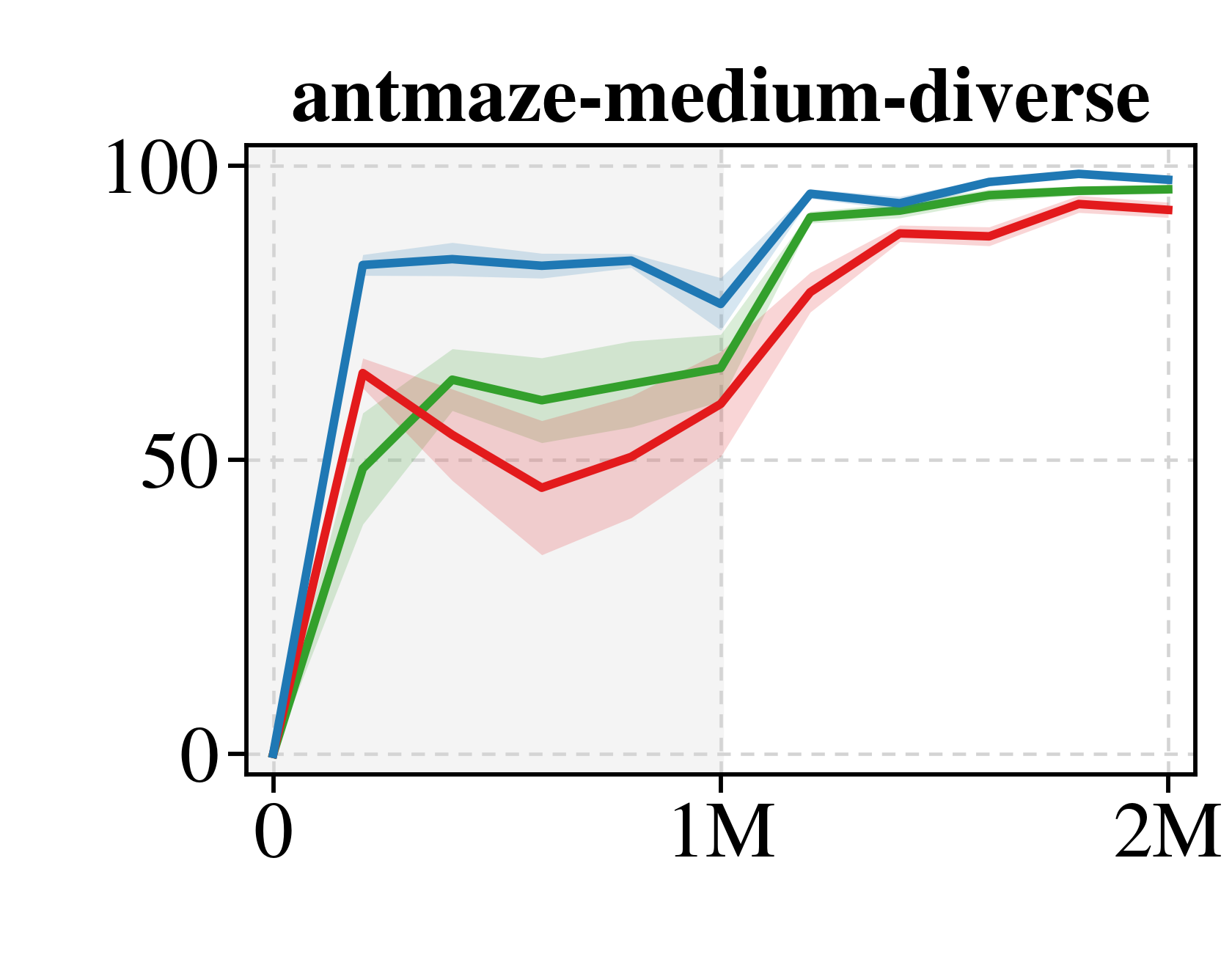}\hfill
\includegraphics[width=0.195\textwidth]{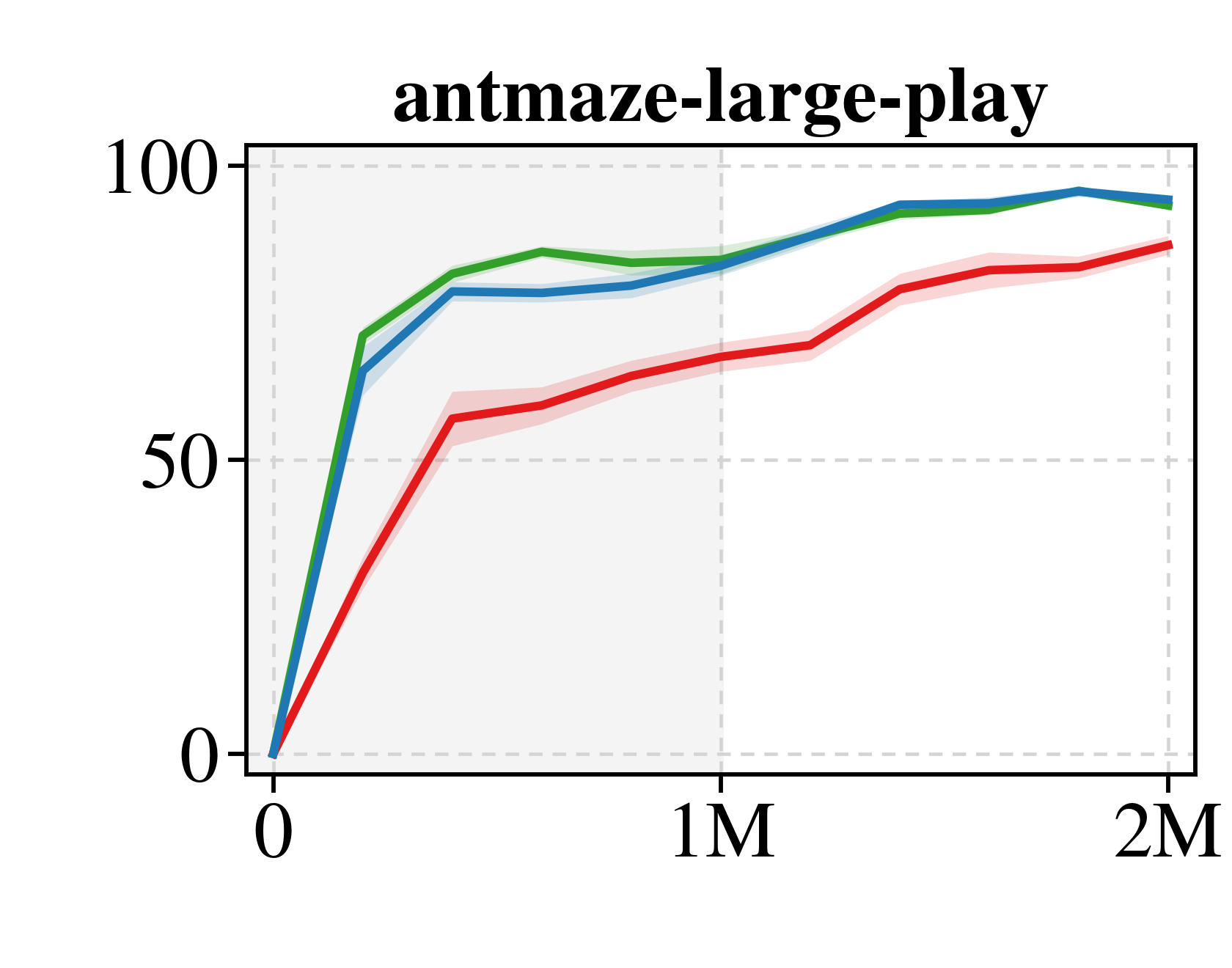}\\[2pt]
\includegraphics[width=0.195\textwidth]{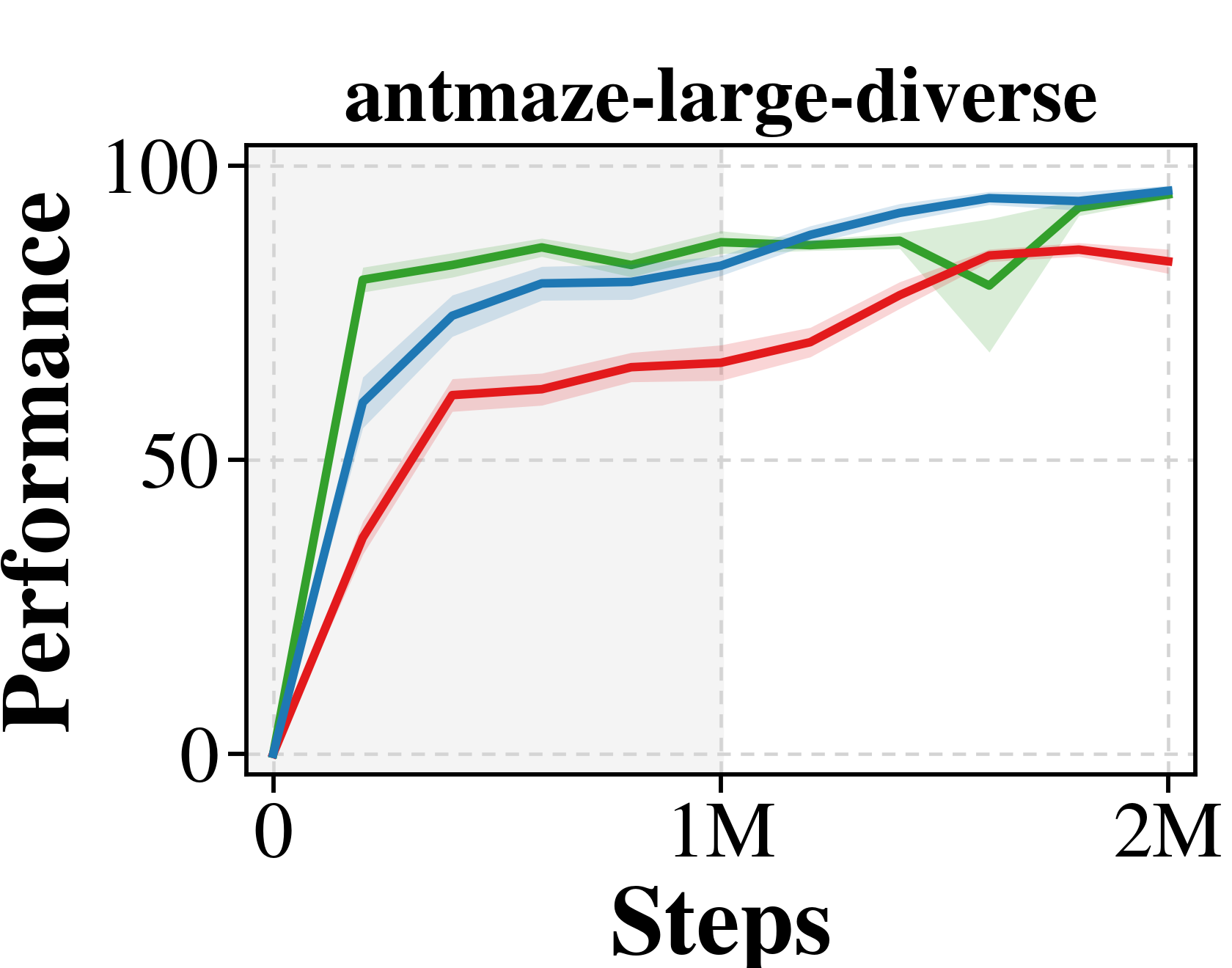}\hfill
\includegraphics[width=0.195\textwidth]{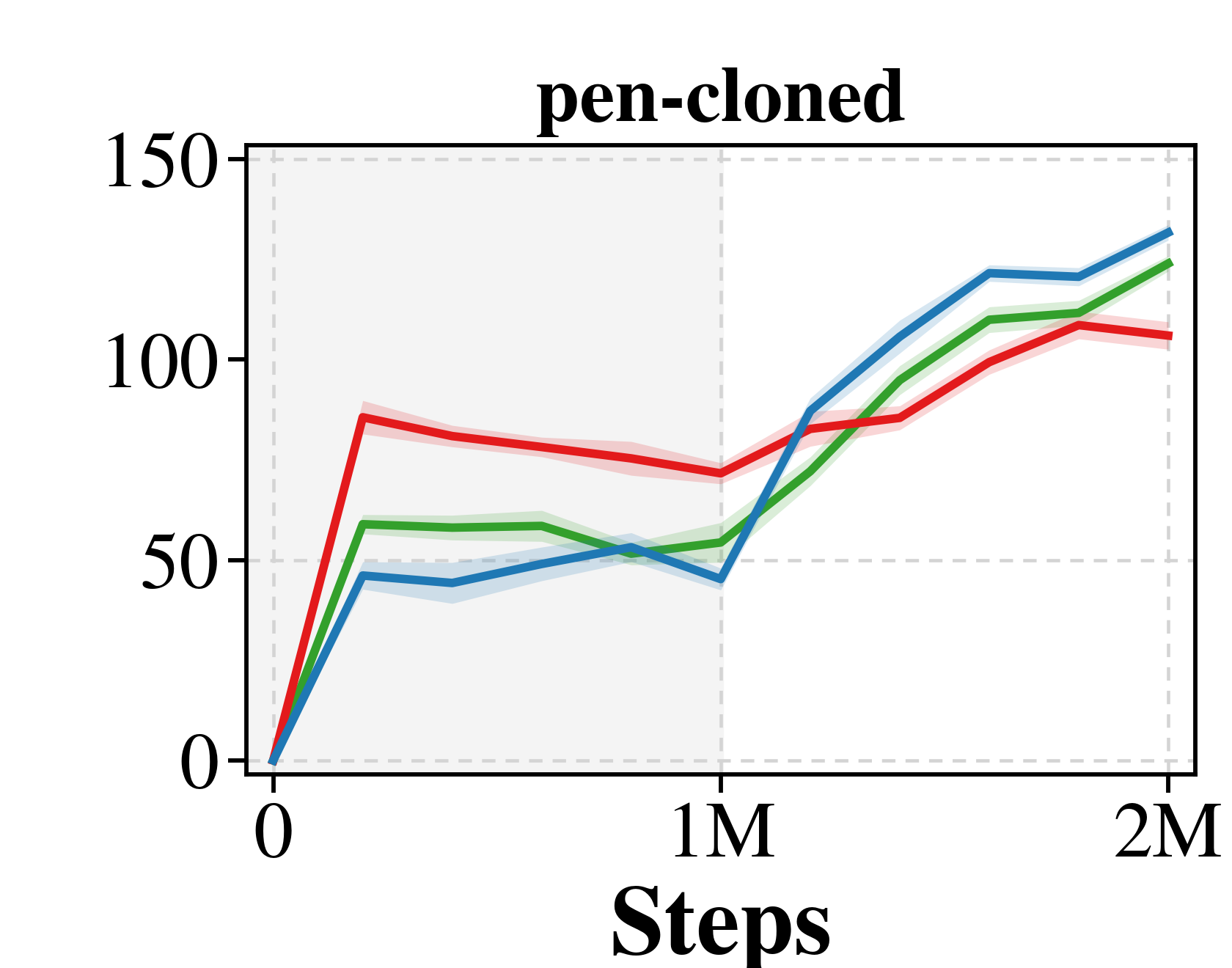}\hfill
\includegraphics[width=0.195\textwidth]{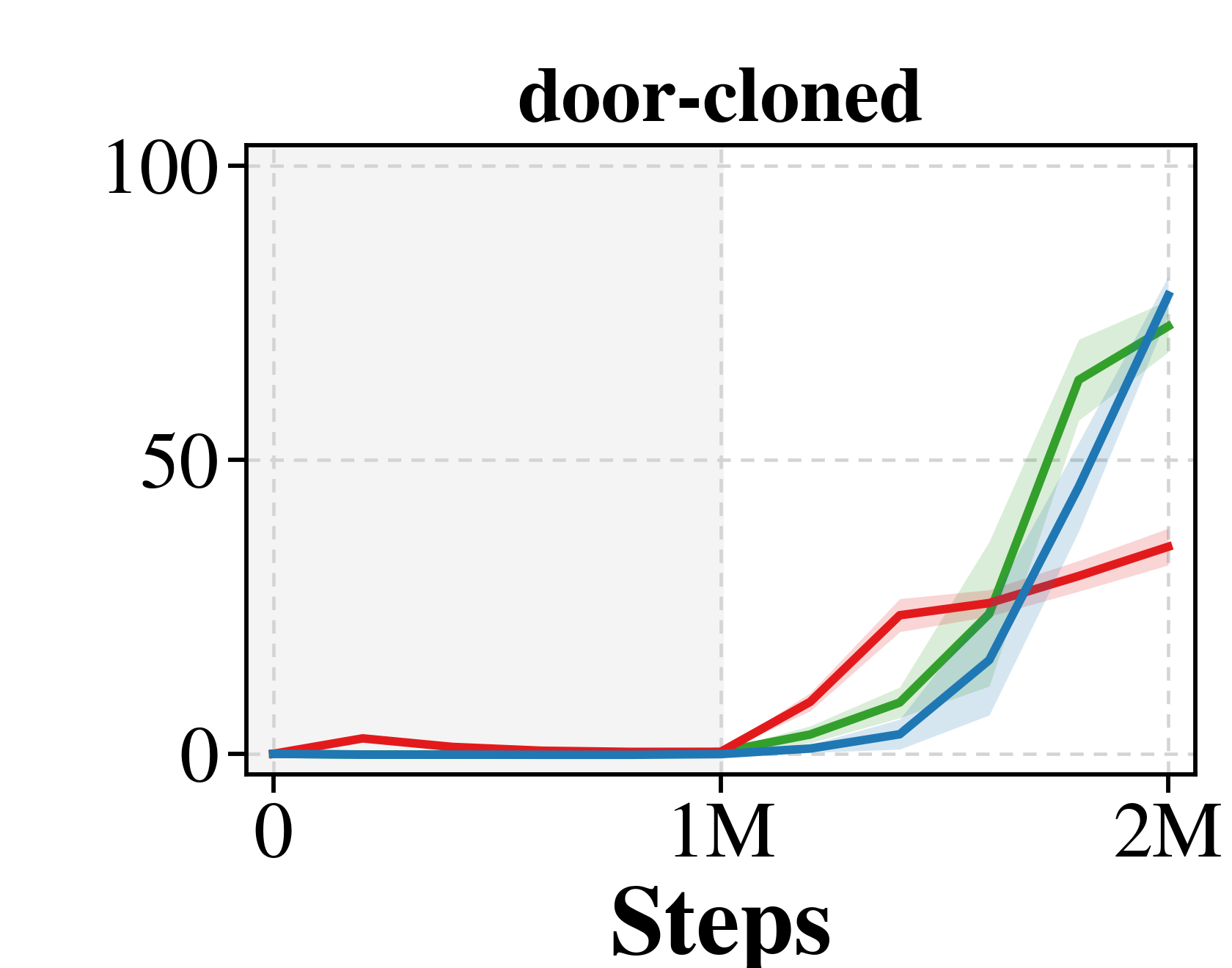}\hfill
\includegraphics[width=0.195\textwidth]{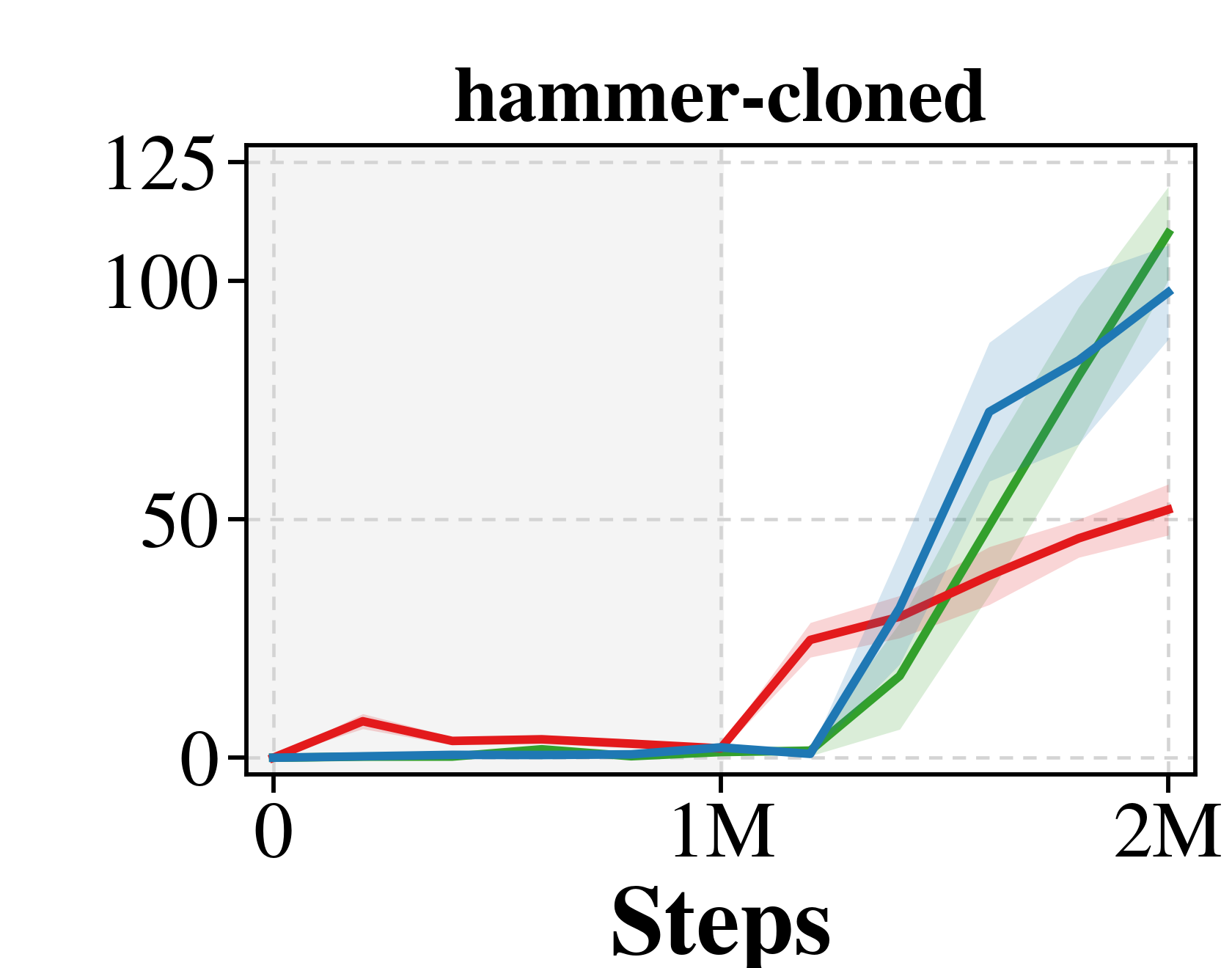}\hfill
\includegraphics[width=0.195\textwidth]{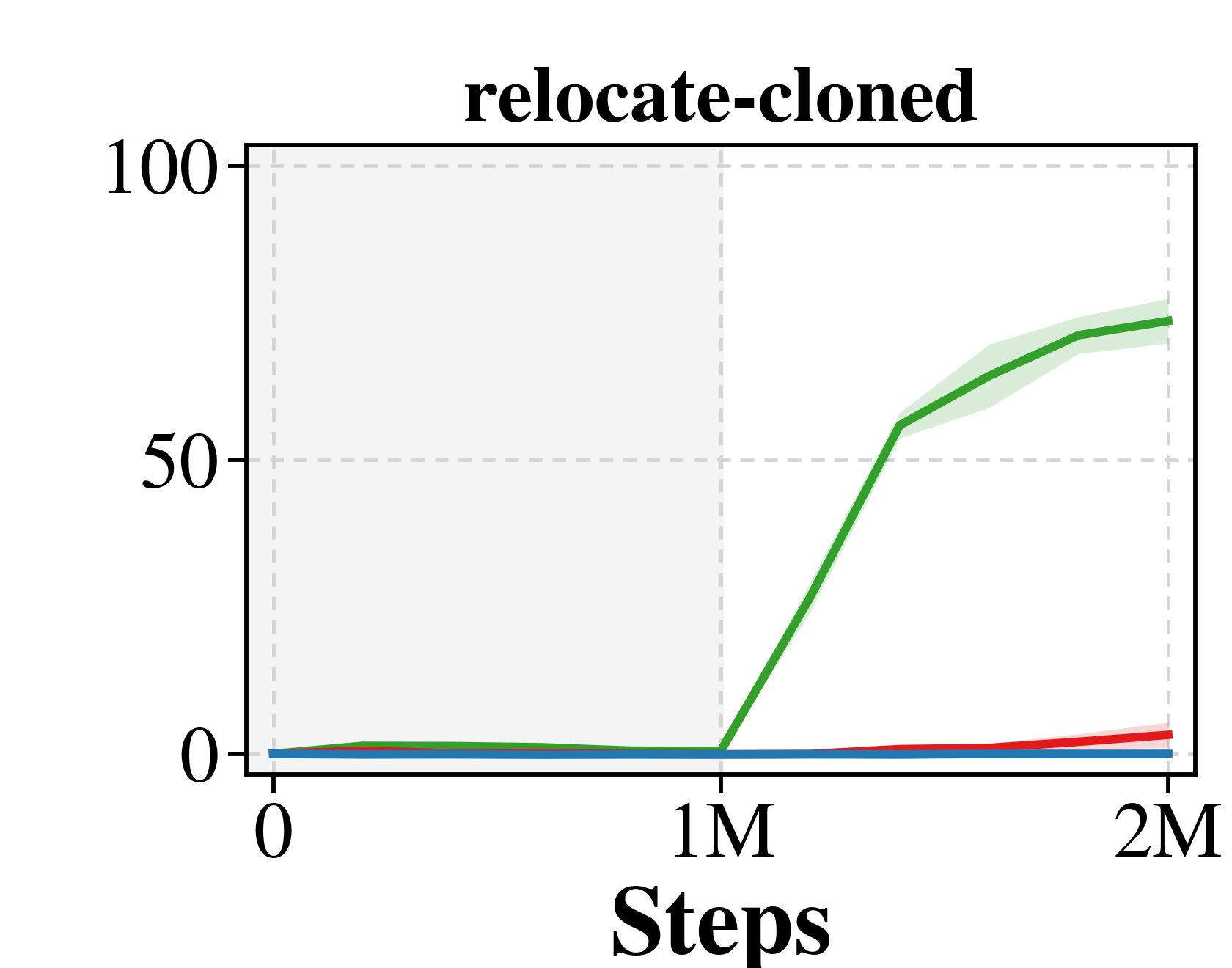}
\caption{\textbf{Offline-to-online RL results.} Online fine-tuning starts at 1M steps. The results are averaged over 8 random seeds and plotted every 200K steps.}
\label{fig:o2o}
\end{figure}

We report these 15 tasks to match the offline-to-online task set of FQL~\citep{park2025flow}. Across the evaluated OGBench and D4RL tasks, online fine-tuning improves over the offline checkpoint without modifying the OptiFlow training procedure. Newly collected transitions are incorporated directly into critic updates, VaBC reference-policy training, and transport-based target construction. These results suggest that the OT coupling structure is compatible with online fine-tuning and can extend beyond the strictly offline setting.

Some exceptions are puzzle-4$\times$4-play-singletask-task4-v0, which exhibits high seed variance under the default offline hyperparameters: five of eight seeds reach a score of 100, while three remain at 0; and adroit tasks that require hyperparameter tuning. Relocate-cloned task can be further tuned to show improving performance. We report this result without task-specific online tuning unless otherwise stated in Table~\ref{tab:o2o_hparams}. Overall, OptiFlow shows robust offline-to-online fine-tuning behavior across the evaluated tasks.

\newpage
\section*{NeurIPS Paper Checklist}

The checklist is designed to encourage best practices for responsible machine learning research, addressing issues of reproducibility, transparency, research ethics, and societal impact. Do not remove the checklist: {\bf The papers not including the checklist will be desk rejected.} The checklist should follow the references and follow the (optional) supplemental material.  The checklist does NOT count towards the page
limit. 

Please read the checklist guidelines carefully for information on how to answer these questions. For each question in the checklist:
\begin{itemize}
    \item You should answer \answerYes{}, \answerNo{}, or \answerNA{}.
    \item \answerNA{} means either that the question is Not Applicable for that particular paper or the relevant information is Not Available.
    \item Please provide a short (1--2 sentence) justification right after your answer (even for \answerNA). 
\end{itemize}

{\bf The checklist answers are an integral part of your paper submission.} They are visible to the reviewers, area chairs, senior area chairs, and ethics reviewers. You will also be asked to include it (after eventual revisions) with the final version of your paper, and its final version will be published with the paper.

The reviewers of your paper will be asked to use the checklist as one of the factors in their evaluation. While \answerYes{} is generally preferable to \answerNo{}, it is perfectly acceptable to answer \answerNo{} provided a proper justification is given (e.g., error bars are not reported because it would be too computationally expensive'' or ``we were unable to find the license for the dataset we used''). In general, answering \answerNo{} or \answerNA{} is not grounds for rejection. While the questions are phrased in a binary way, we acknowledge that the true answer is often more nuanced, so please just use your best judgment and write a justification to elaborate. All supporting evidence can appear either in the main paper or the supplemental material, provided in appendix. If you answer \answerYes{} to a question, in the justification please point to the section(s) where related material for the question can be found.

IMPORTANT, please:
\begin{itemize}
    \item {\bf Delete this instruction block, but keep the section heading ``NeurIPS Paper Checklist"},
    \item  {\bf Keep the checklist subsection headings, questions/answers and guidelines below.}
    \item {\bf Do not modify the questions and only use the provided macros for your answers}.
\end{itemize}


\begin{enumerate}

\item {\bf Claims}
    \item[] Question: Do the main claims made in the abstract and introduction accurately reflect the paper's contributions and scope?
    \item[] Answer: \answerYes{} 
    \item[] Justification: We clearly state the contribution points and the scope in the abstract and introduction that are discussed and demonstrated throughout the paper.

\item {\bf Limitations}
    \item[] Question: Does the paper discuss the limitations of the work performed by the authors?
    \item[] Answer: \answerYes{} 
    \item[] Justification: We provide a limitations section before the conclusion of the main paper in section \ref{sec:limitations}.

\item {\bf Theory assumptions and proofs}
    \item[] Question: For each theoretical result, does the paper provide the full set of assumptions and a complete (and correct) proof?
    \item[] Answer: \answerNA{} 
    \item[] Justification: The paper does not present formal theoretical results such as theorems, propositions, or convergence guarantees.

    \item {\bf Experimental result reproducibility}
    \item[] Question: Does the paper fully disclose all the information needed to reproduce the main experimental results of the paper to the extent that it affects the main claims and/or conclusions of the paper (regardless of whether the code and data are provided or not)?
    \item[] Answer: \answerYes{} 
    \item[] Justification: We provide the full code-base needed to exactly replicate the benchmark experiments. We include all hyperparameters and protocols in the Table \ref{tab:OptiFlow_shared_hparams}. The provided code-base can reproduce the experiments with simple guideline in the README.

\item {\bf Open access to data and code}
    \item[] Question: Does the paper provide open access to the data and code, with sufficient instructions to faithfully reproduce the main experimental results, as described in supplemental material?
    \item[] Answer: \answerYes{} 
    \item[] Justification: We include a git repository of the code-base which contains all experiment setups needed to run the experiments as we denote in the details within the Table \ref{tab:OptiFlow_shared_hparams} in the appendix.

\item {\bf Experimental setting/details}
    \item[] Question: Does the paper specify all the training and test details (e.g., data splits, hyperparameters, how they were chosen, type of optimizer) necessary to understand the results?
    \item[] Answer: \answerYes{} 
    \item[] Justification: We specify every eval protocol, setup, and hyperparameters in cleanly organized tables starting with the Table \ref{tab:OptiFlow_shared_hparams}.

\item {\bf Experiment statistical significance}
    \item[] Question: Does the paper report error bars suitably and correctly defined or other appropriate information about the statistical significance of the experiments?
    \item[] Answer: \answerYes{} 
    \item[] Justification: We report errors for all experimental results where appropriate. We state for each experimental table that the results are reported in standard error.

\item {\bf Experiments compute resources}
    \item[] Question: For each experiment, does the paper provide sufficient information on the computer resources (type of compute workers, memory, time of execution) needed to reproduce the experiments?
    \item[] Answer: \answerYes{} 
    \item[] Justification: We provide the details on compute in the discussion section \ref{app:compute}.
    
\item {\bf Code of ethics}
    \item[] Question: Does the research conducted in the paper conform, in every respect, with the NeurIPS Code of Ethics \url{https://neurips.cc/public/EthicsGuidelines}?
    \item[] Answer: \answerYes{} 
    \item[] Justification: Yes, we understand and abide by rules and conform with the NeurIPS Code of Ethics.

\item {\bf Broader impacts}
    \item[] Question: Does the paper discuss both potential positive societal impacts and negative societal impacts of the work performed?
    \item[] Answer: \answerYes{} 
    \item[] Justification: We include our statement as a separate section in Appendix~\ref{app:broad}.
    
\item {\bf Safeguards}
    \item[] Question: Does the paper describe safeguards that have been put in place for responsible release of data or models that have a high risk for misuse (e.g., pre-trained language models, image generators, or scraped datasets)?
    \item[] Answer: \answerNA{} 
    \item[] Justification: We release our code-base, but it is not subject to safeguards requirement.

\item {\bf Licenses for existing assets}
    \item[] Question: Are the creators or original owners of assets (e.g., code, data, models), used in the paper, properly credited and are the license and terms of use explicitly mentioned and properly respected?
    \item[] Answer: \answerYes{} 
    \item[] Justification: We properly cite all calls or uses of existing assets whenever appropriate.

\item {\bf New assets}
    \item[] Question: Are new assets introduced in the paper well documented and is the documentation provided alongside the assets?
    \item[] Answer: \answerYes{} 
    \item[] Justification: We provide our code-base as supplement materials as a link to anonymized repository with self-contained, explanatory README.

\item {\bf Crowdsourcing and research with human subjects}
    \item[] Question: For crowdsourcing experiments and research with human subjects, does the paper include the full text of instructions given to participants and screenshots, if applicable, as well as details about compensation (if any)? 
    \item[] Answer: \answerNA{} 
    \item[] Justification: Our work does not involve crowdsourcing nor research with human subjects.

\item {\bf Institutional review board (IRB) approvals or equivalent for research with human subjects}
    \item[] Question: Does the paper describe potential risks incurred by study participants, whether such risks were disclosed to the subjects, and whether Institutional Review Board (IRB) approvals (or an equivalent approval/review based on the requirements of your country or institution) were obtained?
    \item[] Answer: \answerNA{} 
    \item[] Justification: Our work does not involve crowdsourcing nor research with human subjects.

\item {\bf Declaration of LLM usage}
    \item[] Question: Does the paper describe the usage of LLMs if it is an important, original, or non-standard component of the core methods in this research? Note that if the LLM is used only for writing, editing, or formatting purposes and does \emph{not} impact the core methodology, scientific rigor, or originality of the research, declaration is not required.
    \item[] Answer: \answerNA{} 
    \item[] Justification: We do not use LLM for any important, original, or non-standard components.

\end{enumerate}

\end{document}